\documentclass{article}

 \usepackage[preprint]{neurips_2026}

\usepackage[utf8]{inputenc} %
\usepackage[T1]{fontenc}    %
\usepackage{hyperref}       %
\usepackage{url}            %
\usepackage{booktabs}       %
\usepackage{amsfonts}       %
\usepackage{nicefrac}       %
\usepackage{microtype}      %
\usepackage{xcolor}         %

\title{Robust Task-Aware State Estimation\\from Pretrained Perception}

\author{%
  Guy Azran \\
  CLAIR Lab \\
  Taub Faculty of Computer Science \\
  Technion -- Israel Institute of Technology \\
  Haifa 3200003, Israel \\
  \texttt{guy.azran@campus.technion.ac.il}
  \And
  Sarah Keren \\
  CLAIR Lab \\
  Taub Faculty of Computer Science \\
  Technion -- Israel Institute of Technology \\
  Haifa 3200003, Israel \\
  \texttt{sarahk@cs.technion.ac.il}
}

\usepackage{graphicx}
\usepackage{amsmath}
\usepackage{amsthm}
\usepackage[capitalize,noabbrev]{cleveref}
\usepackage{algorithm}
\usepackage{algpseudocode}
\usepackage{subcaption}
\usepackage{wrapfig}
\usepackage{comment}
\usepackage{thm-restate}
\usepackage{dsfont}
\usepackage{multirow}
\usepackage{rotating}
\usepackage{placeins}

\crefname{appendix}{Appendix}{Appendices}
\Crefname{appendix}{Appendix}{Appendices}

\newtheorem{example}{Example}
\newtheorem{definition}{Definition}

\newtheorem{claim}{Claim}
\newtheorem{lemma}{Lemma}

\newtheorem{corollary}{Corollary}

\DeclareMathOperator*{\argmin}{arg\,min}

\usepackage{xspace}

\newcommand{\pseFn}{\ensuremath{\phi}\xspace}

\newcommand{\fluentSet}[0]{\ensuremath{F}\xspace}

\newcommand{\fluent}[0]{\ensuremath{f}\xspace}

\newcommand{\beliefSet}{\ensuremath{b}\xspace}

\newcommand{\beliefState}[0]{\ensuremath{\beta}\xspace}
\newcommand{\taskBeliefState}[0]{\ensuremath{\beliefState^{\constraintSet}}\xspace}
\newcommand{\unconstBeliefState}[0]{\ensuremath{\tilde{\beliefState}}\xspace}

\newcommand{\factoredBelief}[0]{\ensuremath{\beliefState^\fluentSet}}

\newcommand{\probThresh}{\ensuremath{\theta}\xspace}

\newcommand{\prob}{\ensuremath{p}\xspace}
\newcommand{\constraintSet}{\ensuremath{\mathcal{C}}\xspace}
\newcommand{\constraint}{\ensuremath{c}\xspace}
\newcommand{\normalizer}{\ensuremath{Z}\xspace}
\newcommand{\dnormalizer}{\ensuremath{Z_{\text{relaxed}}}\xspace}

\newcommand{\obsSpace}[0]{\ensuremath{O}\xspace}
\newcommand{\obs}[0]{\ensuremath{o}\xspace}
\newcommand{\stateSpace}[0]{\ensuremath{S}\xspace}

\newcommand{\state}[0]{\ensuremath{s}\xspace}

\newcommand{\beliefSize}[0]{\ensuremath{k}\xspace}

\newcommand{\queue}[0]{\ensuremath{Q}\xspace}
\newcommand{\queueExtracted}[0]{\ensuremath{\stateSpace^{\text{ext}}}\xspace}

\newcommand{\indicator}[0]{\ensuremath{\mathds{1}}\xspace}

\newcommand{\absThresh}{\ensuremath{\tau}\xspace}

\usepackage[acronym,numberedsection]{glossaries}
\glsdisablehyper

\makeglossaries

\newglossaryentry{symState}{
  name={symbolic state},
  description={A symbolic representation of the world state in terms of features relevant to the planning task, typically expressed in a formal language such as STRIPS}%
}

\newglossaryentry{symGround}{
  name={symbolic grounder},
  description={A function that maps \glspl{wState} to symbolic features}
}

\newglossaryentry{taskAction}{
  name={task action},
  description={A high-level action that changes the \gls{symState} of the world}
}

\newglossaryentry{queryAction}{
  name={query action},
  description={An action that provides information about the current \gls{symState} without changing it}
}

\newglossaryentry{tgtFluent}{
  name={target fluent},
  description={A fluent whose truth value becomes known to the agent after executing a \gls{queryAction}}
}

\newglossaryentry{belSet}{
  name={belief set},
  description={A set of possible initial \glspl{symState} representing uncertainty about the true initial state in conformant planning}
}

\newglossaryentry{belState}{
  name={belief},
  description={A probability distribution over possible initial \glspl{symState} representing uncertainty about the true initial state in conformant probabilistic planning}
}

\newglossaryentry{factoredBelState}{
  name={factored belief},
  description={A function that assigns independent probabilities to individual features, representing the marginal probability that each feature is true}
}

\newglossaryentry{cumErr}{
  name={cumulative \gls{factoredBelState} error},
  description={A measure of the total discrepancy between a \gls{symState} and the probabilities assigned by a \gls{factoredBelState} across all features}
}

\newglossaryentry{unconstBelState}{
  name={unconstrained belief},
  description={A \gls{belState} computed via a \gls{factoredBelState} by assuming conditional independence between features, before applying task-specific constraints (see \cref{eq:unconst-belief})}
}

\newglossaryentry{taskBelState}{
  name={task belief},
  description={A \gls{belState} computed by conditioning the \gls{unconstBelState} on task-specific constraints (see \cref{eq:calc-belief})}
}

\newacronym[description=A multimodal foundation model that combines text and image input]
  {vlm}
  {VLM}
  {Vision-Language Model}

\newacronym[description=A formalism for classical planning]
  {strips}
  {STRIPS}
  {Stanford Research Institute Problem Solver}

\newacronym[description=A planning paradigm where the initial state is not fully known]
  {cp}
  {CP}
  {Conformant Planning}

\newacronym[description=A planning paradigm where the initial state is not fully known and additional information is acquired through sensing during execution]
  {ctp}
  {CtP}
  {Contingent Planning}

\newacronym[description=\Acrlong{cp} where the initial state is represented as a probability distribution over possible states]
  {cpp}
  {CPP}
  {Conformant Probabilistic Planning}

\newacronym[description=The smallest subset of states whose cumulative probability meets or exceeds a given threshold (see \cref{def:mlss})]
  {mbs}
  {MBS\textsubscript{\probThresh}}
  {\probThresh-confidence Minimal Belief Set}

\newacronym[description=The algorithm that produces the \acrshort{mbs}]
  {pravas}
  {PRaVaSS}
  {Probability-Ranked Valid State Search}
  
\newacronym[description=A language for defining task planning problems]
  {pddl}
  {PDDL}
  {Planning Domain Definition Language}

\newacronym[description=A function that provides probabilistic estimates of symbolic features given sensory observations]
  {sse}
  {SSE}
  {Symbolic State Estimator}

  \newacronym[description=A planning system for classical planning problems]
  {fd}
  {FD}
  {Fast Downward}

\newacronym[description=A model for partially observable decision-making with Markovian assumptions of action transitions\, rewards\, and sensing]
  {pomdp}
  {POMDP}
  {Partially Observable Markov Decision Process}
 
\newacronym[description=The probability-maximizing state feature assignment]
  {map}
  {MAP}
  {Maximum A Posteriori}

\newacronym[description=An adaptive choosing of the \gls{mbs} via binary search for the largest subset for which there exists a plan]
  {ams}
  {AMS}
  {Adaptive MBS Search}

\newcommand{\coderepo}{\url{https://github.com/CLAIR-LAB-TECHNION/pravass_neurips2016}}

\begin{document}

\maketitle

\begin{abstract}
  Agents that perform state estimation using pretrained perception models, e.g., \glspl{vlm}, face the fundamental challenge that predictions are uncertain. Still, most approaches determinize these predictions into a single best-guess state and make decisions as if it were ground truth.
This is brittle because when the guess is wrong, the agent's decisions may be unsafe or infeasible.
At the other extreme, maintaining a full probability distribution over symbolic states is often intractable.
We propose a middle path, drawing on the underlying probabilities the perception model already produces (e.g., a \gls{vlm}'s output-layer token distributions) to construct a \emph{\gls{mbs}}, a minimal subset of states whose cumulative probability meets a user-specified belief-mass threshold.
The minimality of an \gls{mbs} ensures that the agent considers only the most likely states, enhancing robustness in downstream decision-making while maintaining computational tractability.
Using a novel three-layered \gls{belState} system, we efficiently approximate individual state probabilities from per-feature predictions and task-level constraints.
On this belief construction, we base our \gls{mbs} construction, while maintaining only a lean factored representation of the \gls{belState}.
Our algorithm, \emph{\gls{pravas}}, takes advantage of the unique structure of this \gls{belState} system to construct an \gls{mbs} with correctness guarantees and polynomial complexity in the unconstrained case.
Using a \gls{vlm} as the backbone of our state estimator, we validate our method's theoretical properties, scalability with problem size, and improvement of downstream decision-making when used in a task-planning pipeline.

\end{abstract}

\glsresetall
\section{Introduction}

Autonomous agents increasingly rely on pretrained perception for state estimation from sensory input.
For example, an agent may query a \gls{vlm} to determine the truth value of semantic facts from images \cite{yangGuidingLongHorizonTask2025,azranS3ESemanticSymbolic2025,liangVisualPredicatorLearningAbstract2024}.
The resulting symbolic facts are passed to a task planner that selects high-level actions.
This pipeline assumes, almost universally, that the perception output can be collapsed to a single most likely interpretation of the scene.
However, this is a brittle assumption.

Pretrained perception models' predictions are uncertain.
With \glspl{vlm}, visual ambiguity, occlusion, lighting, noise, model error, and other factors all contribute to noisy predictions, where confidence can vary widely across state features.
When the decision maker commits to a single guessed state and that guess is wrong, this may result in attempting unsafe actions, violating user-defined guardrails, or failing outright.

\begin{wrapfigure}{l}{0.4\textwidth}
    \centering
    \includegraphics[width=1\linewidth]{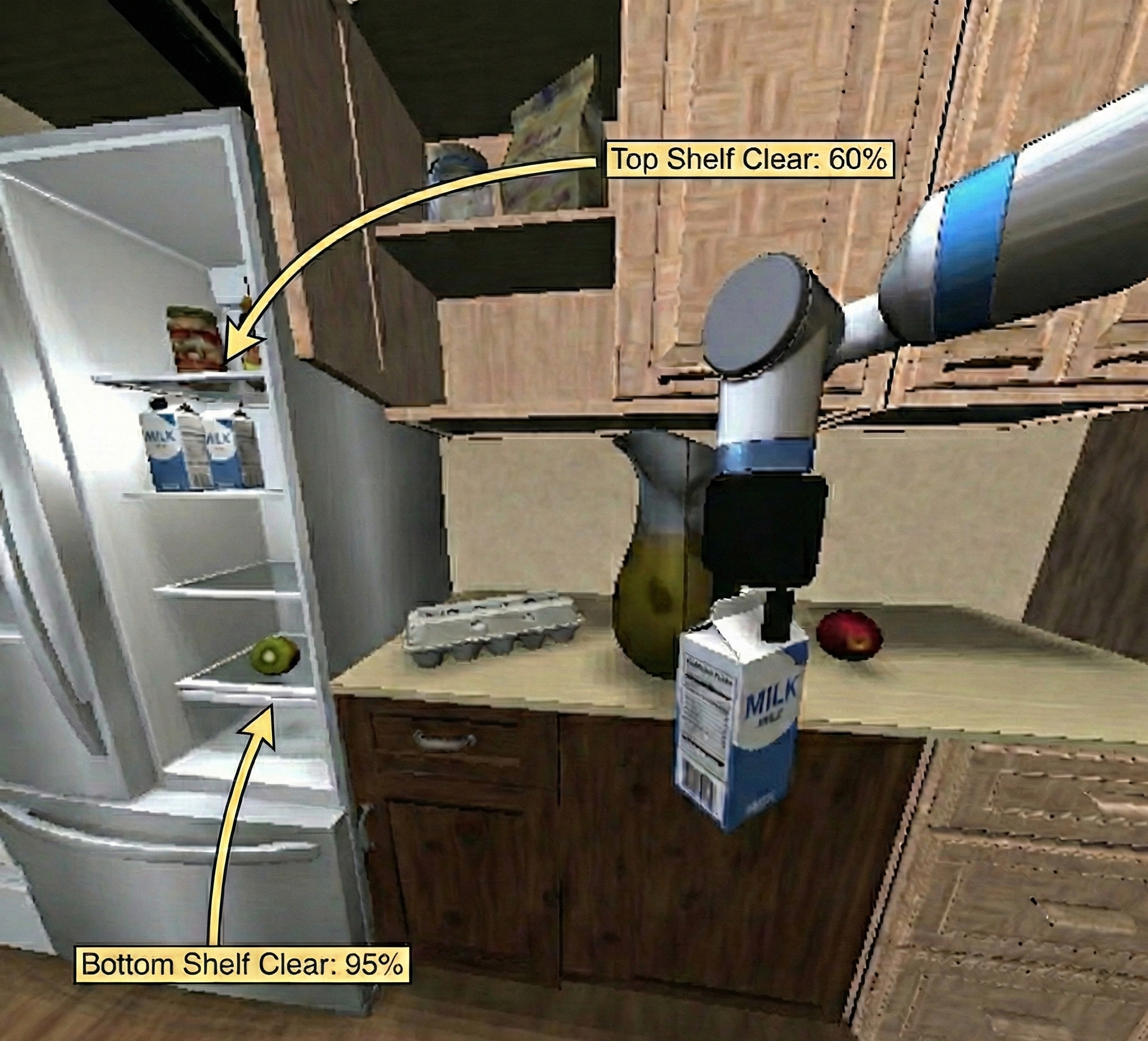}
    \caption{Grocery task from \cref{ex:kitchen}.}
    \label{fig:ex-kitchen}
    \vspace{-1\baselineskip}
\end{wrapfigure}

\begin{example}[Kitchen Robot]\label{ex:kitchen}
    A mobile manipulator is told to ``put away the groceries'', as depicted in \cref{fig:ex-kitchen}.
    It uses a task planner that operates over state features, e.g., \verb|on-counter(eggs)|, \verb|is-open(fridge)|, \verb|clear(shelf)|.
    Using the onboard camera, the agent attempts to ground high-level state features by querying a \gls{vlm} with semantic queries, e.g., ``Is the top shelf clear?'', ``Are the eggs on the counter?'', ``Is the fridge door open?''.
    The common robotics pipeline collapses the answers into a single most likely state by accepting the \gls{vlm}'s textual output ``Yes'' or ``No'', and plans as if that were the ground truth.
But the scene is perceptually ambiguous: a glare on the fridge handle, a partially occluded carton, a bag blocking the shelf, etc.
Suppose the \gls{vlm}'s underlying probabilistic outputs indicate that \verb|clear(top-shelf)| is true with only 60\% confidence, while \verb|clear(bottom-shelf)| is true with 95\% confidence.
If the state estimation determinizes \verb|clear(top-shelf)=true|, the agent may execute a \verb|place(milk,top-shelf)| action and collide with a hidden object, knocking a glass jar off the shelf.
\end{example}

In \cref{ex:kitchen}, the planner did not fail because it planned poorly, but because it planned from the wrong initial state.
The root cause is brittle state estimation.
At the other extreme, maintaining a full distribution over symbolic states and performing belief-space planning would be safer, but is generally intractable in combinatorial task-planning spaces.

We propose a practical middle ground and construct a \emph{\gls{mbs}}: a minimal subset of states whose cumulative probability satisfies a user-specified belief-mass threshold.
\Gls{mbs} is a compact, task-consistent abstraction of the underlying uncertainty that captures the most likely states while excluding low-probability ones, enabling robust downstream decision-making without incurring the full complexity of belief-space planning.
We present a novel three-layered \gls{belState} system over which the \gls{mbs} is constructed, where the distribution over all states is represented by a lean factored representation and task-level constraints.
We further propose an algorithm, \emph{\gls{pravas}}, which constructs an \gls{mbs} with correctness guarantees, and polynomial complexity in the unconstrained case, by taking advantage of the unique structure of this \gls{belState} system.

Returning to \cref{ex:kitchen}, rather than committing to a single state where the top shelf is clear, \gls{pravas} selects plausible states that include both possibilities.
This uncertainty-aware state representation supports the safer decision to place the milk on the less accessible bottom shelf.

This problem sits at the intersection of neurosymbolic abstraction, \gls{vlm}-based planning, and planning under uncertainty.
In neurosymbolic abstraction, early approaches learned state abstractions by interacting with the environment \cite{ahmetogluDeepSymDeepSymbol2022,konidarisSkillsSymbolsLearning2018,asaiClassicalPlanningDeep2022,asaiUnsupervisedGroundingPlannable2019}.
In \gls{vlm}-based planning, recent work uses pretrained \glspl{vlm} rather than relying on dangerous environment interactions, either directly as planners that generate action sequences from visual observations \cite{yangGuidingLongHorizonTask2025,duanManipulateAnythingAutomatingRealWorld2024,huLookYouLeap2023}, or as grounders that convert visual observations into symbolic states for downstream planning \cite{azranS3ESemanticSymbolic2025,liangVisualPredicatorLearningAbstract2024,chenLLMStateOpenWorld2024,dingRobotTaskPlanning2024,zhangDKPROMPTDomainKnowledge2024}.
Across these lines of work, perception output is typically a single deterministic symbolic state.
In contrast, planning under uncertainty takes into consideration the uncertainty in the system's knowledge, e.g., by modeling the state space probabilistically \cite{kurniawatiSARSOPEfficientPointBased2008,kaelblingPlanningActingPartially1998} or considering multiple states at a time \cite{bonetPlanningPartialObservability2011,palaciosCompilingUncertaintyAway2009}, but typically assumes a small state space due to computational intractability, inspiring recent work to consider less restrictive belief representations \cite{kerenReasoningPlanRobustness2020,taigCompilingConformantProbabilistic2013}.
Our work combines these ideas by using the probabilistic outputs of pretrained perception models to construct a compact, task-consistent \gls{belSet} over symbolic states that can be used for robust downstream decision-making.

Methodologically, the \gls{mbs} objective is related to highest-density regions \cite{hyndmanComputingGraphingHighest1996} and to set-valued prediction, which seeks the smallest prediction set containing the true label with guaranteed coverage \cite{sadinleLeastAmbiguousSetValued2019}.
Here, however, the ``labels'' are structured symbolic states from a combinatorial space, subject to task-specific validity constraints.
Highest-density and top-$k$ enumeration do not handle such constraints without enumerating the space, and set-valued classifiers do not exploit task structure, whereas \gls{pravas} extracts the most likely task-valid states lazily, without materializing the exponentially large state space (\cref{thm:mlss}).

The contributions of this work are as follows:
\begin{itemize}
    \item \textbf{Constrained belief representation}: We propose a three-layered \gls{belState} system based on task-level information and per-feature predictions from a single observation, which conditions a factored product distribution on task-derived validity constraints to efficiently extract individual state probabilities from a compact \gls{belState} representation.
    \item \textbf{\Gls{mbs}}: We define \gls{mbs} as the minimum-cardinality set of valid states whose cumulative probability meets a user-specified threshold and demonstrate its utility as a compact, task-consistent abstraction for downstream decision-making.
    \item \textbf{\Gls{pravas} algorithm}: We present \gls{pravas}, an algorithm for computing an \gls{mbs} with correctness and complexity guarantees, and release an open-source implementation.\footnote{\coderepo}
\end{itemize}

The \gls{mbs} can serve as input to a range of downstream reasoners.
Our empirical evaluation demonstrates the effectiveness of \gls{mbs} as the initial \gls{belSet} in a task-planning pipeline on a variety of domains, showing that the uncertainty abstraction materially improves plan safety and success.
Using a \gls{vlm} as the backbone of our state estimator, we evaluate \gls{pravas} on three domains spanning noisy visible state features and genuinely hidden initial-state features, demonstrating faithful coverage, compactness, scalability, and downstream utility.

\glsresetall
\section{From perception to task-consistent belief}\label{sec:problem}

\begin{figure}[t]
   \centering
    \includegraphics[width=1\linewidth]{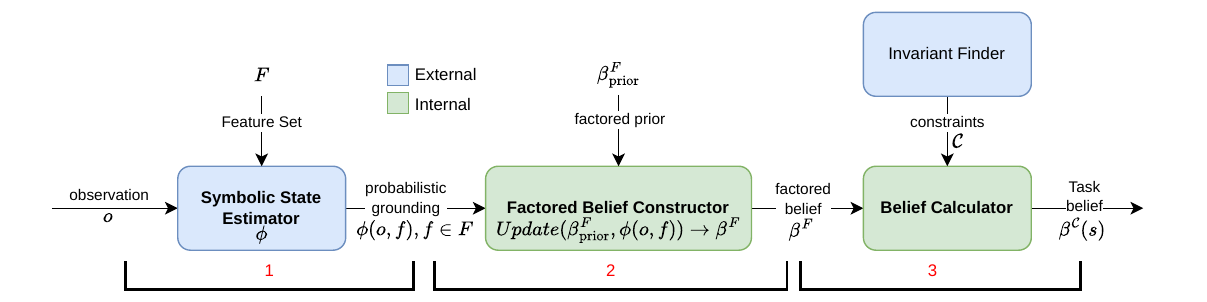}
    \caption{The three-layered \gls{belState} pipeline. (1) An observation is processed by the \acrshort{sse} to obtain per-feature probabilities. (2) These are updated with an optional prior to produce the \gls{factoredBelState}. (3) The \gls{factoredBelState} and a set of task-specific constraints are then used to compute the \gls{belState}.}
    \label{fig:belief-pipeline}
\end{figure}

To provide effective state estimation, the agent must assign probabilities to individual states. Because this is intractable for large state spaces, we propose a novel \gls{factoredBelState} which is represented by a distribution over a set of state features.
Using a pretrained perception model, the agent converts a raw sensory observation into per-feature probability distributions, which can be updated with an optional prior.
We combine this \gls{factoredBelState} with autonomously extracted task-specific constraints to calculate a \gls{belState} for individual states without explicitly enumerating all possible states. The full pipeline for this three-layered \gls{belState} system is illustrated in \cref{fig:belief-pipeline}.

Robotic manipulation is a natural motivating domain, since a \gls{vlm} can provide useful information about the environment, as in \cref{ex:kitchen}.
However, the formulation applies to any system that produces symbolic facts from pretrained perception.
The key requirement is a function that maps observations to per-feature confidence scores.
We show how this probability mapping is used to compute the probability for a given state using our three-layered \gls{belState} system.

\subsection{Factored belief}\label{sec:factored-belief}

\textbf{Probabilistic grounding.}
In the first stage of the pipeline, we convert a raw observation into per-feature probability distributions. For simplicity, we treat features as Boolean-valued.
As such, for a feature space \fluentSet, the state space is simply $2^\fluentSet$, where a state $\state \in 2^\fluentSet$ is a subset of features that are true in that state.
Alternatively, we write $\state(\fluent) = \indicator[\fluent \in \state]$.
This formulation naturally extends to other feature types, e.g., categorical features, by representing their distributions appropriately.

Let \obsSpace be an observation space and let \fluentSet be a feature space. We have access to a \emph{\gls{sse}} $\pseFn: \obsSpace \times \fluentSet \to [0,1] \cup \{\textsc{null}\}$.
For a given observation $\obs \in \obsSpace$ (e.g., an image) and for each feature $\fluent \in \fluentSet$ (e.g., ``the top shelf is clear''), the \gls{sse} (e.g., a \gls{vlm}) is queried to produce $\pseFn(\obs, \fluent) \in [0,1]$, representing its confidence that $\fluent$ holds in the observed scene.
We also allow $\pseFn(\obs, \fluent) = \textsc{null}$ when the observation provides insufficient evidence to assess $\fluent$, e.g., when querying whether a shelf is clear when the fridge door is closed.

In practice, the \gls{sse} is commonly realized by a \gls{vlm} \cite{azranS3ESemanticSymbolic2025,merlerViPlanBenchmarkVisual2025,zhangDKPROMPTDomainKnowledge2024}.
Given an image observation $\obs$ and a feature-specific query (e.g., ``Is the top shelf clear?''), a forward pass yields log-probabilities over the vocabulary, from which we extract:
\begin{equation*}
    \Pr_{\text{VLM}}(\text{``Yes''} | \obs, \text{``Is the top shelf clear?''})
    \quad
    \Pr_{\text{VLM}}(\text{``No''} | \obs, \text{``Is the top shelf clear?''})
\end{equation*}
Normalizing these two values gives the estimated probability $\pseFn(\obs, \fluent)$ that $\fluent$ is true.

\textbf{Incorporating prior information.}
The probabilistic grounding step produces per-feature estimates.
However, this does not take into account any prior information already available, e.g., from previous observations or user-provided hints.
The second stage of the pipeline, therefore, combines the grounding output with an optional prior distribution using a principled update rule, such as Bayesian update.
We call the combined result \emph{\gls{factoredBelState}} and formally define it as follows:

\begin{definition}[\Gls{factoredBelState}]
    Let \fluentSet be a set of state features. The \emph{\gls{factoredBelState}} $\factoredBelief: \fluentSet \to [0,1]$ defines a multivariate Bernoulli distribution over individual features.
\end{definition}

When no prior information is available, the \gls{factoredBelState} equals the probabilistic grounding output directly: $\factoredBelief(\fluent) = \pseFn(\obs, \fluent)$ for each observable feature.
When a prior distribution $\factoredBelief_\text{prior}$ is available, the \gls{factoredBelState} can be obtained by combining it with the grounding output, for example via Bayesian update.
For features where $\pseFn(\obs, \fluent) = \textsc{null}$, the \gls{factoredBelState} retains the prior value if one exists, receives a default uninformative value (e.g., $0.5$), or is handled downstream.

\subsection{Extracting task constraints}\label{sec:belief:constraints}
In stage 3 of the pipeline in \cref{fig:belief-pipeline}, we convert the \gls{unconstBelState} into a more plausible \gls{taskBelState}.
We do so by incorporating task-specific constraints that rule out impossible states, thus concentrating probability mass on valid states.
We define constraints on the state space as a function $\constraintSet: 2^{\fluentSet} \to \{0,1\}$, indicating whether a given state satisfies the constraints or not.
Equivalently, we write $\state \models \constraintSet$ to indicate $\constraintSet(\state) = 1$, and $\state \not\models \constraintSet$ to indicate $\constraintSet(\state) = 0$.

To obtain task-level constraints, we need to leverage the structure of the task.
Many symbolic domains impose structural constraints on which states are valid.
Crucially, this validity notion is task-relative rather than perceptual.
The task description determines which symbolic states are admissible and should therefore receive nonzero probability under the downstream uncertainty representation.
In many cases, there are obvious constraints that can be identified by inspection and supplied manually or via a simple heuristic.
For example, in \cref{ex:kitchen}, we can easily identify that an object cannot be in two places at once, and thus cannot occupy both the top and bottom shelves simultaneously.
However, given a formal symbolic task description, we can often extract these constraints automatically.

A principled source of task-level constraints is the \gls{fd} planning system \cite{helmertFastDownwardPlanning2006}.
\Gls{fd} accepts as input \gls{pddl} \cite{mcdermottPDDLthePlanningDomain1998}, a common formalism for describing planning problems that defines the state features, actions and their effects, the initial state, and goal conditions.
\Gls{fd} analyzes the \gls{pddl} description and identifies invariants in the domain.
Importantly, \gls{fd} induces these invariants by determining which states are reachable from the initial state.
However, to obtain useful constraints, this initial state does not have to be the true initial state of the problem, only an ancestor state from which the current state is reachable.

One kind of constraint provided by \gls{fd} is called mutex groups \cite{helmertConciseFinitedomainRepresentations2009}, defined as a set of feature sets $\{\constraint_1, \ldots, \constraint_m\}\subseteq 2^{\fluentSet}$. For each $i$, at most one feature in $\constraint_i$ can be true at a time, i.e., $\constraintSet(\state) = \indicator[\forall i, |\constraint_i \cap \state| \leq 1]$.
An interesting property of disjoint mutex groups, i.e., $\constraint_i \cap \constraint_j = \emptyset$ for all $i \neq j$, is that the cumulative probability of constraint-satisfying states can be computed in linear time relative to the number of features $|\fluentSet|$, in the worst case.
This property will help us correctly calculate state probabilities after applying constraints, as explained in the next section.

\subsection{Belief computation}\label{sec:belief:comp}

\begin{wrapfigure}{l}{0.43\textwidth}
    \vspace{-3\baselineskip}
    \centering
    \includegraphics[width=1\linewidth]{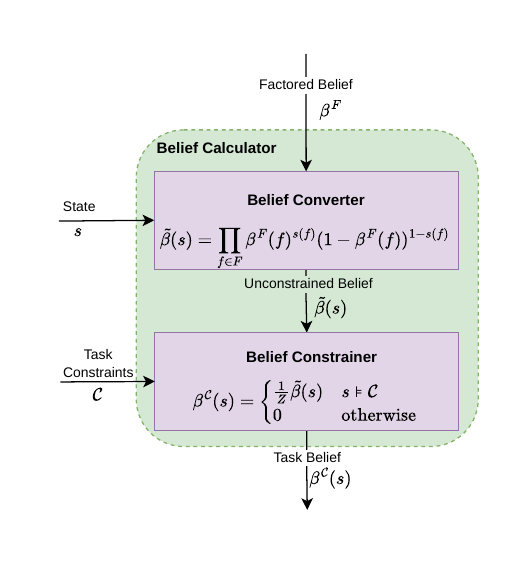}
    \vspace{-3\baselineskip}
    \caption{Belief calculator (from \cref{fig:belief-pipeline}).}
    \label{fig:belief-calc}
    \vspace{-1\baselineskip}
\end{wrapfigure}

The \gls{factoredBelState} provides probabilities for individual features, but downstream reasoning often requires uncertainty over states rather than state features in isolation \cite{palaciosCompilingUncertaintyAway2009,kaelblingPlanningActingPartially1998}.
The \gls{belState} computation stage therefore lifts these marginals to a distribution over symbolic states in $2^\fluentSet$ and then refines that distribution using task structure, without enumerating the full state space.
The \gls{belState} computation is broken down in \cref{fig:belief-calc}, and described in detail below.

\textbf{Belief converter.}
The true joint distribution over all state features is unknown.
That would require modeling the complex dependencies between features, which is often intractable.
Instead, we make a simplifying assumption of conditional independence between features.
This assumption is likely violated in practice, but it allows us to approximate the probability of a single state in linear time, which we can later refine using task-level constraints.
We quantify the sensitivity of the resulting \gls{mbs} to correlated perception errors in \cref{ap:correlated}.
In this setting, we define the \emph{\gls{unconstBelState}} of a state $\state \in 2^{\fluentSet}$ as:
\begin{equation}\label{eq:unconst-belief}
    \unconstBeliefState(\state) = \prod_{\fluent \in \fluentSet} \factoredBelief(\fluent)^{\state(\fluent)} (1 - \factoredBelief(\fluent))^{1 - \state(\fluent)}
\end{equation}

\textbf{Belief constrainer.}
The \gls{unconstBelState} captures the perceptual evidence but ignores task-specific structure.
It assigns a nonzero probability to every state in $2^\fluentSet$, including states that violate the constraints of the domain.
For example, in \cref{ex:kitchen}, an item cannot be both on the counter and in the fridge simultaneously.
To address this, we incorporate task constraints to obtain a \gls{belState} that concentrates probability mass on valid states.

To incorporate given task constraints $\constraintSet: 2^{\fluentSet} \to \{0,1\}$, we define the \emph{\gls{taskBelState}} $\taskBeliefState$ by conditioning on \constraintSet.
States that violate the constraints are assigned zero probability, and the remaining states retain their \gls{unconstBelState}, normalized accordingly:
\begin{equation}\label{eq:calc-belief}
    \taskBeliefState(\state) = \begin{cases}
    \frac{1}{\normalizer} \unconstBeliefState(\state) & \text{if } \state \models \constraintSet \\
    0 & \text{otherwise}
    \end{cases}
\end{equation}
where $\normalizer = \sum_{s\models \constraintSet} \unconstBeliefState(s)$ is a normalizing constant enforcing that \taskBeliefState is a valid probability distribution.
For this to be well-defined, at least one state satisfying $\constraintSet$ must have nonzero \gls{unconstBelState}.

Note that in the general case, computing $\normalizer$ exactly requires summing over all valid states, which may be expensive or intractable.
However, using disjoint mutex groups, \normalizer can be computed in linear time, as shown in \cref{ap:constraints:norm}.
Furthermore, we can use a relaxed normalization approach where $\dnormalizer \geq \normalizer$ to compute a \gls{taskBelState} that is not a true probability distribution, but rather a lower bound on the true constrained probability of each state.
We can always choose $\normalizer_{\text{relaxed}} = 1$, which corresponds to no normalization.
Regardless of the choice of normalizer, the resulting \gls{taskBelState} preserves the relative ordering of states by probability.
The tradeoff, however, is that the \gls{taskBelState} underestimates the true probabilities, leading to more conservative downstream reasoning that may consider more candidate states than strictly necessary.
Consequently, whenever the reported cumulative mass of a set reaches a threshold \probThresh, so does its true constrained mass, so the relaxation costs minimality rather than validity (\cref{cor:relaxed}).

\section{Selecting a compact uncertainty set}\label{sec:mlss}

The \gls{taskBelState} $\taskBeliefState$ assigns coherent probabilities to valid symbolic states, jointly capturing perceptual uncertainty and task structure.
However, $\taskBeliefState$ is still a distribution over a combinatorial state space.
Taking into account the entire state space, whose size is exponential in $|\fluentSet|$, will render downstream reasoning computationally intractable.
On the other hand, collapsing the belief to a single most likely state discards perceptual uncertainty and yields fragile decisions under noisy perception.

We seek a principled middle ground that is the smallest set of valid states whose cumulative probability meets a user-specified belief-mass threshold $\probThresh$, yielding a robust handoff that is compact enough to remain tractable yet faithful to the underlying uncertainty.
We introduce the \emph{\gls{mbs}}, which formalizes this middle ground.

\begin{definition}[\gls{mbs}]\label{def:mlss}
    Let $\beliefState: 2^{\fluentSet} \to [0,1]$ be a \gls{belState}, $\probThresh \in [0,1]$ be a probability threshold, and \constraintSet be a set of constraints on the \gls{symState} space $2^\fluentSet$.  A \emph{\acrfull{mbs}}, denoted $\textup{\gls{mbs}}(\beliefState,\constraintSet)$, is a subset of states $\beliefSet \subseteq 2^\fluentSet$ of minimal cardinality such that $\sum_{\state \in \beliefSet} \beliefState(\state) \geq \probThresh$ and $\forall \state \in \beliefSet, \forall \state' \notin \beliefSet$, it holds that $\beliefState(\state) \geq \beliefState(\state')$.
\end{definition}

\textbf{Attributed mass vs.\ coverage.}
In \cref{def:mlss}, \probThresh bounds the probability mass that the \gls{belState} \beliefState attributes to the returned set.
This mass equals the probability that the true state is contained in the set, which we call \emph{true-state coverage}, only insofar as \beliefState is well calibrated.
We therefore measure coverage directly against ground truth in \cref{sec:experiments}, and report the empirical relation between attributed mass and coverage in \cref{ap:static-full}.

Naively searching for an \gls{mbs} by enumerating all valid states and sorting them according to their probabilities would be highly inefficient.
We propose \gls{pravas}, a novel algorithm described in \cref{alg:mlss}.
\Gls{pravas} conducts a best-first search over symbolic states, where two states are considered neighbors when they differ in the value of exactly one feature.
\gls{pravas} disassembles the \gls{belState} calculator depicted in \cref{fig:belief-calc}, sorting states by their \gls{unconstBelState} and selecting them according to their \gls{taskBelState}.
The search begins from the most likely state, obtained by setting each feature to its most probable value, and progressively expands neighbors in non-increasing order of probability by flipping the state bits (feature values).
At each step, the highest-priority state is extracted from a max-heap.
If it satisfies the task constraints, it is added to the growing \gls{mbs}, and otherwise it is skipped but still expanded so that valid neighbors remain reachable.
The search halts as soon as the cumulative probability of the collected states crosses the threshold \probThresh.
See \cref{ap:ex-prevas} for a step-by-step numerical example describing the search process on a small instance with a guiding illustration (\cref{fig:mlss-example}).

\begin{algorithm}[t]
\caption{\acrfull{pravas}}\label{alg:mlss}
\begin{algorithmic}[1]
    \Require $\factoredBelief: \fluentSet \rightarrow [0,1]$, $\probThresh \in [0, 1]$, $\constraintSet: 2^{\fluentSet} \to \{0,1\}$, $\normalizer \in (0,1]$
    \Ensure $\beliefSet \in \argmin_{\beliefSet' \subseteq 2^\fluentSet}\left\{|\beliefSet'| \middle| \sum_{\state \in \beliefSet'} \taskBeliefState(\state) \geq \probThresh\right\} \text{ s.t. } \forall \state \in \beliefSet, \forall\state' \notin \beliefSet \quad \taskBeliefState(\state) \geq \taskBeliefState(\state')$.
    \State $\state_0(\fluent) \leftarrow \indicator[\factoredBelief(\fluent) > 0.5] \quad \forall \fluent \in \fluentSet$ \label{alg:mlss:init-s0} \Comment{Start with a most likely state}
    \State $\unconstBeliefState(\state_0) \gets \text{CalculateUnconstrainedBelief}(\factoredBelief, \state_0)$ \Comment{Calculated via \cref{eq:unconst-belief}}
    \State $\queue \leftarrow$ Empty max-heap; $\text{Visited} \leftarrow \{\state_0\}$ \label{alg:mlss:init-sets}
    \State $\queue$.Insert($\state_0$, $\unconstBeliefState(\state_0)$) \label{alg:mlss:init-q}
    \State $\beliefSet \leftarrow \emptyset$; $\prob \leftarrow 0$ \label{alg:mlss:init-p}  \Comment{Selected states and their cumulative probability}
    \While{$\prob < \probThresh \wedge \queue \neq \emptyset$} \label{alg:mlss:while}
        \State $\state, \unconstBeliefState(\state) \leftarrow$ $\queue$.ExtractMax() \label{alg:mlss:extract-q}
        \If{$\state \models \constraintSet$} \Comment{Only include constraint-satisfying states}
            \State $\taskBeliefState(\state) \gets \frac{1}{\normalizer} \unconstBeliefState(\state)$ \Comment{Calculated \gls{taskBelState} via \cref{eq:calc-belief}}
            \State $\beliefSet \leftarrow \beliefSet \cup \{\state\}$; $\prob \leftarrow \prob + \taskBeliefState(\state)$ \label{alg:mlss:append-res}
        \EndIf
        \For{$\fluent \in \fluentSet$} \label{alg:mlss:for} %
            \State $\state' \leftarrow$ Flip($\state$, $\fluent$) \Comment{Flip $\fluent$'s value: $\state'(\fluent) = 1 - \state(\fluent)$}
            \If{$\state' \notin \text{Visited}$}
                \State $\text{Visited} \leftarrow \text{Visited} \cup \{\state'\}$ \label{alg:mlss:add-visit}
                \State $\unconstBeliefState(\state') \gets \text{CalculateUnconstrainedBelief}(\factoredBelief, \state')$ \Comment{Calculated via \cref{eq:unconst-belief}}
                \State $\queue$.Insert($\state'$, $\unconstBeliefState(\state')$) \label{alg:mlss:insert-q}
            \EndIf
        \EndFor
    \EndWhile
    \State \Return $\beliefSet$
\end{algorithmic}
\end{algorithm}

In more detail, in \cref{alg:mlss}, lines \ref{alg:mlss:init-s0}--\ref{alg:mlss:init-q} initialize the search at $\state_0$ with a max-heap containing only this state.
Line \ref{alg:mlss:extract-q} extracts the highest-priority state from the heap, based on its \gls{unconstBelState} (computed via \cref{eq:unconst-belief}).
The extracted state is added to the output \gls{belSet} if it satisfies the constraints (line \ref{alg:mlss:append-res}).
The loop at line \ref{alg:mlss:for} expands the extracted state by flipping each feature in turn, enqueuing any unvisited neighbor (line \ref{alg:mlss:insert-q}).
The termination condition (line \ref{alg:mlss:while}) halts the search once the cumulative probability of selected states meets or exceeds \probThresh.

The key insight behind \gls{pravas} is that the heap prioritizes by the \gls{unconstBelState} \unconstBeliefState (computed via \cref{eq:unconst-belief}) rather than the \gls{taskBelState} \taskBeliefState, yet constraint-satisfying states are extracted in non-increasing order of \taskBeliefState as long as the heap is not empty.
This lets us prioritize exploration while guaranteeing that the output is an \gls{mbs} without enumerating the full state space.

\begin{restatable}{theorem}{thmmlss}\label{thm:mlss}
    In \cref{alg:mlss}, when a constraint-satisfying state $\state$ is extracted from the heap $\queue$, then for all $\state'\notin \beliefSet$ it holds that $\taskBeliefState(\state) \geq \taskBeliefState(\state')$, where \taskBeliefState is the \gls{taskBelState} computed by \cref{eq:calc-belief}.
\end{restatable}

\begin{proof}[Sketch]
    \cref{alg:mlss} explores states via a max-heap ordered by the \gls{unconstBelState} \unconstBeliefState.
    The \emph{canonical path} from $\state_0$ to any state $\state$, formed by flipping differing features, is monotonically non-increasing in \unconstBeliefState since each flip moves a feature from its most likely value.
    Consequently, when $\state$ is extracted, any unextracted $\state'$ is either in the heap with $\unconstBeliefState(\state') \leq \unconstBeliefState(\state)$, or has an unextracted ancestor $\state''$ in the heap on its canonical path with $\unconstBeliefState(\state) \geq \unconstBeliefState(\state'') \geq \unconstBeliefState(\state')$.
    Thus, states are extracted in non-increasing order of \unconstBeliefState, meaning constraint-satisfying states are extracted in non-increasing order of their proportional \gls{taskBelState} \taskBeliefState.
\end{proof}

From \cref{thm:mlss}, it follows that \gls{pravas} correctly returns an \gls{mbs} as defined in \cref{def:mlss}.

\begin{restatable}{corollary}{cormlss}\label{cor:mlss}
    The \gls{pravas} algorithm described in \cref{alg:mlss} returns an \gls{mbs} as defined in \cref{def:mlss} relative to the \gls{taskBelState} \taskBeliefState as calculated in \cref{eq:calc-belief}.
\end{restatable}

\cref{thm:mlss,cor:mlss} are exactness guarantees relative to the stated \gls{taskBelState}: \gls{pravas} returns exactly the \gls{mbs} of \taskBeliefState.
They do not guarantee robustness to misspecification of \taskBeliefState, e.g., through violations of the conditional independence assumption of \cref{sec:belief:comp}, whose effect we measure empirically in \cref{sec:experiments,ap:correlated}.

In the worst case, the number of explored states is exponential in $|\fluentSet|$.
Even if we bound the \gls{mbs} size to \beliefSize, there might still be many constraint-violating states that need to be explored before finding \beliefSize valid states.
In the absence of constraints, we can provide a tighter bound on the runtime. With constraints, we hope to encounter a manageable number of constraint-violating states in practice.

\begin{restatable}{proposition}{thmefficient}\label{thm:efficient}
    If $\constraintSet = \emptyset$, \cref{alg:mlss} runs in $O(|\fluentSet|\beliefSize \log(|\fluentSet|\beliefSize))$ time and $O(|\fluentSet|^2\beliefSize)$ bits, where $\beliefSize$ is the number of states in the \gls{mbs}.
\end{restatable}

In summary, the exact-minimality guarantee of \cref{cor:mlss} holds for any constraint set, the polynomial bound of \cref{thm:efficient} is proven for the unconstrained case, and the efficiency of \gls{pravas} in constrained domains is demonstrated empirically in \cref{sec:experiments}.
Full proofs are in \cref{ap:proof-mlss,ap:proof-cor,ap:proof-efficiency}.
With this, the \gls{mbs} can be computed efficiently and used as input to downstream reasoners, e.g., as the initial belief set for planning under initial-state uncertainty.

\section{Experiments}\label{sec:experiments}

The goals of our empirical evaluation are threefold: to show that (1) \gls{mbs} is a good \gls{belState} representation of a pretrained perception model; (2) \gls{pravas} is an efficient method for computing an \gls{mbs}; and (3) \gls{pravas} can be integrated into downstream decision-making to achieve improved robustness.
All experiment code and data are available at \coderepo.

\begin{figure}[t]
\begin{minipage}[t][][b]{0.35\textwidth}
    \centering
    \includegraphics[width=1\linewidth]{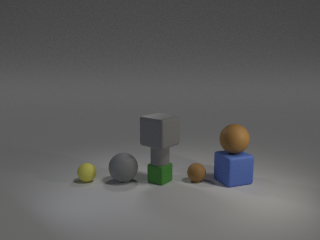}
    \subcaption{PRB}
    \label{fig:domains:prb}
\end{minipage}\hfill
\begin{minipage}[t][][b]{0.35\textwidth}
    \centering
    \includegraphics[width=1\linewidth]{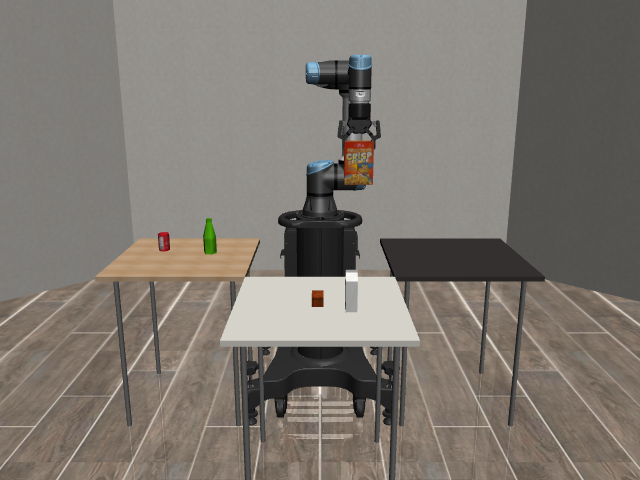}
    \subcaption{GW}
    \label{fig:domains:gw}
\end{minipage}\hfill
\begin{minipage}[t][][b]{0.26\textwidth}
    \centering
    \includegraphics[width=1\linewidth]{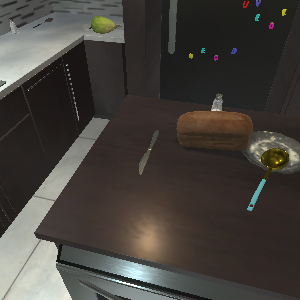}
    \subcaption{ALF}
    \label{fig:domains:alf}
\end{minipage}\hfill
\caption{Example observations from each evaluation domain.}
\label{fig:domains}
\vspace{-1\baselineskip}
\end{figure}

{\bf Domains.}
We evaluate three domains that reflect different aspects of uncertainty.
\textbf{Photorealistic Blocksworld (PRB)} \cite{asaiPhotoRealisticBlocksworldDataset2018} is a blocksworld environment with photorealistic rendering, providing clean symbolic constraints and enabling controlled scaling. It is also visually challenging due to realistic lighting, textures, and occlusions, leading to uncertain state estimation.
\textbf{Groceries World (GW)} \cite{azranS3ESemanticSymbolic2025} is a tabletop manipulation domain with realistic visual ambiguity. A robot must rearrange grocery items on different colored tables.
The symbolic description does not fully capture the uncertain environmental dynamics.
\textbf{ALFWorld (ALF)} \cite{shridharALFWorldAligningText2020}, adapted from the ALFRED benchmark \cite{ALFRED20}, is used as a bounded one-shot stress test for hidden initial-state features, including task-relevant objects that may be hidden in a receptacle, producing genuinely unresolvable uncertainty.
Example observations from each domain are shown in \cref{fig:domains}, while full details, including state features, actions, and objects, are provided in \cref{ap:domains}. Full details on data construction are in \cref{ap:data}.

\textbf{State estimator and calibration.}
The \gls{sse} was implemented using the S3E \cite{azranS3ESemanticSymbolic2025} Python library.
Raw per-feature probability estimates are calibrated via Platt scaling \cite{plattProbabilisticOutputsSupport1999}, which fits a logistic recalibration of the per-feature logit on a held-out validation split. These calibrated marginals are shared across all baselines, ensuring that differences in output quality reflect only the state-set construction method, not the perception model.
All experiments presented in this section use an \gls{sse} based on a single fixed \gls{vlm} backbone and prompting scheme, with GPT-4.1 \cite{openaiIntroducingGPT41API2025} serving as the underlying model.
\cref{ap:backbones} repeats Experiments 1 and 3 with five open-weight \glspl{vlm} under the same pipeline and calibration protocol.

\textbf{Constraint extraction.}
For PRB and GW, task constraints are extracted from the \gls{pddl} domain and problem files using the \gls{fd} translator \cite{helmertFastDownwardPlanning2006,helmertConciseFinitedomainRepresentations2009}, which discovers mutex groups and structural invariants.
For ALF, \gls{fd} was not applicable due to the problem structure, so we used a domain-specific heuristic to extract key structural invariants, e.g., the agent can only hold one object at a time. 
\cref{ap:constraints} details the process of extracting constraints and computing the normalizer for each domain.

\textbf{Experiment 1: ``How steep is the tradeoff between true-state coverage and \gls{belSet} compactness for \gls{mbs} as controlled by \probThresh, and how does it compare to other belief set constructions?''}

We compare \gls{mbs} to three baselines. \textbf{\gls{map}}: the single most likely state, i.e., the state with each feature set to its most likely value, yielding a singleton \gls{belSet}.
\textbf{Feature Abstention}: the set of states obtained by abstaining on all features whose marginal probability is below a threshold \absThresh \cite{el-yanivFoundationsNoisefreeSelective2010}.
Features with a marginal above \absThresh are determinized to their most likely value, and the rest are treated as ambiguous, i.e., both assignments are included in the returned set.
For each \probThresh, we find the smallest \absThresh such that the returned set exceeds the \probThresh mass on average.
\textbf{Random \probThresh-Mass}: valid states are randomly selected until their cumulative \gls{taskBelState} exceeds \probThresh. All methods accept as input the same calibrated marginals from the \gls{sse} and the resulting probabilistic grounding (the output from stage 1 in \cref{fig:belief-pipeline}).
Features assigned \textsc{null} by the \gls{sse} are treated as ambiguous, i.e., both assignments are included in the \gls{belSet}.
\gls{map} is a point-estimate method, and thus ignores \textsc{null} assignments and forces a truth assignment for each feature.

\begin{figure}[t]
        \centering
        \includegraphics[width=1\linewidth]{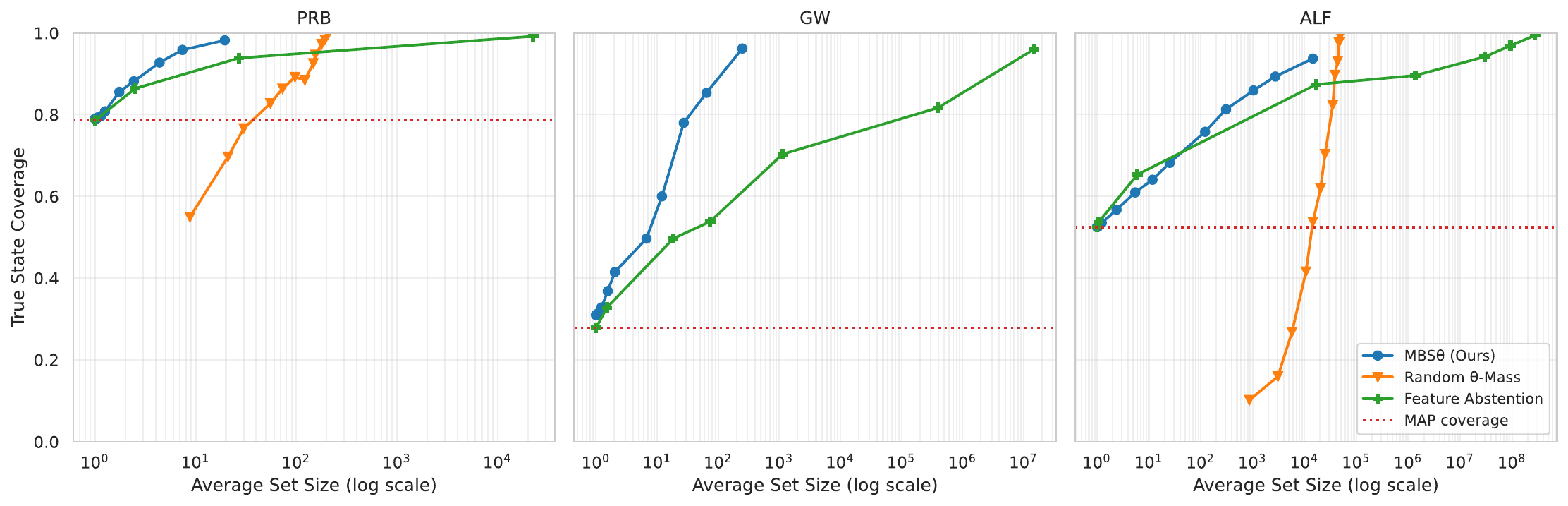}
        \caption{True-state coverage vs.\ average output set size. %
        Top left is better.}
        \label{fig:coverage-size}
    \vspace{-2\baselineskip}
\end{figure}

\textbf{Results.}
As the \gls{belSet} grows, the true-state coverage generally increases, but at the cost of degraded robustness and heavier computation.
Therefore, we want coverage to increase more for little growth in \gls{belSet} size.
\cref{fig:coverage-size} illustrates this coverage-compactness tradeoff.
Each curve corresponds to a method.
Each point on the curve corresponds to a different \probThresh (or \absThresh calibrated for \probThresh), which affects the size of the \gls{belSet}.
\Gls{map} is a singleton, and thus does not grow in size, but its coverage is a lower bound on that of any \gls{belSet} that includes the \gls{map}, including \gls{mbs}.
Feature abstention erratically jumps in set size for small changes in coverage, since slightly changing the abstention threshold \absThresh can vastly increase the number of abstained features.
In contrast, the \gls{mbs} curve is smooth and generally dominates feature abstention, indicating that \gls{mbs} achieves a better coverage-compactness tradeoff.
Note that in ALF, there is a small window where feature abstention dominates \gls{mbs} for low \probThresh values, but this is because the feature-abstention subset is not minimal (see \cref{ap:static-full}).

In the GW domain, Random \probThresh-Mass is omitted because it fails to finish running on 90\% of instances within its timeout window.
In the PRB and ALF domains, it has the steepest coverage-compactness curve, but at the cost of extremely large set sizes.
For example, at $\probThresh=0.9$, the set size of Random \probThresh-Mass is over 40 times that of \gls{mbs} in both PRB and ALF.
This is even more pronounced at lower \probThresh values.
Moreover, unlike the other methods, Random \probThresh-Mass does not necessarily contain the most likely state, and thus its coverage can be lower than that of \gls{map} despite its large size.
A complete table with coverage and set size values is provided in \cref{ap:static-full}.
That appendix also plots realized set mass against empirical true-state coverage, which agree within 0.11 for GPT-4.1 at \probThresh between 0.9 and 0.99, and \cref{ap:backbones} shows that \gls{mbs} retains its coverage-compactness advantage on all six \gls{vlm} backbones.

\textbf{Experiment 2: ``Is \gls{pravas} an efficient method for constructing an \gls{mbs} compared to naive approaches, and how does it scale with problem size?''}

We compare \gls{pravas} to an exhaustive approach that computes the \gls{mbs} by enumerating all $2^{|\fluentSet|}$ states, computing their \gls{taskBelState} \taskBeliefState, sorting them, and finding the minimal prefix that crosses the \probThresh mass.
Each method is given a 10\,s window for each test instance to compute the \gls{mbs}.
We measure how long each method takes to compute the \gls{mbs}, and how many states it maintains during the search (which corresponds to memory usage). Hardware configurations are detailed in \cref{ap:compute}.
Here, we want to compute an \gls{mbs}, and thus do not compare to the baselines from other experiments.

\begin{figure}[!b]
    \begin{minipage}[t][][b]{0.32\textwidth}
        \centering
        \includegraphics[width=1\linewidth]{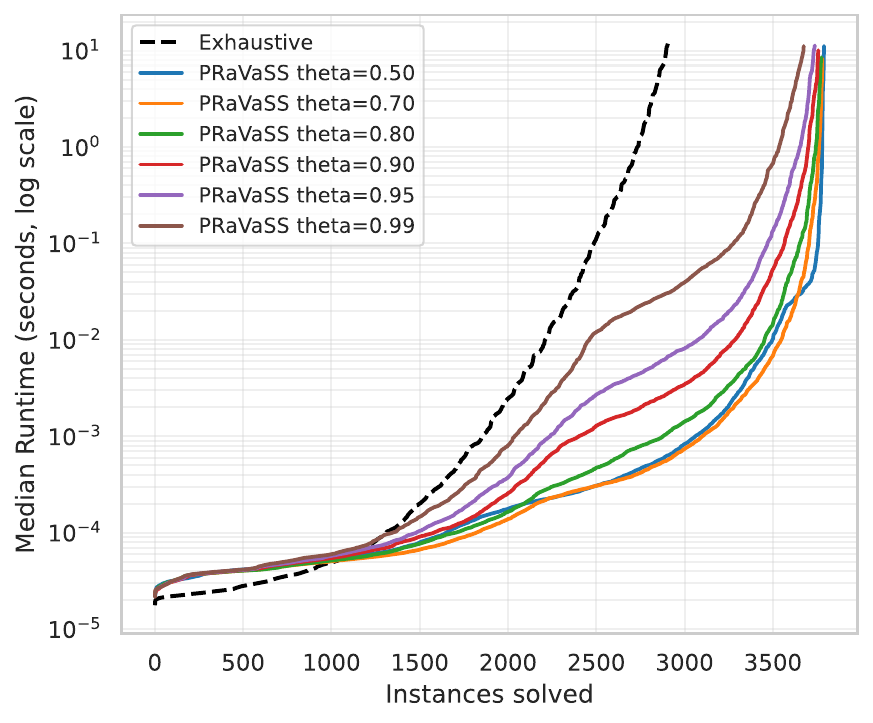}
        \caption{Runtime vs.\ instances solved.}
        \label{fig:scaling:cactus}
    \end{minipage}\hfill
    \begin{minipage}[t][][b]{0.33\textwidth}
        \centering
        \includegraphics[width=1\linewidth]{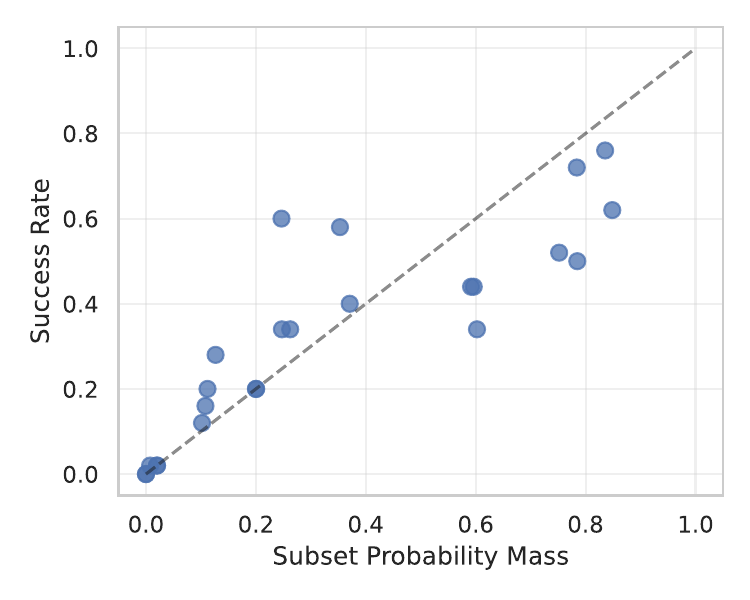}
        \caption{Average attributed mass vs.\ plan success rate for \acrfull{ams}.}
        \label{fig:success-conf-cor}
    \end{minipage}
    \hfill
    \begin{minipage}[t][][b]{0.32\textwidth}
        \centering
        \captionof{table}{Downstream planning success rate.}
        \label{tab:planning}
    \resizebox{\columnwidth}{!}{
\begin{tabular}{@{}lcc@{}}
\toprule
Method & PRB & GW \\
\midrule
MAP & 16.7\% & 70.7\% \\
 & {\scriptsize [14.4\%, 18.8\%]} & {\scriptsize [67.2\%, 73.8\%]} \\
\addlinespace[0.2em]
Adaptive Abst. & \underline{16.8\%} & \underline{73.3\%} \\
 & {\scriptsize [14.7\%, 19.1\%]} & {\scriptsize [69.8\%, 76.8\%]} \\
\addlinespace[0.2em]
\midrule
AMS (Ours) & \textbf{32.6\%} & \textbf{80.0\%} \\
 & {\scriptsize [30.1\%, 35.3\%]} & {\scriptsize [76.7\%, 83.0\%]} \\
\bottomrule
\end{tabular}
}

    \end{minipage}\hfill
\end{figure}

\textbf{Results.}
\cref{fig:scaling:cactus} shows a cactus plot of the runtime of \gls{pravas} and exhaustive enumeration.
A point $(x,y)$ on the cactus plot indicates that $x$ instances were solved by the method within $y$ seconds.
The plot shows the dominance of \gls{pravas} even at the highest tested threshold $\probThresh=0.99$, which forgoes only 1\% of the probability mass.
This shows that \gls{pravas} efficiently drops a large mass of low-probability states that waste precious computation time and memory. See \cref{ap:runtime-full} for a scaling analysis based on problem size, as well as a complete table of runtime and memory values across domains and \probThresh values. There, we show that \gls{pravas} can scale to problems with 42\% more features than exhaustive enumeration at $\probThresh=0.99$, and much more at lower \probThresh.
Although all three domains carry constraints, the median number of states visited remains below 160 at every \probThresh (\cref{tab:scaling}), so the exponential worst case discussed in \cref{sec:mlss} is not met in practice.

\textbf{Experiment 3: ``Can \gls{pravas} be integrated into downstream decision-making tasks and achieve improved robustness compared to point estimates?''}

To demonstrate downstream effects, we use \gls{pravas} to construct an \gls{mbs}, and use a robust planner \cite{palaciosCompilingUncertaintyAway2009,smithConformantGraphplan1998} that finds plans that achieve the goal from every state in a given \gls{belSet}.
The ALF domain is excluded from this experiment because its partially observable nature requires the agent to actively collect additional information to ensure robust decision-making. %
Building on its efficiency, we can use \gls{pravas} iteratively to find the largest \probThresh for which a robust plan exists, an approach we call \gls{ams}.
\Gls{ams} provides the attributed mass of the selected \gls{belSet} as a confidence score for plan success instead of relying on a user-specified threshold.
We compare this approach to \gls{map}, where we plan from the single most likely state, and an adaptive version of feature abstention that finds the largest \absThresh for which a plan is found.
We use the CPOR planner \cite{maliahComputingContingentPlan2022} with an underlying \gls{fd} planner \cite{helmertFastDownwardPlanning2006}.
\Gls{ams} uses the full set of constraints extracted by \gls{fd}, whose mutex groups overlap, so the exact normalizer requires weighted model counting (\cref{ap:constraints}).
We therefore use the relaxed normalizer of \cref{sec:belief:comp}, which preserves the ordering of states, the only quantity the search consumes, while reported masses remain lower bounds.
A closed-loop observe-update-replan study on ALF, in which \gls{ams} stays above \gls{map} by 20.9 points before and 16.5 points after a single observation, is reported in \cref{ap:closed-loop}.
For full details, see \cref{ap:ex3-full}.

\textbf{Results.}
\cref{tab:planning} summarizes the downstream planning results.
Bracketed rows are 95\% bootstrap confidence intervals (CI) using 10,000 paired resamples over test instances, where success is measured by whether the found plan reaches the goal.
Bold values indicate the best-performing method in each domain, and underlined values the second-best.
\gls{ams} achieves a 13\% improvement over \gls{map} in GW, and a 95\% improvement in PRB.
This shows the power of constraint-based belief set construction and the versatility of \gls{pravas}, even in the presence of approximated probability masses obtained with a relaxed normalizer.
Two ablations in \cref{ap:ex3-full} separate the belief-set construction from the constraint information.
With identical calibrated marginals and identical full constraints, \gls{ams} exceeds adaptive abstention by 15.7 points on PRB (paired 95\% CI $[13.6, 17.8]$) and by 5.2 points on GW ($[3.2, 7.3]$), and exceeds its own loose-constraint variant by 16.2 and 4.5 points, respectively, whereas replacing the relaxed normalizer with the exact one changes success by at most 0.7 points.
The smaller GW gain is largely a headroom effect, since \gls{map} already succeeds on over 70\% of GW instances against a 95.0\% oracle ceiling, and the gain grows on backbones with degraded yet informative perception (\cref{ap:backbones}).

Running \gls{pravas} iteratively enables \gls{ams} to communicate its confidence in a plan through the probability mass of the \gls{belSet}. \cref{fig:success-conf-cor} scatters the average success rate as a function of \gls{ams}'s attributed mass across test instances in the PRB domain.
Each point on the plot represents the success rate of \gls{ams} for a given \probThresh on all PRB instances of a certain horizon and object count configuration, corresponding to different levels of problem difficulty. The results show a strong positive correlation between attributed mass and success rate.
The Pearson correlation is $r=0.866$, with $p\approx4\times10^{-8}$, indicating that the attributed mass is a strong predictor of plan success.
\cref{ap:ex3-full} contains additional results for this experiment, including results showing that planning over \gls{mbs} can retain its performance when \gls{sse} classification is wrong, highlighting how \gls{mbs} enhances robustness.

\glsresetall
\section{Conclusion}

By translating the probabilistic outputs of foundation models into rigorous symbolic representations and providing a mathematically grounded handoff from uncertain perception to symbolic planning, our framework offers a practical stepping stone toward more reliable neurosymbolic  AI systems. %
Rather than forcing a choice between brittle single-state determinization and intractable full belief-space planning, our \gls{mbs}, extracted via the \gls{pravas} algorithm, yields an uncertainty abstraction permissive enough to respect perceptual ambiguity, structured enough to obey task logic, and compact enough for tractable downstream decision-making.
Crucially, it achieves this with formal correctness guarantees relative to the stated belief and polynomial complexity in the unconstrained case, significantly outperforming baselines in coverage and compactness, while scaling to problems with up to 154\% more state features than exhaustive approaches when capturing the majority of the state space probability mass.

\textbf{Limitations and future work.} We note that the quality of \gls{mbs} is dependent on the accuracy of the \gls{sse}, and that \probThresh bounds the mass attributed by a belief that assumes conditional independence between features, rather than certified true-state coverage. Furthermore, our belief system is static and does not take exploration into account. Future extensions will support distribution-free calibration of \probThresh against true-state coverage, temporal belief refinement, learned inter-feature dependencies, and active information-gathering actions, ultimately enabling closed-loop robotic agents that reason safely through an ambiguous world.

\textbf{Broader impact.} For a detailed analysis of societal impacts, see \cref{ap:impact}.

\begin{ack}
This work received no third-party funding or support.
The authors declare no competing interests.
We thank the anonymous reviewers and area chair, whose comments led directly to the ablation, multi-backbone, sensitivity, and closed-loop studies added in this version.
\end{ack}

\bibliography{references}
\bibliographystyle{unsrt}

\newpage

\appendix
\crefalias{section}{appendix}
\crefalias{subsection}{appendix}

\section{\texorpdfstring{\Gls{pravas}}{PRaVaSS} numerical example}\label{ap:ex-prevas}

\begin{figure}[!h]
    \centering
    \includegraphics[width=1\textwidth]{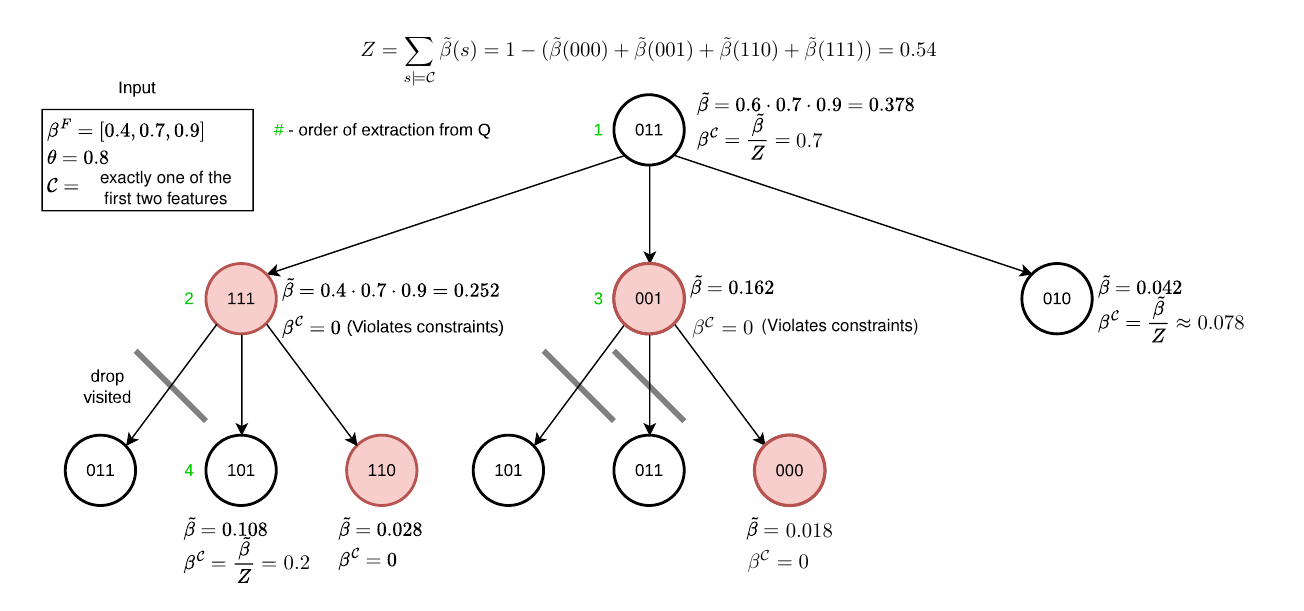}
    \caption{Search for an \gls{mbs} on three features using the \gls{pravas} algorithm, expanding from state 011 via bit flips. Indices appearing to the left of certain nodes indicate the order in which they are extracted from the max-heap, based on the \gls{unconstBelState} $\unconstBeliefState$. Red-shaded nodes violate constraints and are excluded from the \gls{mbs}.}
    \label{fig:mlss-example}
\end{figure}

\cref{fig:mlss-example} shows a numerical example of how the state space is explored in \cref{alg:mlss} using bit-flip operations.
The example consists of a three-feature state space.
The \gls{factoredBelState} is given as $\factoredBelief = [0.4, 0.7, 0.9]$ for the first, second, and third features, respectively.
The desired probability threshold is set to $\probThresh = 0.8$.
The constraints are that the first and second features are mutually exclusive and exhaustive, i.e., exactly one of them is true in any valid state.
The \gls{belState} normalizer for \cref{eq:calc-belief} is
$$
\normalizer = 1 - (\unconstBeliefState(000) + \unconstBeliefState(001) + \unconstBeliefState(110) + \unconstBeliefState(111)) = 1 - (0.018 + 0.162 + 0.028 + 0.252) = 0.54
$$

The max-heap is initialized with the state 011, which is the most probable state.
This is the first state extracted from the max-heap and added to the \gls{mbs}.
It is expanded via bit flips to states 111, 001, and 010, which are added to the heap.
States 111 and 001 violate the constraints, and are assigned zero probability in the \gls{taskBelState}.
The other \glspl{taskBelState} are computed by normalizing their \glspl{unconstBelState}.
After this iteration, $\beliefSet = \{011\}$, and the cumulative probability of states in this set is $\prob = \taskBeliefState(011) = 0.7$.

State 111 has the highest \gls{unconstBelState} among the three nodes currently in the heap, and is thus extracted next.
However, since it violates the constraints, it is not added to the \gls{mbs}.
Expanding 111 reveals states 101, 110, and the already visited 011, so states 101 and 110 are added to the heap.
After this iteration, we still have $\beliefSet = \{011\}$.

Next, state 001 is extracted from the heap, but it also violates the constraints and is not added to the \gls{mbs}.
Expanding 001 reveals the new state 000.
After this iteration, we still have $\beliefSet = \{011\}$.

Finally, state 101 is extracted from the heap and added to the \gls{mbs}, increasing the cumulative probability to $\prob = \taskBeliefState(011) + \taskBeliefState(101) = 0.7 + 0.2 = 0.9$.
This crosses the threshold $\probThresh = 0.8$, and the algorithm terminates with $\beliefSet = \{011, 101\}$ as the resulting \gls{mbs}.

Had we used the \gls{taskBelState} as the heap key instead of the \gls{unconstBelState}, state 010 would have been extracted before 111 and 001, so it would have been added to the \gls{mbs} before 101 was discovered.
This would have resulted in a cumulative probability of approximately $0.778$, just below the threshold, so the search would have continued and added a third state.
However, $\{011, 101\}$ reaches the threshold with only two states, so the three-state output would not have been an \gls{mbs} by definition.

\section{Proof of \texorpdfstring{\cref{thm:mlss}}{\ref*{thm:mlss}}}\label{ap:proof-mlss}

\begin{definition}[Bit diffset]
    Let $\state,\state' \in 2^\fluentSet$. The \emph{bit diffset} $D(\state, \state')$ is defined as the set of features that differ between \state and $\state'$, i.e., 
    $$
    D(\state, \state') = \{\fluent \in \fluentSet | \state(\fluent) \neq \state'( \fluent)\}
    $$
\end{definition}

\begin{definition}[Bit-flip operator]\label{def:bit-flip}
    Let $\state \in 2^\fluentSet$ and $\fluent \in \fluentSet$. The \emph{bit-flip operator} $\text{Flip}(\state, \fluent)$ returns a new state $\state' \in 2^\fluentSet$ such that:
    $$
    \forall \fluent' \in \fluentSet \quad \state'(\fluent') = 
    \begin{cases}
        1 - \state(\fluent) & \text{if } \fluent' = \fluent \\
        \state(\fluent') & \text{otherwise}
    \end{cases}
    $$
\end{definition}

\begin{definition}[Canonical path]\label{def:canon-path}
    Let $\state,\state' \in 2^\fluentSet$. Let $D(\state, \state') = \{\fluent_1, \ldots, \fluent_n\}$ where the features are numbered according to a fixed total ordering over \fluentSet.
    The \emph{canonical path} from \state to $\state'$ is the path of states $\state = \state_0 \rightarrow \state_1 \rightarrow \cdots \rightarrow \state_n = \state'$ achieved by flipping the bits of state $\state_0$ one by one in the order of the features.
    That is, for $i \in \{1, \ldots, n\}$, define $\state_i = \text{Flip}(\state_{i - 1}, f_i)$, from $\state_0 = \state$ leading to $\state_n = \state'$.
\end{definition}

\begin{lemma}\label{le:res-vis}
    In \cref{alg:mlss}, let \queueExtracted denote the set of states that have been extracted from the heap \queue thus far. Then at the moment of max-extraction from \queue, it holds that $\queue \cup \queueExtracted = \text{Visited}$.
\end{lemma}

\begin{proof}
Note that every insertion to the heap is accompanied by an insertion of the same state to the visited set.
All states that have ever been added to the heap are either still in the heap or have been extracted from it.
Since no state is ever removed from the visited set, the lemma holds.
\end{proof}

\begin{lemma}\label{le:path-mono}
    In \cref{alg:mlss}, the \gls{unconstBelState} values of the states on the canonical path from $\state_0$ to any state \state are monotonically non-increasing, i.e.,
    $$
    \unconstBeliefState(\state_0) \geq \cdots \geq \unconstBeliefState(\state_n) = \unconstBeliefState(\state)
    $$
\end{lemma}

\begin{proof}
    Assume by contradiction that there exists $0 < j \leq n$ such that $0 \leq \unconstBeliefState(\state_{j - 1}) < \unconstBeliefState(\state_j)$.
    Then all of the components that comprise the product in $\unconstBeliefState(\state_j)$ are nonzero.
    Thus, we can safely divide by
    $$
    \prod_{\fluent \in \fluentSet \setminus \{\fluent_j\}} \factoredBelief(\fluent)^{\state_{j - 1}(\fluent)} (1 - \factoredBelief(\fluent))^{1 - \state_{j - 1}(\fluent)}
    $$
    From here, we have:
    \begin{align*}
        & \unconstBeliefState(\state_{j - 1}) < \unconstBeliefState(\state_j) \\
        \overset{1}{\iff} & \prod_{\fluent \in \fluentSet} \factoredBelief(\fluent)^{\state_{j - 1}(\fluent)} (1 - \factoredBelief(\fluent))^{1 - \state_{j - 1}(\fluent)} < \prod_{\fluent \in \fluentSet} \factoredBelief(\fluent)^{\state_{j}(\fluent)} (1 - \factoredBelief(\fluent))^{1 - \state_{j}(\fluent)} \\
        \overset{2}{\iff} & \factoredBelief(\fluent_j)^{\state_{j - 1}(\fluent_j)} (1 - \factoredBelief(\fluent_j))^{1 - \state_{j - 1}(\fluent_j)} < \factoredBelief(\fluent_j)^{\state_{j}(\fluent_j)} (1 - \factoredBelief(\fluent_j))^{1 - \state_{j}(\fluent_j)} \\
        \overset{3}{\iff} & \factoredBelief(\fluent_j)^{\state_{j-1}(\fluent_j)} (1 - \factoredBelief(\fluent_j))^{1 - \state_{j-1}(\fluent_j)} < \factoredBelief(\fluent_j)^{1 - \state_{j-1}(\fluent_j)} (1 - \factoredBelief(\fluent_j))^{\state_{j-1}(\fluent_j)}
    \end{align*}
    Transition 1 follows from the definition of \unconstBeliefState in \cref{eq:unconst-belief}. 
    Transition 2 is a division by all components in the product except for the one corresponding to $\fluent_j$, which are nonzero, and are equal since $\state_{j-1}(\fluent) = \state_j(\fluent)$ for all $\fluent \neq \fluent_j$ by the definition of the canonical path in \cref{def:canon-path}.
    Transition 3 follows from the definition of the bit-flip operation in \cref{def:bit-flip}, which states that $\state_j(\fluent_j) = 1 - \state_{j - 1}(\fluent_j)$.

    If $\state_{j-1}(\fluent_j) = 1$, then the above simplifies to
    \begin{align*}
    & \factoredBelief(\fluent_j)^1 (1 - \factoredBelief(\fluent_j))^0 < \factoredBelief(\fluent_j)^0 (1 - \factoredBelief(\fluent_j))^1 \\
    & \Rightarrow \factoredBelief(\fluent_j) < 0.5
    \end{align*}
    If $\state_{j-1}(\fluent_j) = 0$, then the above simplifies to
    \begin{align*}
    & \factoredBelief(\fluent_j)^0 (1 - \factoredBelief(\fluent_j))^1 < \factoredBelief(\fluent_j)^1 (1 - \factoredBelief(\fluent_j))^0 \\
    & \Rightarrow \factoredBelief(\fluent_j) > 0.5
    \end{align*}
    Since $\state_{j-1}(\fluent_j) = \state_0(\fluent_j)$, then both cases contradict the definition of $\state_0$, which for all features is defined as $\state_0(\fluent) = \indicator[\factoredBelief(\fluent) > 0.5]$.
\end{proof}

\begin{lemma}\label{le:order}
    In \cref{alg:mlss}, when a state $\state \in 2^\fluentSet$ is extracted from the max-heap \queue, then for all $\state' \in 2^\fluentSet$ not yet extracted from \queue, it holds that $\unconstBeliefState(\state) \geq \unconstBeliefState(\state')$.
\end{lemma}

\begin{proof}
Let $\queueExtracted_{\beliefSize} = \{\state_0, \ldots, \state_{\beliefSize - 1}\}$ be the set of the first $\beliefSize$ states extracted from \queue, numbered by the order in which they were extracted.
We must show that for all \beliefSize, for all states $\state \in \queueExtracted_{\beliefSize}$, and for all $\state' \notin \queueExtracted_{\beliefSize}$, it holds that $\unconstBeliefState(\state) \geq \unconstBeliefState(\state')$.
We do this by induction on $\beliefSize$.

\paragraph{Basis:} When $\beliefSize = 1$, this is the first iteration, where the only value in the heap is $\state_0$. By definition, for any $\state \neq \state_0$, it holds that $\unconstBeliefState(\state_0) \geq \unconstBeliefState(\state)$.

\paragraph{Assumption:} Assume that for $\beliefSize$ it holds that for all $\state \in \queueExtracted_{\beliefSize}$ and $\state' \notin \queueExtracted_{\beliefSize}$ we have $\unconstBeliefState(\state) \geq \unconstBeliefState(\state')$.

\paragraph{Induction step:} Let $\queueExtracted_{\beliefSize + 1} = \{\state_0, \ldots, \state_{\beliefSize}\}$.
By the induction assumption, $\queueExtracted_{\beliefSize} = \queueExtracted_{\beliefSize + 1} \setminus \{\state_{\beliefSize}\}$ contains the top $\beliefSize$ most probable states, i.e., for all $\state \in \queueExtracted_{\beliefSize}$ and for all $\state' \notin \queueExtracted_{\beliefSize}$, it holds that $\unconstBeliefState(\state) \geq \unconstBeliefState(\state')$.
It is left to show that for all $\state \notin \queueExtracted_{\beliefSize + 1}$, it holds that $\unconstBeliefState(\state_{\beliefSize}) \geq \unconstBeliefState(\state)$.

Assume by contradiction that there exists a state $\state \notin \queueExtracted_{\beliefSize + 1}$ such that $ \unconstBeliefState(\state) > \unconstBeliefState(\state_{\beliefSize})$.
Then $\state \notin \queue$ because otherwise it would have been extracted before $\state_{\beliefSize}$.
By \cref{le:res-vis}, $\state$ has not yet been visited.

On the canonical path $\state_0 = u_0 \rightarrow \cdots \rightarrow u_n = \state$, since $u_0$ has been visited and $u_n = \state$ has not yet been visited, there exists $0<j\leq n$ such that $u_j$ has not yet been visited and $u_{j-1}$ has.
\begin{itemize}
    \item Since $u_{j-1}$ is in the visited set, then by \cref{le:res-vis}, it is either in the heap or has been extracted from the heap. It cannot have been extracted from the heap because then it would have been expanded, and $u_j$ would have been visited. Therefore, $u_{j-1}$ is in the heap.
    \item Since $\state_\beliefSize$ was extracted but $u_{j-1}$ was not, it follows that $\unconstBeliefState(\state_\beliefSize) \geq \unconstBeliefState(u_{j - 1})$.
    \item By \cref{le:path-mono}, $\unconstBeliefState(u_{j - 1}) \geq \unconstBeliefState(u_n) = \unconstBeliefState(\state)$.
\end{itemize}

Putting it all together, we get:
$$
\unconstBeliefState(\state_\beliefSize) \geq \unconstBeliefState(u_{j - 1}) \geq \unconstBeliefState(\state) > \unconstBeliefState(\state_\beliefSize)
$$
which is a contradiction.
Thus, for all $\state \notin \queueExtracted_{\beliefSize + 1}$, it holds that $\unconstBeliefState(\state_{\beliefSize}) \geq \unconstBeliefState(\state)$.
\end{proof}

\thmmlss*

\begin{proof}
Let $\state$ be a constraint-satisfying state extracted from the heap at some iteration.
We must show that for all $\state' \notin \beliefSet$, it holds that $\taskBeliefState(\state) \geq \taskBeliefState(\state')$.

If $\state'$ violates the constraints, then trivially $\taskBeliefState(\state') = 0 \leq \taskBeliefState(\state)$. Otherwise, note that $\beliefSet$ is comprised of all constraint-satisfying states that were extracted from the heap before \state. Therefore, $\state'$ has not yet been extracted from the heap. By \cref{le:order}, we have:
$$
\unconstBeliefState(\state) \geq \unconstBeliefState(\state') \iff \frac{\unconstBeliefState(\state)}{\normalizer} \geq \frac{\unconstBeliefState(\state')}{\normalizer} \iff \taskBeliefState(\state) \geq \taskBeliefState(\state')
$$
Thus, the theorem holds for all $\state' \notin \beliefSet$.
\end{proof}

\section{Proof of \texorpdfstring{\cref{cor:mlss}}{\ref*{cor:mlss}}}\label{ap:proof-cor}

\cormlss*

\begin{proof}
    By \cref{thm:mlss}, each state added to $\beliefSet$ has $\taskBeliefState$ at least as large as that of any state not yet in $\beliefSet$, so $\beliefSet$ contains the $|\beliefSet|$ states with the largest \taskBeliefState.
    The algorithm halts the moment the cumulative probability first reaches \probThresh, which means that before the last addition, the cumulative probability was strictly below \probThresh.
    Since no other set of $|\beliefSet|-1$ states can have larger cumulative probability than the top $|\beliefSet|-1$ already in $\beliefSet$, no smaller set crosses the threshold, so $\beliefSet$ has minimal cardinality and satisfies the ordering condition of \cref{def:mlss}.
\end{proof}

\section{Proof of \texorpdfstring{\cref{thm:efficient}}{\ref*{thm:efficient}}}\label{ap:proof-efficiency}

\thmefficient*

\subsection{Time complexity}

\begin{claim}
The time complexity of \cref{alg:mlss} is $O(|\fluentSet|\beliefSize\log(|\fluentSet|\beliefSize))$.
\end{claim}

\begin{proof}
    \textbf{Pre-iteration cost.}
In lines \ref{alg:mlss:init-s0}--\ref{alg:mlss:init-p} of the algorithm, we linearly compute $\state_0$ and its probability, and initialize collections with at most one item.
This requires $O(|\fluentSet|)$ operations.

\textbf{Inner-iteration cost.}
The internal \textit{for} loop in line \ref{alg:mlss:for} consists of a bit flip, a ``visited'' check, an insertion to the visited set, and an insertion to the heap. Using random access, the bit flip is $O(1)$. Checking membership in the visited set is $O(1)$. Inserting into the visited set is $O(\log(M))$ where $M$ is the size of the set. The heap insertion is $O(\log(N))$ where $N$ is the size of the heap. Assuming none of the states are visited (worst case), then we require $O(|\fluentSet|(\log(N) + \log(M)))$ for all iterations. We must also acknowledge the time to compute the probability at each iteration. If naively implemented, this would take $O(|\fluentSet|)$, meaning throughout the iteration, this would take $O(|\fluentSet|^2)$. However, by updating the probability incrementally according to only the changed bit (divide by unflipped value and multiply by flipped value), this operation takes $O(1)$.

\textbf{Outer-iteration cost.}
The \textit{while} loop in line \ref{alg:mlss:while} consists of a max extraction from the heap, an insertion of this value to \beliefSet, and the internal \textit{for} loop. Since \beliefSet has size at most $\beliefSize$, the insertion can be implemented in $O(\log(\beliefSize))$. The max extraction from the heap requires $O(\log(N))$ where $N$ is the size of the heap.  The \textit{while} conditions are checkable in $O(1)$. The inner \textit{for} loop is $O(|\fluentSet|(\log(N) + \log(M)))$. Considering that at most $|\fluentSet|$ values are inserted into the visited set and heap at every iteration, then $M = N = |\fluentSet|\beliefSize$ at most. Overall, each iteration in the \textit{while} loop requires:
\begin{align*}
    & O(\log(\beliefSize)) + O(\log(N)) + O(|\fluentSet|(\log(N) + \log(M))) \\
    = & O(\log(\beliefSize)) + O(|\fluentSet|(\log(|\fluentSet|\beliefSize) + \log(|\fluentSet|\beliefSize))) \\
    = & O(|\fluentSet|\log(|\fluentSet|\beliefSize)).
\end{align*}
There are at most $\beliefSize$ iterations, so the overall time complexity of the \textit{while} loop is $O(|\fluentSet|\beliefSize\log(|\fluentSet|\beliefSize))$.

Since the pre-iteration costs are negligible compared to the iteration cost, the final upper bound on the time complexity stands at $T(\fluentSet, \beliefSize) = O(|\fluentSet|\beliefSize\log(|\fluentSet|\beliefSize))$.
\end{proof}

\subsection{Space complexity}

\begin{claim}
    The space complexity of \cref{alg:mlss} is $O(|\fluentSet|^2\beliefSize)$ bits.
\end{claim}
\begin{proof}
    As seen in the proof of time complexity, we store the following values at any one time:
\begin{itemize}
    \item The visited set, containing $O(|\fluentSet|\beliefSize)$ states. 
    \item The heap, containing $O(|\fluentSet|\beliefSize)$ states.
\end{itemize}
Each state can be represented using a bitmask of $|\fluentSet|$ bits. Therefore $S(\fluentSet, \beliefSize) = O(|\fluentSet|^2\beliefSize)$ bits.
\end{proof}

\section{Evaluation domains}\label{ap:domains}

This appendix gives a high-level overview of the three evaluation domains used in our experiments: Photorealistic Blocksworld (PRB), Groceries World (GW), and ALFWorld (ALF).
All of these domains are available under permissive open-source licenses (Apache 2.0 and MIT).
The complete \gls{pddl} domain files for all three are included in our code repository (\coderepo), which we recommend consulting for the precise predicate signatures, action preconditions, and effects.
Data collection from these domains is described in \cref{ap:data}, and constraint extraction and normalizer computation are described in \cref{ap:constraints}.

\subsection{Photorealistic Blocksworld (PRB)}\label{ap:domains:prb}

\textbf{PRB} \cite{asaiPhotoRealisticBlocksworldDataset2018} is a shape-stacking domain in which the agent rearranges blocks and spheres on a table.
The challenge is primarily \emph{perceptual}: the photorealistic rendering induces strong observation-level ambiguity (lighting, textures, occlusions) over what are otherwise clean and well-understood symbolic dynamics.
The state vocabulary captures the support relation between shapes, distinguishing whether a shape rests directly on the table or on top of a specific block, and which shapes are clear on top.
Blocks and spheres are typed separately, since spheres cannot be stacked upon.
The action set consists of moves that pick up a clear shape and place it onto the table or onto a clear block; preconditions only check the destination, while universally quantified conditional effects update the previous support without requiring the planner to commit to where the shape started.
Because of this design, the size of the state space grows quickly with the number of shapes (each pair of distinct shapes contributes a candidate stacking literal), making PRB our main vehicle for stress-testing scalability under controlled conditions.

\subsection{Groceries World (GW)}\label{ap:domains:gw}

\textbf{GW} \cite{azranS3ESemanticSymbolic2025} is a tabletop manipulation domain in which a robot sorts grocery items between a fixed set of tables using a single gripper.
In contrast to PRB, the visual scene is clearer: the difficulty comes from \emph{abstraction uncertainty}, where the symbolic vocabulary does not fully capture the underlying physical dynamics, e.g., an item may fall on the floor, but the task description enforces that it must be on a table or in the robot's gripper.
The vocabulary is intentionally minimal: item-on-table relations together with gripper-state predicates indicating whether the gripper is empty and, if not, which specific item is held. There is no internal structure on items or tables (no containers, no item attributes).
The action set covers grasping and releasing, with conditional effects that gracefully handle uncertainty over which table an item currently occupies or which item the gripper holds.
The number of features per instance is fixed, which makes GW useful as a complementary stress test to PRB: the symbolic structure is much smaller, but each individual feature carries more dynamical weight.

\subsection{ALFWorld (ALF)}\label{ap:domains:alf}

\textbf{ALF} \cite{shridharALFWorldAligningText2020}, adapted from the ALFRED benchmark \cite{ALFRED20}, is a household task domain in which an agent navigates a kitchen-style environment, manipulates items, and performs object transformations such as cleaning, heating, cooling, slicing, and toggling.
We use ALF as a bounded one-shot stress test for \emph{hidden} initial-state features: task-relevant objects can be placed inside closed receptacles, producing genuinely unresolvable uncertainty from a single observation.
ALF has a substantially richer symbolic structure than the other two domains.
Objects are organized under a typed hierarchy that distinguishes manipulable items from receptacles, and both are further parameterized by type-codes (e.g., fridge, sink, microwave for receptacles; knife, butter knife for items) governed by static type-compatibility relations.
The dynamic state vocabulary covers visibility and proximity to the agent, containment and open/closed status of receptacles, the agent's grasping state, and per-object physical-state attributes (clean, hot, cool, toggled, sliced) alongside the affordance predicates that gate them.
Static type and affordance predicates are fixed within an instance and act as priors for the \gls{sse}; the dynamic predicates are the ones over which probabilistic estimates are produced and over which the \gls{mbs} is computed.
The action set spans navigation to a target item or receptacle, opening and closing receptacles, picking up and putting down items, and the transformation actions, all using visibility-based gating and conditional effects to remain safe under partial observability.

\section{Data construction}\label{ap:data}

For each domain, we generate single-observation instances by sampling a hidden ground-truth state and rendering one observation.
In PRB, instances are parameterized by the number of objects $N \in \{3,4,5,6,7,8\}$ and the goal horizon $H \in \{1, 2,4,6\}$. We generate 50 test and 10 validation instances per $(N, H)$ cell, yielding 1,200 test and 240 validation images.
In GW, instances are parameterized by $H \in \{1,2,4,6\}$; we generate 150 test and 50 validation instances per $H$, yielding 600 test and 200 validation images.
In ALF, we randomly sample 2,000 test and 1,000 validation instances from the 7,080 available tasks.
All data was processed by the GPT-4.1 \gls{vlm} \cite{openaiIntroducingGPT41API2025} to extract symbolic state feature probabilities. The model was accessed through the OpenAI API and is subject to OpenAI's Terms of Use, Usage Policies, and Sharing \& Publication Policy, which permit API use and welcome research publications related to the OpenAI API when compliant with those policies.
The flow of the prompts and extraction of the probabilities were handled by S3E \cite{azranS3ESemanticSymbolic2025}, which is available under the MIT license.
For the multi-backbone study in \cref{ap:backbones}, the same images were additionally processed through the same S3E pipeline by five open-weight \glspl{vlm} run locally: Qwen3-VL-8B-Instruct and Qwen3-VL-32B-Instruct \cite{baiQwen3VLTechnicalReport2025}, and Gemma 4 E2B, E4B, and 31B \cite{gemmateamGemma4Technical2026}, all distributed under the Apache 2.0 license.
The final dataset is linked from our code repository (\coderepo).

\section{Domain constraint extraction}\label{ap:constraints}

This appendix describes how we extract symbolic state-validity constraints from each planning task and how we compute the normalizing constant \normalizer used when conditioning the \gls{unconstBelState} on those constraints (\cref{sec:problem}).

\subsection{Notation and two constraint sets}\label{ap:constraints:notation}

Recall from \cref{sec:problem} that for a feature set \fluentSet, an at-most-one mutex group $\constraint \subseteq \fluentSet$ requires that at most one of its atoms is true in any valid state. We maintain two related constraint sets per task instance:
\begin{itemize}
    \item $\constraintSet_{\mathrm{all}}$: \emph{all} extracted at-most-one groups, possibly overlapping. These are the strongest available validity constraints and are stored in the JSON field \texttt{constraints}.
    \item $\constraintSet_Z$: a \emph{pairwise-disjoint} subfamily, $\constraint_i \cap \constraint_j = \emptyset$ for all $i \neq j$, stored as \texttt{selected\_groups}. Disjointness is what makes the normalizer of \cref{ap:constraints:norm} computable in closed form.
\end{itemize}

\subsection{Fast Downward extraction (PRB and GW)}\label{ap:constraints:fd}

For PRB and GW, $\constraintSet_{\mathrm{all}}$ and $\constraintSet_Z$ are extracted using the \gls{fd} translator's invariant machinery~\cite{helmertConciseFinitedomainRepresentations2009}.
We use \gls{fd} version 22.06, which is publicly available under the GPLv3 license. 
We parse the \gls{pddl} domain and problem, run \gls{fd}'s grounding/reachability procedure, and then run its invariant finder on the reachable grounded task. The translator is invoked with \texttt{-{}-keep-unimportant-variables} and \texttt{-{}-skip-variable-reordering} so that no variables are dropped or relabeled before fact-group extraction, and invariant generation is supplied with the reachable action-parameter information returned by grounding (omitting it would yield fewer invariants). \Gls{fd} produces (i) a possibly overlapping set of grounded mutex groups, used as $\constraintSet_{\mathrm{all}}$, and (ii) a disjoint cover of reachable facts selected for the SAS\textsuperscript{+} encoding, used as $\constraintSet_Z$.

\paragraph{Synthetic seed initial state.}
\Gls{fd} invariant generation is tied to facts and actions reachable from the supplied initial state. Some benchmark problems encode only a partial relevant initial state, which can cause \gls{fd} to prune objects that have no asserted truth value, in turn dropping the mutex groups that involve them. To avoid this, we synthesize, before invoking \gls{fd}, a deterministic seed initial state from the domain schema and the object universe of the original problem, and feed that synthetic problem to \gls{fd} instead of the original. The synthetic seed is constructed only from publicly available structural information (object names and predicate signatures) and does not read the hidden ground truth, so the extracted constraints remain task-structural rather than label-derived. For GW, every item is placed on a single canonical table, and the gripper is empty; for PRB, every shape is directly on the table and clear on top.

\paragraph{What \gls{fd} recovers.}
In GW, \gls{fd} recovers per-object placement mutexes (an object is on at most one table or in the gripper) and a gripper-capacity mutex (the gripper is empty or holds at most one object). The selected disjoint representation splits these into per-object table-location groups plus one gripper-state group, since the full groups overlap on \texttt{robot-gripping}$(o)$ atoms. In PRB, \gls{fd} recovers per-shape support mutexes (a shape is directly on the table or on at most one block) and per-block occupancy/clearance mutexes (a block is clear or supports at most one shape), with overlap on \texttt{shape-on-top-of-block}$(s,b)$ atoms; the selected groups choose a disjoint cover from the same fact-group computation.

\subsection{Custom ALF constraints}\label{ap:constraints:alf}

For ALF, \gls{fd}'s invariant finder does not recover useful nontrivial mutex groups, even when run over all reachable grounded atoms. Two of the most natural-looking candidate constraints are also unsound on this dataset: ALF's \texttt{inreceptacle}$(o,r)$ predicate is not single-valued (e.g., a pan can be associated with multiple stove burners simultaneously), and visibility predicates likewise admit multiple simultaneously true atoms. We therefore add a small set of lifted constraints that is supported by both the ALF action model and the observed ground-truth states.

Let $H = \{\texttt{holds}(o) : o \in O\}$ be the set of grounded holding atoms.
\begin{itemize}
    \item \emph{Global gripper-capacity:} $\mathrm{AtMostOne}(H)$. Added to both $\constraintSet_{\mathrm{all}}$ and $\constraintSet_Z$ when $|H| \geq 2$; otherwise omitted as vacuous.
    \item \emph{Held-vs-contained exclusions:} for each reachable \texttt{holds}$(o)$ and each reachable \texttt{inreceptacle}$(o,r)$, the pairwise mutex $\mathrm{AtMostOne}(\{\texttt{holds}(o),\texttt{inreceptacle}(o,r)\})$. These are added only to $\constraintSet_{\mathrm{all}}$, since they overlap on \texttt{holds}$(o)$ and would destroy the disjoint factorization of $\constraintSet_Z$.
\end{itemize}

\subsection{Normalizer computation}\label{ap:constraints:norm}

The normalizer for a constraint set $\constraintSet$ is $\normalizer_\constraintSet = \sum_{\state \subseteq \fluentSet} \indicator[\state \models \constraintSet] \, \unconstBeliefState(\state)$. Two computations are run.

\paragraph{Closed-form on disjoint selected groups.}
Because the groups in $\constraintSet_Z = \{\constraint_1,\ldots,\constraint_m\}$ are pairwise disjoint and atoms outside any selected group integrate to one under the independent Bernoulli product distribution \unconstBeliefState, the normalizer factorizes as $\normalizer_{\constraintSet_Z} = \prod_{i=1}^m \normalizer_{\constraint_i}$. For a single at-most-one group, only the all-false assignment and the $|\constraint|$ single-true assignments contribute nonzero mass:
\begin{equation*}
    \normalizer_\constraint
    = \prod_{\fluent \in \constraint}(1-\factoredBelief(\fluent))
    + \sum_{\fluent \in \constraint} \factoredBelief(\fluent) \prod_{\fluent' \in \constraint,\ \fluent' \neq \fluent}(1-\factoredBelief(\fluent')).
\end{equation*}
A two-state dynamic program tracking ``zero processed atoms true'' and ``exactly one processed atom true'' computes $\normalizer_\constraint$ in $O(|\constraint|)$ time, so $\normalizer_{\constraintSet_Z}$ is computed in $O(\sum_i |\constraint_i|) \leq O(|\fluentSet|)$.

\paragraph{Exact variable elimination on the full constraint set.}
$\constraintSet_{\mathrm{all}}$ generally contains overlapping groups, so the closed-form product above does not apply. We additionally attempt an exact weighted model count of $\normalizer_{\constraintSet_{\mathrm{all}}}$ by variable elimination over the factor graph whose unary factors are the per-atom Bernoulli probabilities $\factoredBelief(\fluent)$ and whose hard factors are the at-most-one indicators $\indicator[\sum_{\fluent \in \constraint_i} \state(\fluent) \leq 1]$. The graph is decomposed into connected components (two constraints share a component iff they share an atom), each component is solved independently using a min-scope variable ordering, and the per-component results are multiplied. Variable elimination is exact but exponential in the induced width; we enforce a configurable bound on the largest intermediate factor scope (default $24$). When this bound is exceeded, the variable-elimination attempt is abandoned and the closed-form $\normalizer_{\constraintSet_Z}$ is reported as the primary normalizer instead.

\subsection{Selected groups as a relaxed normalizer}\label{ap:constraints:relax}

When variable elimination terminates within the scope bound, the primary normalizer is the exact $\normalizer_{\constraintSet_{\mathrm{all}}}$ and no relaxation is involved. When it does not, the fallback $\normalizer_{\constraintSet_Z}$ is the relaxed normalizer \dnormalizer of \cref{sec:belief:comp}. The fallback is sound because every selected at-most-one group is either one of the full groups or a subset of one, so $\mathrm{AtMostOne}(\constraint) \implies \mathrm{AtMostOne}(\constraint')$ for any $\constraint' \subseteq \constraint$. This gives $\Omega(\constraintSet_{\mathrm{all}}) \subseteq \Omega(\constraintSet_Z)$ where $\Omega(\constraintSet)$ denotes the set of states satisfying \constraintSet, and hence by nonnegativity of \unconstBeliefState,
\begin{equation*}
    \normalizer_{\constraintSet_{\mathrm{all}}}
    = \sum_{\state \models \constraintSet_{\mathrm{all}}} \unconstBeliefState(\state)
    \leq \sum_{\state \models \constraintSet_Z} \unconstBeliefState(\state)
    = \normalizer_{\constraintSet_Z} = \dnormalizer.
\end{equation*}
Disjointness and the relaxation play different roles: disjointness makes $\normalizer_{\constraintSet_Z}$ cheap to compute, while the logical implication $\constraintSet_{\mathrm{all}} \models \constraintSet_Z$ makes it a sound upper bound on $\normalizer_{\constraintSet_{\mathrm{all}}}$.

A naive alternative would multiply the per-group normalizers of overlapping full constraints. This is unsound: under the factored Bernoulli belief, the validity events of overlapping groups are not independent, and the resulting product can be \emph{smaller} than $\normalizer_{\constraintSet_{\mathrm{all}}}$. For example, with atoms $x,y,z$, all marginals equal to $0.5$, and overlapping groups $\constraint_1=\{x,y\}$, $\constraint_2=\{y,z\}$, the exact probability of satisfying both is $5/8$, while $\Pr(\constraint_1\text{ valid})\Pr(\constraint_2\text{ valid}) = (3/4)^2 = 9/16 < 5/8$. Overlapping constraints must therefore either be normalized by a method that accounts for the overlap, such as variable elimination, or replaced by a disjoint family that is logically implied by the full constraints.

The relaxation preserves the \probThresh guarantee of \cref{def:mlss} in a one-sided form, which we state formally.

\begin{corollary}[One-sided guarantee under relaxed normalization]\label{cor:relaxed}
    Let $\dnormalizer \geq \normalizer$, let \taskBeliefState be the exactly normalized \gls{taskBelState} of \cref{eq:calc-belief}, and let \beliefSet be the set returned by \cref{alg:mlss} run with \dnormalizer in place of \normalizer.
    If the algorithm terminates with reported cumulative mass $\prob \geq \probThresh$, then $\sum_{\state \in \beliefSet} \taskBeliefState(\state) \geq \probThresh$, and \beliefSet is a prefix of the same probability ordering that contains the exact-normalizer $\textup{\gls{mbs}}(\taskBeliefState, \constraintSet)$.
    If instead the total reported mass $\normalizer / \dnormalizer$ is below \probThresh, the algorithm terminates with an empty heap and $\prob < \probThresh$, a condition that is detectable and must be reported rather than silently approximated.
\end{corollary}

\begin{proof}
    For every valid state, $\unconstBeliefState(\state)/\dnormalizer \leq \unconstBeliefState(\state)/\normalizer = \taskBeliefState(\state)$, so the reported mass of any set is a lower bound on its exact mass, which gives the first claim.
    Both normalizers are constants, so they do not change the ordering of states by belief, and \cref{alg:mlss} extracts the same sequence of valid states under either one (\cref{thm:mlss}).
    The algorithm halts at the first prefix of this sequence whose reported mass reaches \probThresh.
    Since the reported mass of every prefix is at most its exact mass, this prefix is at least as long as the first prefix whose exact mass reaches \probThresh, which is the exact-normalizer \gls{mbs} by \cref{cor:mlss}.
    Finally, the reported mass summed over all valid states equals $\normalizer/\dnormalizer$.
    If this is below \probThresh, no prefix reaches \probThresh, so the loop of \cref{alg:mlss} runs until the heap is empty and exits with $\prob < \probThresh$.
\end{proof}
The cost of the relaxation is therefore minimality, never validity: the returned set can be a superset of the exact-normalizer \gls{mbs}, but never smaller.

\subsection{Summary by domain}\label{ap:constraints:summary}

\cref{tab:constraints-summary} summarizes the constraint sources by domain.
This design separates validity checking from tractable normalization: $\constraintSet_{\mathrm{all}}$ provides the strongest available symbolic consistency tests and feeds the variable-elimination exact normalizer, while the disjoint $\constraintSet_Z$ provides an always-available closed-form normalizer for a sound relaxation.

\begin{table}[t]
\centering
\small
\caption{Constraint sources by domain. The primary normalizer is the variable-elimination exact $\normalizer_{\constraintSet_{\mathrm{all}}}$ when it stays within the intermediate-scope bound, with the closed-form $\normalizer_{\constraintSet_Z}$ as a sound relaxed fallback.}
\label{tab:constraints-summary}
\begin{tabular}{lp{0.28\linewidth}p{0.28\linewidth}p{0.24\linewidth}}
\toprule
\textbf{Domain} & \textbf{$\constraintSet_{\mathrm{all}}$ (\texttt{constraints})} & \textbf{$\constraintSet_Z$ (\texttt{selected\_groups})} & \textbf{Source} \\
\midrule
GW  & \gls{fd} all mutex groups: object-placement and gripper-capacity & \gls{fd} selected SAS\textsuperscript{+} groups & \gls{fd} on synthetic seed \\
PRB & \gls{fd} all mutex groups: per-shape support and per-block occupancy/clearance & \gls{fd} selected SAS\textsuperscript{+} groups & \gls{fd} on synthetic seed \\
ALF & Global holding mutex plus pairwise held-vs-contained exclusions & Global holding mutex only (when $|H|\geq 2$) & Custom lifted constraints \\
\bottomrule
\end{tabular}
\end{table}

\section{Full size vs.\ coverage results}\label{ap:static-full}

\begin{figure}
\centering
\includegraphics[width=\textwidth]{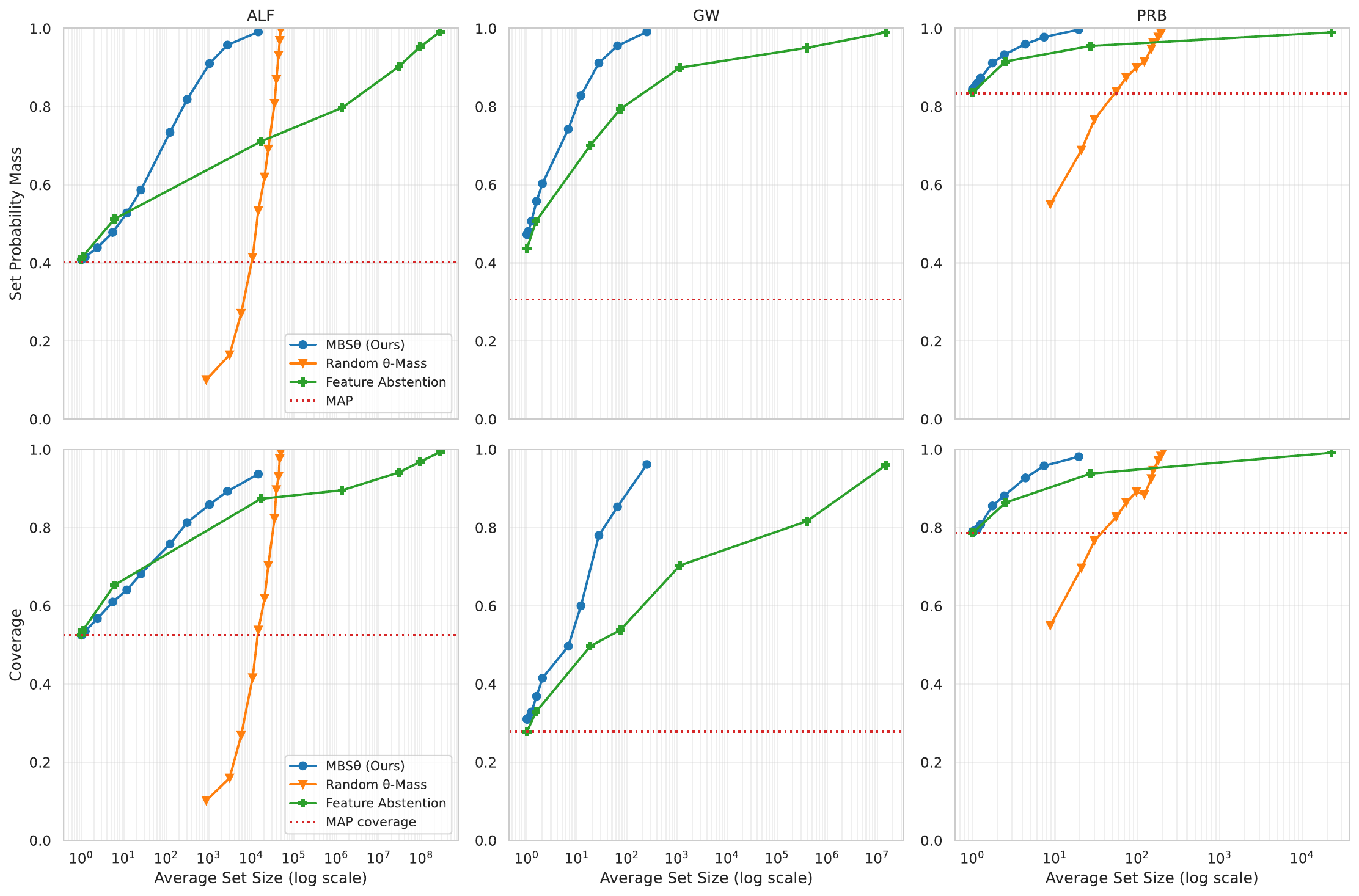}
\caption{Set size tradeoffs for a wide range of \probThresh values. (Bottom) Set size vs.\ coverage (probability of the true state being included in the \gls{belSet}). (Top) Set size vs.\ attributed mass (the probability mass of the states in the \gls{belSet} according to the \gls{taskBelState}).}
\label{fig:cov-conf-size}
\end{figure}

\cref{tab:static} summarizes the static state-set quality on a wide range of \probThresh values from 0.01 to 0.99. This is the data used to plot \cref{fig:coverage-size}, providing precise numbers for accurate comparison between methods. We extend this in \cref{fig:cov-conf-size}, where we also show the tradeoff between set size and the probability mass of the \gls{belSet} returned by each method. Here, we see a strong positive correlation between coverage and attributed mass, indicating that our three-layered belief system has merit in real-world tasks, provided the \gls{sse} is well calibrated.

\begin{figure}[t]
\centering
\includegraphics[width=\textwidth]{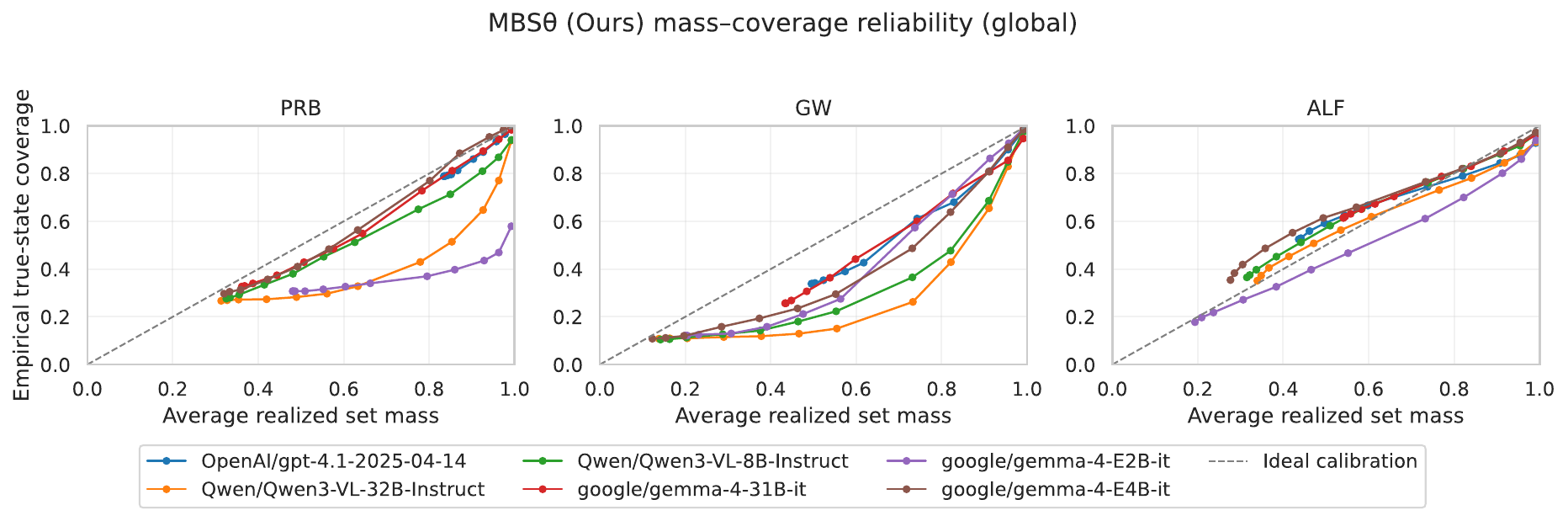}
\caption{Average realized set mass vs.\ empirical true-state coverage of \gls{mbs} for each \gls{vlm} backbone under global Platt calibration (one logistic recalibration per domain, shared by all features). Each point is one \probThresh value from the grid of \cref{tab:static}, and the dashed diagonal is ideal calibration.}
\label{fig:mass-coverage}
\end{figure}

\begin{figure}[p]
\centering
\includegraphics[width=0.78\textwidth]{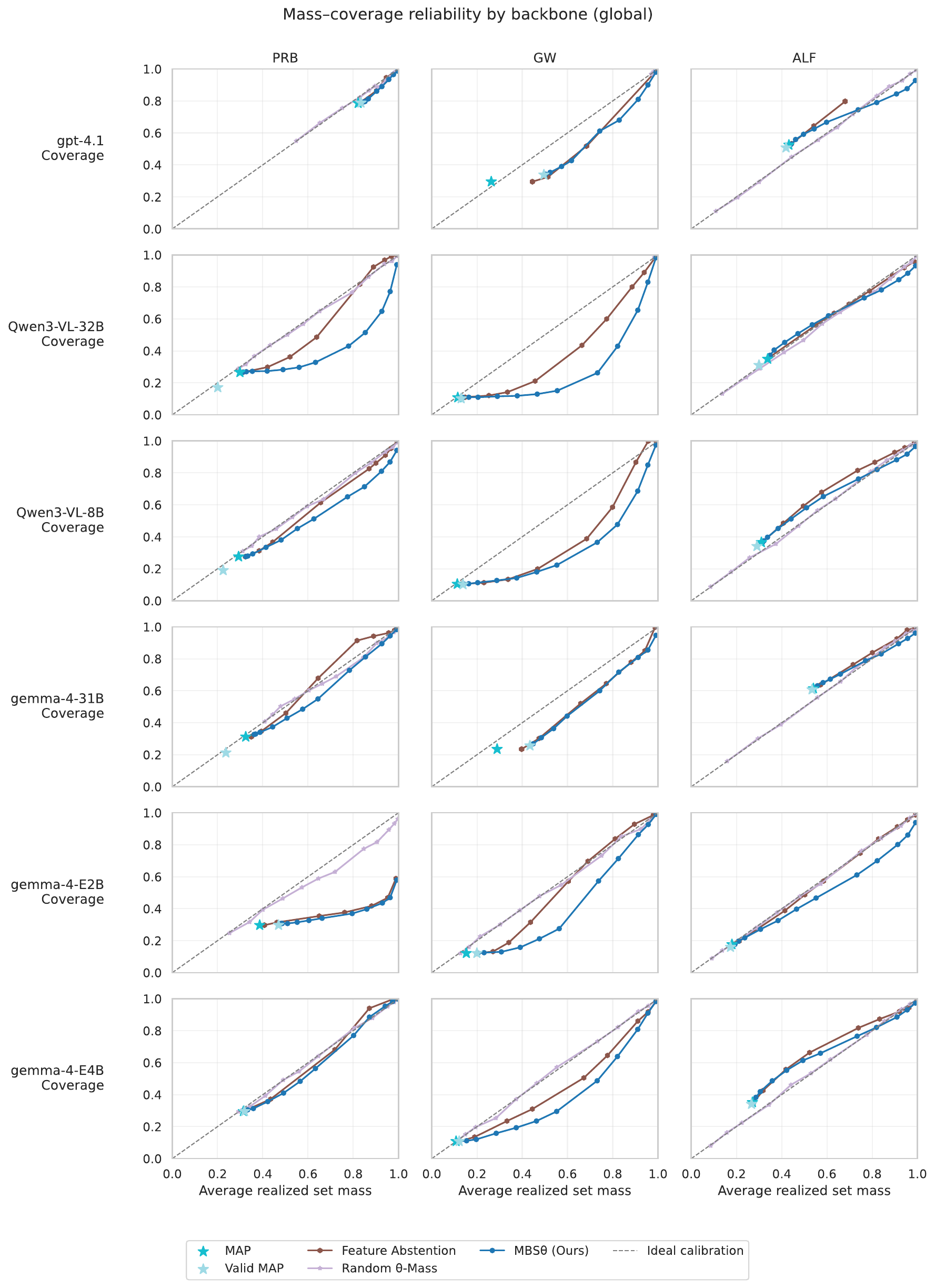}
\caption{Average realized set mass vs.\ empirical true-state coverage per \gls{vlm} backbone (rows) and domain (columns), with all methods overlaid. Curves sweep \probThresh and stars mark the \probThresh-independent \gls{map} and valid \gls{map} points.}
\label{fig:mass-coverage-by-model}
\end{figure}

\paragraph{Attributed mass vs.\ true-state coverage.}
\probThresh bounds the mass that the \gls{taskBelState} attributes to the returned set (\cref{sec:mlss}), whereas coverage is measured against ground truth.
\cref{fig:mass-coverage} plots, for \gls{mbs}, the average realized set mass against the empirical true-state coverage, where each point is one \probThresh value and the diagonal corresponds to a perfectly calibrated state-level belief.
For GPT-4.1, the absolute gap between the two is at most 0.11 across all three domains for \probThresh between 0.9 and 0.99, the operating points used in Experiment~3, and at most 0.20 anywhere on the \probThresh grid.
Deviations from conditional independence and residual feature-level miscalibration surface as this gap, which grows for the weaker open-weight backbones of \cref{ap:backbones}.
\cref{fig:mass-coverage-by-model} shows the same view for each backbone with all baselines overlaid.
The curves in both figures are computed on the common support of instances completed at every \probThresh using the later pipeline run described in \cref{ap:backbones}, and the GPT-4.1 values differ from those of \cref{tab:static}.

\begin{sidewaystable}[p]
\centering
\caption{\Gls{belSet} quality across all belief-mass thresholds. Coverage is the probability of the true state being included in the \gls{belSet}, and set size refers to the number of states in that belief set. Values in brackets are 95\% bootstrap confidence intervals using 10,000 paired resamples over test instances. Higher coverage and lower set size are better.}
\label{tab:static}
\resizebox{\columnwidth}{!}{
\begin{tabular}{lcccccccccc}
\toprule
 & Domain & \multicolumn{3}{r}{ALF} & \multicolumn{3}{r}{GW} & \multicolumn{3}{r}{PRB} \\
 &  & Coverage & Average Set Size & Average Mass & Coverage & Average Set Size & Average Mass & Coverage & Average Set Size & Average Mass \\
$\theta$ & Method &  &  &  &  &  &  &  &  &  \\
\midrule
\multirow[t]{3}{*}{0.01} & \gls{mbs} (Ours) & 0.52 [0.50, 0.55] & 1.00 [1.00, 1.00] & 0.41 [0.40, 0.42] & 0.31 [0.28, 0.35] & 1.00 [1.00, 1.00] & 0.47 [0.46, 0.49] & 0.79 [0.77, 0.81] & 1.00 [1.00, 1.00] & 0.84 [0.83, 0.86] \\
 & Random $\theta$-Mass & 0.10 [0.09, 0.12] & 866.04 [696.02, 1069.11] & 0.10 [0.09, 0.11] & 0.95 [0.89, 1.00] & 23.27 [1.00, 59.12] & 0.96 [0.92, 1.00] & 0.55 [0.52, 0.58] & 8.77 [6.31, 11.52] & 0.55 [0.52, 0.58] \\
 & Feature Abstention & 0.52 [0.50, 0.55] & 1.00 [1.00, 1.00] & 0.41 [0.40, 0.42] & 0.28 [0.24, 0.31] & 1.00 [1.00, 1.00] & 0.44 [0.42, 0.46] & 0.79 [0.77, 0.81] & 1.00 [1.00, 1.00] & 0.84 [0.82, 0.85] \\
\cline{1-11}
\multirow[t]{3}{*}{0.05} & \gls{mbs} (Ours) & 0.53 [0.50, 0.55] & 1.04 [1.03, 1.06] & 0.41 [0.40, 0.42] & 0.31 [0.28, 0.34] & 1.00 [1.00, 1.01] & 0.47 [0.46, 0.49] & 0.79 [0.77, 0.81] & 1.00 [1.00, 1.00] & 0.84 [0.83, 0.86] \\
 & Random $\theta$-Mass & 0.16 [0.14, 0.18] & 3069.37 [2534.38, 3665.43] & 0.17 [0.15, 0.18] & 0.95 [0.89, 1.00] & 61.69 [1.00, 144.58] & 0.96 [0.90, 1.00] & 0.70 [0.67, 0.72] & 20.95 [15.27, 27.00] & 0.69 [0.67, 0.71] \\
 & Feature Abstention & 0.52 [0.50, 0.55] & 1.00 [1.00, 1.00] & 0.41 [0.40, 0.42] & 0.28 [0.24, 0.31] & 1.00 [1.00, 1.00] & 0.44 [0.42, 0.46] & 0.79 [0.76, 0.81] & 1.00 [1.00, 1.00] & 0.84 [0.82, 0.85] \\
\cline{1-11}
\multirow[t]{3}{*}{0.10} & \gls{mbs} (Ours) & 0.53 [0.51, 0.56] & 1.24 [1.19, 1.28] & 0.42 [0.40, 0.43] & 0.31 [0.28, 0.34] & 1.01 [1.00, 1.02] & 0.47 [0.46, 0.49] & 0.79 [0.77, 0.81] & 1.01 [1.00, 1.02] & 0.84 [0.83, 0.86] \\
 & Random $\theta$-Mass & 0.27 [0.25, 0.29] & 5739.67 [4676.61, 6803.03] & 0.27 [0.26, 0.28] & 0.95 [0.89, 1.00] & 170.23 [39.53, 383.89] & 0.97 [0.92, 1.00] & 0.77 [0.74, 0.79] & 30.08 [21.57, 39.39] & 0.77 [0.75, 0.79] \\
 & Feature Abstention & 0.52 [0.50, 0.55] & 1.00 [1.00, 1.00] & 0.41 [0.40, 0.42] & 0.28 [0.25, 0.31] & 1.00 [1.00, 1.00] & 0.44 [0.42, 0.46] & 0.79 [0.76, 0.81] & 1.00 [1.00, 1.00] & 0.84 [0.82, 0.85] \\
\cline{1-11}
\multirow[t]{3}{*}{0.20} & \gls{mbs} (Ours) & 0.57 [0.55, 0.59] & 2.39 [2.18, 2.58] & 0.44 [0.43, 0.45] & 0.31 [0.27, 0.35] & 1.07 [1.04, 1.09] & 0.48 [0.46, 0.50] & 0.79 [0.77, 0.81] & 1.03 [1.02, 1.06] & 0.85 [0.84, 0.86] \\
 & Random $\theta$-Mass & 0.42 [0.39, 0.44] & 10692.09 [8707.97, 12765.55] & 0.41 [0.40, 0.43] & 0.97 [0.92, 1.00] & 206.36 [1.00, 463.72] & 0.98 [0.95, 1.00] & 0.83 [0.81, 0.85] & 55.33 [38.66, 75.37] & 0.84 [0.82, 0.85] \\
 & Feature Abstention & 0.52 [0.50, 0.55] & 1.00 [1.00, 1.00] & 0.41 [0.40, 0.42] & 0.28 [0.24, 0.31] & 1.00 [1.00, 1.00] & 0.44 [0.42, 0.46] & 0.79 [0.76, 0.81] & 1.00 [1.00, 1.00] & 0.84 [0.82, 0.85] \\
\cline{1-11}
\multirow[t]{3}{*}{0.30} & \gls{mbs} (Ours) & 0.61 [0.59, 0.63] & 5.46 [4.92, 6.10] & 0.48 [0.47, 0.49] & 0.33 [0.29, 0.37] & 1.23 [1.19, 1.28] & 0.51 [0.49, 0.52] & 0.79 [0.77, 0.82] & 1.08 [1.05, 1.12] & 0.85 [0.84, 0.86] \\
 & Random $\theta$-Mass & 0.54 [0.51, 0.56] & 14531.53 [12105.55, 17425.03] & 0.53 [0.52, 0.54] & 0.97 [0.92, 1.00] & 313.11 [1.00, 773.28] & 0.99 [0.97, 1.00] & 0.86 [0.84, 0.88] & 73.04 [49.59, 101.09] & 0.87 [0.86, 0.89] \\
 & Feature Abstention & 0.52 [0.50, 0.55] & 1.00 [1.00, 1.00] & 0.41 [0.40, 0.42] & 0.28 [0.24, 0.31] & 1.00 [1.00, 1.00] & 0.44 [0.42, 0.45] & 0.79 [0.76, 0.81] & 1.00 [1.00, 1.00] & 0.84 [0.82, 0.85] \\
\cline{1-11}
\multirow[t]{3}{*}{0.40} & \gls{mbs} (Ours) & 0.64 [0.62, 0.66] & 11.76 [10.35, 13.24] & 0.53 [0.52, 0.53] & 0.37 [0.33, 0.40] & 1.55 [1.49, 1.63] & 0.56 [0.55, 0.57] & 0.80 [0.77, 0.82] & 1.15 [1.10, 1.20] & 0.86 [0.85, 0.87] \\
 & Random $\theta$-Mass & 0.62 [0.59, 0.64] & 20427.42 [16700.43, 24694.47] & 0.62 [0.61, 0.63] & 0.98 [0.95, 1.00] & 360.30 [1.00, 810.62] & 0.99 [0.98, 1.00] & 0.89 [0.87, 0.91] & 97.19 [60.52, 138.75] & 0.90 [0.89, 0.91] \\
 & Feature Abstention & 0.54 [0.51, 0.56] & 1.07 [1.06, 1.08] & 0.42 [0.40, 0.43] & 0.28 [0.24, 0.32] & 1.00 [1.00, 1.00] & 0.44 [0.42, 0.46] & 0.79 [0.77, 0.81] & 1.00 [1.00, 1.00] & 0.84 [0.82, 0.85] \\
\cline{1-11}
\multirow[t]{3}{*}{0.50} & \gls{mbs} (Ours) & 0.68 [0.66, 0.70] & 25.24 [22.30, 28.44] & 0.59 [0.58, 0.59] & 0.41 [0.38, 0.45] & 2.04 [1.92, 2.15] & 0.60 [0.59, 0.62] & 0.81 [0.78, 0.83] & 1.26 [1.18, 1.35] & 0.87 [0.86, 0.88] \\
 & Random $\theta$-Mass & 0.70 [0.68, 0.73] & 24937.30 [20295.77, 30291.91] & 0.69 [0.68, 0.70] & 0.98 [0.94, 1.00] & 407.08 [1.00, 932.66] & 0.99 [0.99, 1.00] & 0.88 [0.87, 0.90] & 121.63 [74.96, 175.06] & 0.92 [0.91, 0.92] \\
 & Feature Abstention & 0.65 [0.63, 0.67] & 6.02 [5.13, 7.07] & 0.51 [0.50, 0.52] & 0.33 [0.29, 0.36] & 1.51 [1.43, 1.62] & 0.51 [0.49, 0.52] & 0.79 [0.76, 0.81] & 1.00 [1.00, 1.00] & 0.84 [0.82, 0.85] \\
\cline{1-11}
\multirow[t]{3}{*}{0.70} & \gls{mbs} (Ours) & 0.76 [0.74, 0.78] & 121.36 [105.81, 138.25] & 0.73 [0.73, 0.74] & 0.50 [0.46, 0.54] & 6.69 [6.39, 7.04] & 0.74 [0.74, 0.75] & 0.86 [0.84, 0.87] & 1.74 [1.55, 1.97] & 0.91 [0.91, 0.92] \\
 & Random $\theta$-Mass & 0.82 [0.80, 0.84] & 34983.43 [28719.78, 42244.02] & 0.81 [0.80, 0.81] & 1.00 [1.00, 1.00] & 246.00 [1.00, 549.30] & 0.99 [0.98, 1.00] & 0.93 [0.91, 0.94] & 147.71 [89.10, 217.77] & 0.95 [0.94, 0.95] \\
 & Feature Abstention & 0.87 [0.86, 0.89] & 16799.79 [8711.69, 27580.03] & 0.71 [0.70, 0.72] & 0.50 [0.46, 0.54] & 18.38 [15.17, 22.08] & 0.70 [0.69, 0.71] & 0.79 [0.76, 0.81] & 1.00 [1.00, 1.00] & 0.84 [0.82, 0.85] \\
\cline{1-11}
\multirow[t]{3}{*}{0.80} & \gls{mbs} (Ours) & 0.81 [0.80, 0.83] & 307.66 [264.46, 348.33] & 0.82 [0.82, 0.82] & 0.60 [0.56, 0.64] & 11.99 [11.29, 12.71] & 0.83 [0.82, 0.83] & 0.88 [0.86, 0.90] & 2.43 [2.08, 2.86] & 0.93 [0.93, 0.94] \\
 & Random $\theta$-Mass & 0.90 [0.88, 0.91] & 38716.36 [31639.23, 46603.39] & 0.87 [0.87, 0.87] & 0.97 [0.92, 1.00] & 315.78 [1.00, 696.75] & 0.99 [0.98, 1.00] & 0.95 [0.93, 0.96] & 155.09 [99.63, 225.51] & 0.96 [0.96, 0.97] \\
 & Feature Abstention & 0.90 [0.88, 0.91] & $1.37{\times}10^6$ [$8.35{\times}10^5$, $2.02{\times}10^6$] & 0.80 [0.79, 0.80] & 0.54 [0.49, 0.58] & 73.83 [62.88, 84.68] & 0.79 [0.79, 0.80] & 0.79 [0.76, 0.81] & 1.00 [1.00, 1.00] & 0.84 [0.82, 0.85] \\
\cline{1-11}
\multirow[t]{3}{*}{0.90} & \gls{mbs} (Ours) & 0.86 [0.84, 0.87] & 1039.21 [889.86, 1185.09] & 0.91 [0.91, 0.91] & 0.78 [0.75, 0.81] & 27.31 [25.74, 28.95] & 0.91 [0.91, 0.91] & 0.93 [0.91, 0.94] & 4.38 [3.48, 5.45] & 0.96 [0.96, 0.96] \\
 & Random $\theta$-Mass & 0.93 [0.92, 0.94] & 43608.47 [35055.79, 52859.34] & 0.93 [0.93, 0.93] & 1.00 [1.00, 1.00] & 423.30 [1.00, 994.85] & 1.00 [1.00, 1.00] & 0.97 [0.96, 0.98] & 178.98 [116.77, 262.84] & 0.98 [0.98, 0.98] \\
 & Feature Abstention & 0.94 [0.93, 0.95] & $2.96{\times}10^7$ [$1.94{\times}10^7$, $4.33{\times}10^7$] & 0.90 [0.90, 0.91] & 0.70 [0.67, 0.74] & 1138.63 [871.31, 1438.83] & 0.90 [0.89, 0.90] & 0.86 [0.84, 0.88] & 2.51 [1.77, 3.45] & 0.92 [0.91, 0.92] \\
\cline{1-11}
\multirow[t]{3}{*}{0.95} & \gls{mbs} (Ours) & 0.89 [0.88, 0.91] & 2737.81 [2297.09, 3152.69] & 0.96 [0.96, 0.96] & 0.85 [0.82, 0.88] & 64.33 [60.79, 68.02] & 0.96 [0.95, 0.96] & 0.96 [0.95, 0.97] & 7.38 [5.77, 9.28] & 0.98 [0.98, 0.98] \\
 & Random $\theta$-Mass & 0.98 [0.97, 0.98] & 46391.44 [37589.30, 54982.88] & 0.97 [0.97, 0.97] & 1.00 [1.00, 1.00] & 506.17 [1.00, 1115.55] & 1.00 [1.00, 1.00] & 0.98 [0.98, 0.99] & 192.88 [122.52, 279.53] & 0.99 [0.99, 0.99] \\
 & Feature Abstention & 0.97 [0.96, 0.98] & $9.33{\times}10^7$ [$5.99{\times}10^7$, $1.30{\times}10^8$] & 0.95 [0.95, 0.96] & 0.82 [0.79, 0.85] & $3.95{\times}10^5$ [$3.28{\times}10^5$, $4.66{\times}10^5$] & 0.95 [0.95, 0.95] & 0.94 [0.92, 0.95] & 26.92 [7.41, 58.78] & 0.96 [0.95, 0.96] \\
\cline{1-11}
\multirow[t]{3}{*}{0.99} & \gls{mbs} (Ours) & 0.94 [0.92, 0.95] & 14548.28 [12145.79, 16947.27] & 0.99 [0.99, 0.99] & 0.96 [0.94, 0.98] & 248.48 [235.85, 261.54] & 0.99 [0.99, 0.99] & 0.98 [0.97, 0.99] & 19.47 [14.42, 25.76] & 1.00 [1.00, 1.00] \\
 & Random $\theta$-Mass & 0.99 [0.99, 1.00] & 48959.12 [40236.30, 59356.38] & 1.00 [1.00, 1.00] & 1.00 [1.00, 1.00] & 555.42 [1.00, 1290.43] & 1.00 [1.00, 1.00] & 1.00 [0.99, 1.00] & 206.76 [134.24, 303.06] & 1.00 [1.00, 1.00] \\
 & Feature Abstention & 0.99 [0.99, 1.00] & $2.77{\times}10^8$ [$1.79{\times}10^8$, $3.99{\times}10^8$] & 0.99 [0.99, 0.99] & 0.96 [0.94, 0.97] & $1.48{\times}10^7$ [$1.39{\times}10^7$, $1.57{\times}10^7$] & 0.99 [0.99, 0.99] & 0.99 [0.99, 1.00] & 22645.68 [2408.07, 50652.07] & 0.99 [0.99, 0.99] \\
\cline{1-11}
point & MAP & 0.52 [0.50, 0.55] & 1.00 [1.00, 1.00] & 0.40 [0.39, 0.41] & 0.28 [0.24, 0.32] & 1.00 [1.00, 1.00] & 0.31 [0.29, 0.33] & 0.79 [0.76, 0.81] & 1.00 [1.00, 1.00] & 0.83 [0.82, 0.85] \\
\bottomrule
\end{tabular}
}
\end{sidewaystable}

\FloatBarrier
\section{Results across \texorpdfstring{\gls{vlm}}{VLM} backbones}\label{ap:backbones}

The main text uses GPT-4.1 as the \gls{sse} backbone.
To test whether the results depend on this choice, we re-ran Experiments 1 and 3 with five open-weight \glspl{vlm}: Qwen3-VL-8B and Qwen3-VL-32B \cite{baiQwen3VLTechnicalReport2025}, and Gemma 4 E2B, E4B, and 31B \cite{gemmateamGemma4Technical2026}.
All backbones share the prompting scheme, Platt calibration protocol, constraints, and planning budget used for GPT-4.1 (\cref{sec:experiments}).
The results in this appendix come from a single later run of the pipeline covering all six backbones, with Platt scaling re-fit on a held-out calibration split, and are computed on the common support of instances completed at every \probThresh. The GPT-4.1 values differ from those of \cref{tab:static,tab:planning}.

\begin{figure}[p]
\centering
\includegraphics[width=0.78\textwidth]{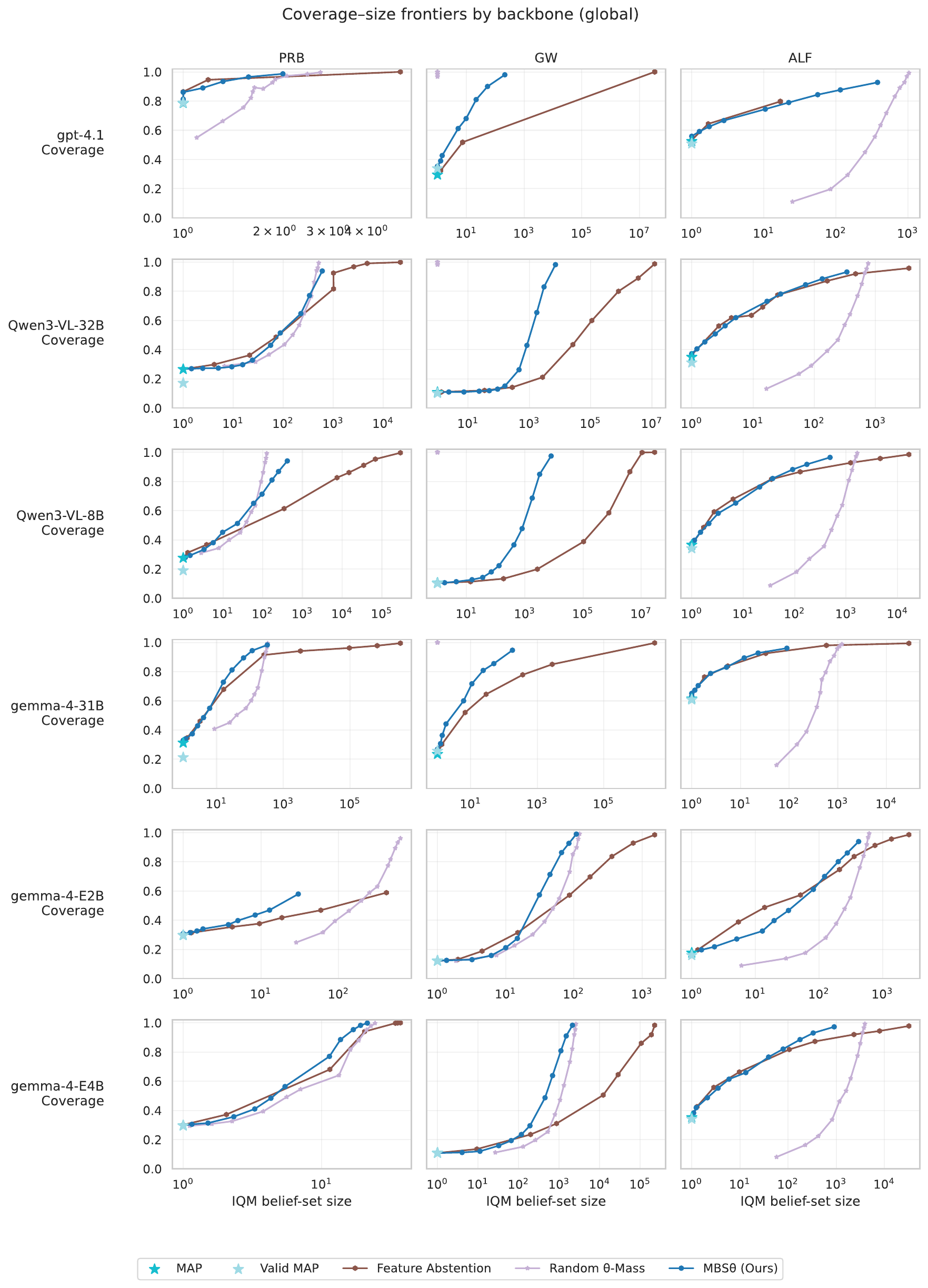}
\caption{Coverage-compactness frontiers per \gls{vlm} backbone (rows) and domain (columns) under global Platt calibration. Curves sweep \probThresh, the horizontal axis is the interquartile mean \gls{belSet} size on a log scale, and stars mark the \probThresh-independent \gls{map} and valid \gls{map} points. Top left is better.}
\label{fig:cov-size-by-model}
\end{figure}

\paragraph{Coverage and compactness.}
\cref{tab:backbones-exp1} reports true-state coverage and \gls{belSet} size at $\probThresh=0.9$ for every backbone, and \cref{fig:cov-size-by-model} shows the full coverage-compactness frontiers.
In addition to the baselines of Experiment 1, we report \emph{valid \gls{map}}, the \gls{map} state projected onto the constraint-satisfying states, i.e., \gls{map} with the same strict constraints as \gls{mbs}.
It does not close the coverage gap to \gls{mbs} on any backbone, so the belief set contributes beyond constraint filtering.
Set sizes are interquartile means over instances, which are robust to the heavy tail of the feature-abstention sets.
For GPT-4.1 only, the Feature Abstention thresholds \absThresh are those calibrated for the earlier run of \cref{tab:static} and were not re-calibrated after the Platt re-fit, so the achieved mass does not track \probThresh on GW and ALF.
On GW at $\probThresh=0.9$, \absThresh exceeds every calibrated confidence, so all features are abstained.
This affects that row of \cref{tab:backbones-exp1} and its curves in \cref{fig:cov-size-by-model,fig:mass-coverage-by-model}, and \cref{tab:static} gives the comparison with matched thresholds.
\gls{mbs} preserves its coverage-per-state advantage on all six backbones, and the advantage is often largest for the open-weight models, whose calibration is weaker and whose \gls{map} estimates degrade most.
For example, on PRB with Qwen3-VL-8B, \gls{mbs} reaches 0.81 coverage where \gls{map} reaches 0.28.
Large sets on poorly calibrated backbones are the method behaving as specified: when the marginals are diffuse, reaching the requested mass requires more states, and the set size is a legible signal of perception quality.

\begin{table}[t]
\centering
\small
\caption{Experiment 1 across \gls{vlm} backbones at $\probThresh=0.9$ under global Platt calibration. Cov.\ is the true-state coverage with its 95\% bootstrap confidence interval over test instances and Size is the interquartile mean \gls{belSet} size. Valid \gls{map} is the \gls{map} state projected onto the constraint-satisfying states. Higher coverage and lower size are better. The GPT-4.1 Feature Abstention row uses thresholds from the earlier run (see text).}
\label{tab:backbones-exp1}
\resizebox{\textwidth}{!}{%
\begin{tabular}{@{}llcccccc@{}}
\toprule
 & & \multicolumn{2}{c}{PRB} & \multicolumn{2}{c}{GW} & \multicolumn{2}{c}{ALF} \\
\cmidrule(lr){3-4} \cmidrule(lr){5-6} \cmidrule(lr){7-8}
Backbone & Method & Cov.\ [95\% CI] & Size & Cov.\ [95\% CI] & Size & Cov.\ [95\% CI] & Size \\
\midrule
\multirow{4}{*}{GPT-4.1} & \gls{mbs} (Ours) & 0.93 [0.92, 0.95] & 1.35 & 0.81 [0.78, 0.84] & 21.9 & 0.84 [0.83, 0.86] & 56.1 \\
 & Valid \gls{map} & 0.79 [0.77, 0.81] & 1 & 0.34 [0.30, 0.38] & 1 & 0.51 [0.48, 0.53] & 1 \\
 & \gls{map} & 0.79 [0.76, 0.81] & 1 & 0.29 [0.26, 0.33] & 1 & 0.52 [0.50, 0.55] & 1 \\
 & Feature Abstention & 0.86 [0.84, 0.88] & 1 & 1.00 [1.00, 1.00] & $3.36{\times}10^{7}$ & 0.80 [0.78, 0.82] & 17 \\
\midrule
\multirow{4}{*}{Qwen3-VL-32B} & \gls{mbs} (Ours) & 0.65 [0.61, 0.68] & 226 & 0.65 [0.61, 0.69] & $1.75{\times}10^{3}$ & 0.84 [0.83, 0.86] & 72.1 \\
 & Valid \gls{map} & 0.17 [0.15, 0.19] & 1 & 0.10 [0.08, 0.13] & 1 & 0.31 [0.29, 0.33] & 1 \\
 & \gls{map} & 0.27 [0.23, 0.30] & 1 & 0.11 [0.08, 0.13] & 1 & 0.35 [0.33, 0.37] & 1 \\
 & Feature Abstention & 0.97 [0.95, 0.98] & $2.59{\times}10^{3}$ & 0.80 [0.77, 0.83] & $8.12{\times}10^{5}$ & 0.87 [0.85, 0.89] & 163 \\
\midrule
\multirow{4}{*}{Qwen3-VL-8B} & \gls{mbs} (Ours) & 0.81 [0.78, 0.84] & 171 & 0.69 [0.65, 0.73] & $1.8{\times}10^{3}$ & 0.88 [0.87, 0.90] & 91.4 \\
 & Valid \gls{map} & 0.19 [0.17, 0.21] & 1 & 0.10 [0.08, 0.13] & 1 & 0.34 [0.32, 0.36] & 1 \\
 & \gls{map} & 0.28 [0.25, 0.31] & 1 & 0.10 [0.08, 0.13] & 1 & 0.37 [0.34, 0.39] & 1 \\
 & Feature Abstention & 0.91 [0.89, 0.93] & $3.48{\times}10^{4}$ & 0.87 [0.84, 0.89] & $4.07{\times}10^{6}$ & 0.93 [0.92, 0.94] & $1.23{\times}10^{3}$ \\
\midrule
\multirow{4}{*}{Gemma 4 31B} & \gls{mbs} (Ours) & 0.89 [0.87, 0.92] & 64.1 & 0.81 [0.78, 0.84] & 23.8 & 0.89 [0.88, 0.91] & 12 \\
 & Valid \gls{map} & 0.21 [0.19, 0.23] & 1 & 0.26 [0.22, 0.29] & 1 & 0.61 [0.59, 0.63] & 1 \\
 & \gls{map} & 0.31 [0.28, 0.35] & 1 & 0.23 [0.20, 0.27] & 1 & 0.61 [0.59, 0.64] & 1 \\
 & Feature Abstention & 0.96 [0.95, 0.98] & $9.5{\times}10^{4}$ & 0.78 [0.74, 0.81] & 366 & 0.93 [0.91, 0.94] & 33.1 \\
\midrule
\multirow{4}{*}{Gemma 4 E2B} & \gls{mbs} (Ours) & 0.44 [0.40, 0.46] & 8.44 & 0.86 [0.83, 0.89] & 66.2 & 0.80 [0.78, 0.82] & 202 \\
 & Valid \gls{map} & 0.30 [0.27, 0.32] & 1 & 0.12 [0.10, 0.15] & 1 & 0.16 [0.14, 0.17] & 1 \\
 & \gls{map} & 0.30 [0.27, 0.32] & 1 & 0.12 [0.10, 0.15] & 1 & 0.18 [0.16, 0.19] & 1 \\
 & Feature Abstention & 0.42 [0.39, 0.45] & 18.6 & 0.84 [0.81, 0.86] & 363 & 0.91 [0.90, 0.93] & 760 \\
\midrule
\multirow{4}{*}{Gemma 4 E4B} & \gls{mbs} (Ours) & 0.95 [0.94, 0.96] & 17 & 0.81 [0.78, 0.84] & $1.11{\times}10^{3}$ & 0.89 [0.87, 0.90] & 177 \\
 & Valid \gls{map} & 0.30 [0.27, 0.32] & 1 & 0.11 [0.09, 0.14] & 1 & 0.34 [0.32, 0.36] & 1 \\
 & \gls{map} & 0.30 [0.27, 0.32] & 1 & 0.11 [0.08, 0.13] & 1 & 0.35 [0.33, 0.37] & 1 \\
 & Feature Abstention & 1.00 [0.99, 1.00] & 34.1 & 0.86 [0.83, 0.89] & $1.06{\times}10^{5}$ & 0.92 [0.91, 0.93] & $2.34{\times}10^{3}$ \\
\bottomrule
\end{tabular}%
}
\end{table}

\paragraph{Downstream planning.}
\cref{tab:backbones-exp3} repeats Experiment 3 with \gls{ams} for every backbone.
It adds an oracle that plans from the true state as a ceiling, and the \gls{ams}-Loose variant, which uses only the disjoint selected constraint groups of \cref{ap:constraints}.
\gls{ams} improves over \gls{map} on all six PRB backbones and on four of six GW backbones.
On the two remaining GW backbones, Gemma 4 E2B and E4B, the marginals are too diffuse to be informative: every method including \gls{map} stays near 10\% under the planning budget while the oracle remains at 95.0\%, and \gls{ams} falls back to \gls{map}-level performance.
The GW gain grows substantially when perception is degraded yet still informative, e.g., \gls{ams} reaches 68.7\% against 31.0\% for \gls{map} with Qwen3-VL-32B, which supports the headroom explanation of the modest GPT-4.1 gain on GW given in \cref{sec:experiments}.
The high timeout rates of \gls{ams}-Loose on the open-weight backbones arise because the disjoint groups prune far fewer of the $2^{|\fluentSet|}$ completions, so the prefix search spends its budget on large candidate sets, which is exactly the failure that the full constraint set prevents.

\begin{table}[t]
\centering
\small
\caption{Experiment 3 across \gls{vlm} backbones under global Platt calibration: downstream planning success rate with 95\% bootstrap confidence intervals over the 1,200 PRB and 600 GW test instances. TO is the fraction of instances on which the method exhausted the planning budget, which count as failures. The oracle plans from the true state. \gls{ams}-Loose uses only the disjoint selected constraint groups, and \gls{ams} the full constraint set with the relaxed normalizer, as in \cref{tab:planning}.}
\label{tab:backbones-exp3}
\begin{tabular}{@{}llcccc@{}}
\toprule
 & & \multicolumn{2}{c}{PRB} & \multicolumn{2}{c}{GW} \\
\cmidrule(lr){3-4} \cmidrule(lr){5-6}
Backbone & Method & Success [95\% CI] & TO & Success [95\% CI] & TO \\
\midrule
\multirow{5}{*}{GPT-4.1} & Oracle & 100.0\% [99.7, 100.0] & 0\% & 95.0\% [93.0, 96.5] & 0\% \\
 & \gls{map} & 16.7\% [14.7, 18.9] & 0\% & 71.7\% [67.9, 75.1] & 0\% \\
 & Adaptive Abst. & 16.8\% [14.7, 19.0] & 0\% & 74.0\% [70.3, 77.3] & 0\% \\
 & \gls{ams}-Loose & 16.2\% [14.2, 18.4] & 0\% & 74.7\% [71.0, 78.0] & 1\% \\
 & \gls{ams} & 32.4\% [29.8, 35.1] & 0\% & 79.2\% [75.7, 82.2] & 2\% \\
\midrule
\multirow{5}{*}{Qwen3-VL-32B} & Oracle & 100.0\% [99.7, 100.0] & 0\% & 95.0\% [93.0, 96.5] & 0\% \\
 & \gls{map} & 18.3\% [16.2, 20.6] & 0\% & 31.0\% [27.4, 34.8] & 0\% \\
 & Adaptive Abst. & 18.5\% [16.4, 20.8] & 0\% & 31.8\% [28.2, 35.7] & 0\% \\
 & \gls{ams}-Loose & 17.0\% [15.0, 19.2] & 57\% & 31.2\% [27.6, 35.0] & 68\% \\
 & \gls{ams} & 37.8\% [35.1, 40.5] & 0\% & 68.7\% [64.8, 72.2] & 6\% \\
\midrule
\multirow{5}{*}{Qwen3-VL-8B} & Oracle & 100.0\% [99.7, 100.0] & 0\% & 95.0\% [93.0, 96.5] & 0\% \\
 & \gls{map} & 22.3\% [20.1, 24.8] & 0\% & 33.7\% [30.0, 37.5] & 0\% \\
 & Adaptive Abst. & 22.8\% [20.5, 25.3] & 0\% & 35.8\% [32.1, 39.7] & 0\% \\
 & \gls{ams}-Loose & 20.2\% [18.1, 22.6] & 52\% & 34.8\% [31.1, 38.7] & 64\% \\
 & \gls{ams} & 55.4\% [52.6, 58.2] & 0\% & 65.3\% [61.4, 69.0] & 0\% \\
\midrule
\multirow{5}{*}{Gemma 4 31B} & Oracle & 100.0\% [99.7, 100.0] & 0\% & 95.0\% [93.0, 96.5] & 0\% \\
 & \gls{map} & 35.0\% [32.4, 37.7] & 0\% & 66.2\% [62.3, 69.8] & 0\% \\
 & Adaptive Abst. & 35.1\% [32.4, 37.8] & 0\% & 69.0\% [65.2, 72.6] & 0\% \\
 & \gls{ams}-Loose & 29.3\% [26.8, 32.0] & 46\% & 69.3\% [65.5, 72.9] & 2\% \\
 & \gls{ams} & 84.1\% [81.9, 86.0] & 0\% & 80.5\% [77.1, 83.5] & 0\% \\
\midrule
\multirow{5}{*}{Gemma 4 E2B} & Oracle & 100.0\% [99.7, 100.0] & 0\% & 95.0\% [93.0, 96.5] & 0\% \\
 & \gls{map} & 24.8\% [22.5, 27.4] & 0\% & 10.2\% [8.0, 12.8] & 0\% \\
 & Adaptive Abst. & 28.1\% [25.6, 30.7] & 0\% & 10.2\% [8.0, 12.8] & 0\% \\
 & \gls{ams}-Loose & 19.8\% [17.7, 22.2] & 2\% & 10.2\% [8.0, 12.8] & 8\% \\
 & \gls{ams} & 42.8\% [40.1, 45.7] & 0\% & 10.2\% [8.0, 12.8] & 6\% \\
\midrule
\multirow{5}{*}{Gemma 4 E4B} & Oracle & 100.0\% [99.7, 100.0] & 0\% & 95.0\% [93.0, 96.5] & 0\% \\
 & \gls{map} & 17.6\% [15.5, 19.8] & 0\% & 11.5\% [9.2, 14.3] & 0\% \\
 & Adaptive Abst. & 17.7\% [15.6, 19.9] & 0\% & 11.5\% [9.2, 14.3] & 0\% \\
 & \gls{ams}-Loose & 16.8\% [14.8, 19.1] & 4\% & 10.3\% [8.1, 13.0] & 62\% \\
 & \gls{ams} & 32.2\% [29.7, 34.9] & 0\% & 10.3\% [8.1, 13.0] & 52\% \\
\bottomrule
\end{tabular}
\end{table}

\FloatBarrier
\section{Sensitivity to correlated perception errors}\label{ap:correlated}

The belief converter of \cref{sec:belief:comp} assumes conditional independence between features.
Task constraints restore the dominant task-level dependencies through mutex structure, but perceptual dependencies, e.g., a single lighting condition that corrupts several features at once, remain unmodeled.
The coverage curves of Experiment 1 already absorb the real error structure of a \gls{vlm}, since coverage is measured against ground truth.
Here we isolate what conditional independence alone costs in a controlled study.

\paragraph{Protocol.}
For each test instance we take the real calibrated marginals $p^0$ and select a feature block that a shared nuisance would corrupt together: either the features of the largest mutex-coupled groups, the adversarial case since the constraint layer already couples them, or the least-decided features, the ``shared lighting'' case.
We then define a correlated joint as a mixture driven by a latent nuisance event $E \sim \text{Bernoulli}(\pi)$,
\begin{equation*}
    P_{\text{true}}(\state) = (1-\pi)\prod_{\fluent \in \fluentSet} p^0_\fluent(\state(\fluent)) + \pi \prod_{\fluent \in \fluentSet} p^1_\fluent(\state(\fluent)),
\end{equation*}
where $p^1$ equals $p^0$ outside the block and inside it each marginal is pulled toward its opposite, $p^1_\fluent = p^0_\fluent + \rho\,(1 - 2p^0_\fluent)$, so that $\rho = 1$ mirrors the marginal.
Conditioned on $E$, the features are independent, but marginally they are not.
The estimator receives the mixture's exact marginals $(1-\pi)p^0 + \pi p^1$, so it is perfectly feature-calibrated by construction and the conditional-independence lift is the only remaining discrepancy.
No ground truth enters the belief.
We run \gls{pravas} on the resulting product belief and score the returned set against the correlated joint and against a certified oracle, the true joint's own minimal \probThresh-mass set over valid states, computed by a best-first search over both mixture components with a stopping certificate.
We report five quantities: the total variation between the product belief and the joint, computed exactly on the affected block; the attributed minus true mass of the returned set, where the true mass is the exact expected coverage under the joint; the fraction of instances whose true mass still reaches \probThresh; the set size relative to the oracle set; and the ground-truth coverage of each method.
We sweep $\pi \in \{0, 0.1, 0.2, 0.3, 0.5\}$ with $\rho = 1$ at $\probThresh = 0.9$ for GPT-4.1 under global Platt calibration.
The case $\pi=0$ is the uncorrelated control, where the joint is the product belief and every distortion metric is exactly zero.
Instances whose oracle set is not certified are dropped for every method, so all contrasts are paired.

\begin{figure}[t]
\centering
\includegraphics[trim=0 0 0 50,clip,width=\textwidth]{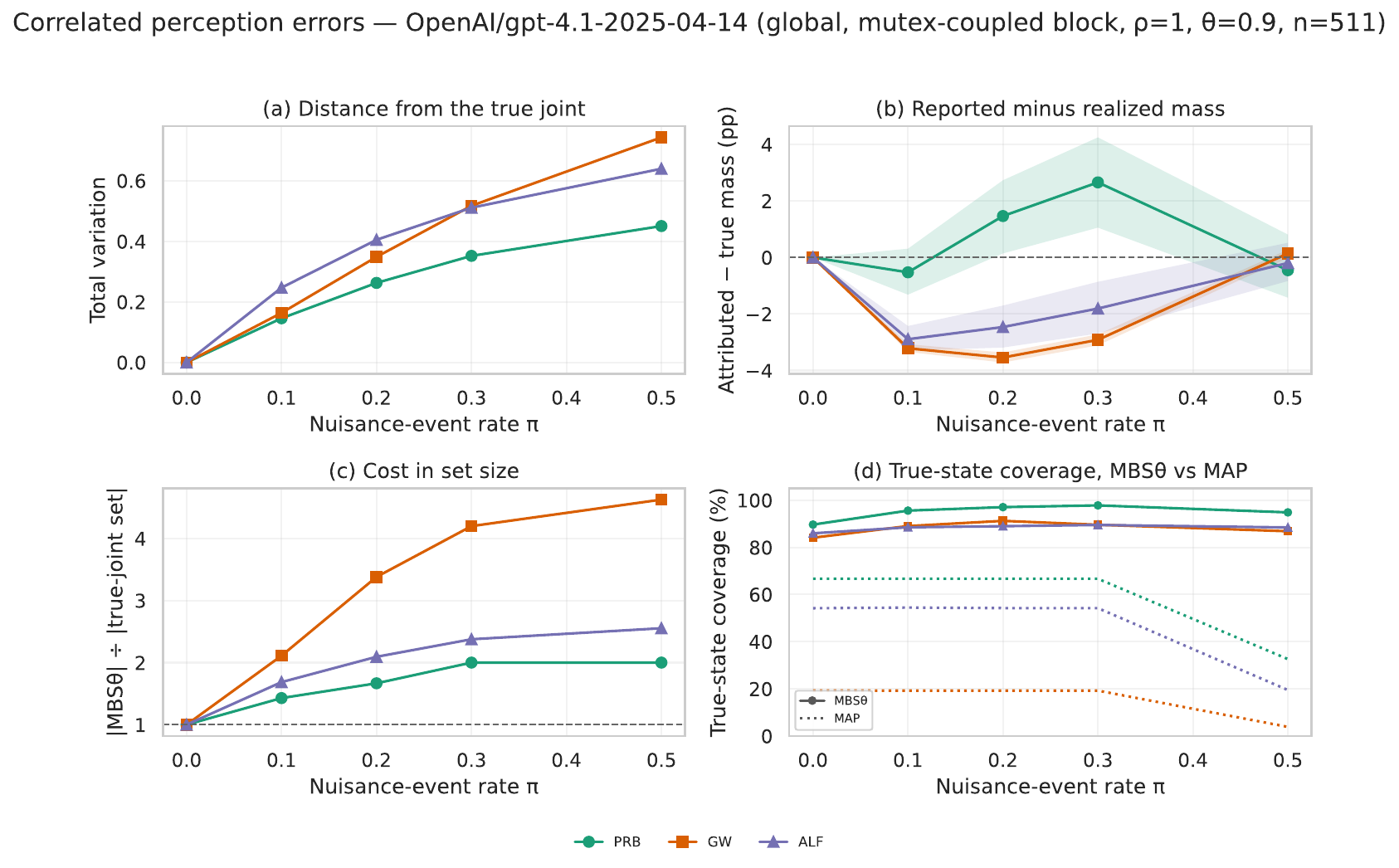}
\caption{Correlated perception errors with the mutex-coupled block, $\rho=1$, $\probThresh=0.9$, and GPT-4.1 under global Platt calibration, as a function of the nuisance-event rate $\pi$. (a) Total variation between the product belief and the correlated joint. (b) Attributed minus true mass of the returned \gls{mbs} in percentage points, with 95\% bootstrap confidence bands. (c) Size of the product-belief \gls{mbs} relative to the true joint's own minimal \probThresh-mass set. (d) Ground-truth coverage of \gls{mbs} (solid) and \gls{map} (dotted).}
\label{fig:correlated}
\end{figure}

\paragraph{Results.}
\cref{fig:correlated} and \cref{tab:correlated-distortion,tab:correlated-coverage,tab:correlated-blocks} summarize the sweep.
Three findings follow.
First, even when corruption drives the total variation between the product belief and the true joint up to 0.74, the gap between attributed and true set mass stays within 4.4 points across all domains, corruption rates, and block choices, and its sign varies across domains and block modes.
The true mass can fall below the requested \probThresh in the hardest cells, but the shortfall tracks this small gap rather than collapsing.
Second, the cost of the independence assumption is set size rather than coverage: the product-belief set grows up to 4.6 times the certified true-joint set at the same mass target.
Third, ground-truth coverage of the product-belief \gls{mbs} stays between 84\% and 98\% across the entire sweep, while \gls{map} and valid \gls{map} collapse to as low as 3.8\% under heavy corruption.
Under exactly this kind of correlated error, determinization collapses and the belief set absorbs the misspecification by growing, which is the design intent.

\paragraph{Scope.}
This is a synthetic sensitivity analysis, not an estimate of the natural error correlation of any \gls{vlm}: $\pi$ and $\rho$ are swept rather than estimated from data, and the nuisance is a single shared binary event over one block.
Learned inter-feature dependencies and distribution-free calibration of \probThresh against coverage remain future work.

\begin{table}[t]
\centering
\small
\caption{Distortion of the product belief under correlated perception errors (mutex-coupled block, $\rho=1$, $\probThresh=0.9$, GPT-4.1). $n$ is the number of paired instances, total variation is the distance between the product belief and the correlated joint, the mass gap is the attributed minus true mass of the returned \gls{mbs} in percentage points with 95\% bootstrap confidence intervals, Reaches \probThresh is the fraction of instances whose true mass still meets \probThresh, and the last column is the size of the product-belief \gls{mbs} relative to the true joint's own minimal \probThresh-mass set.}
\label{tab:correlated-distortion}
\begin{tabular}{@{}lcccccc@{}}
\toprule
Domain & $\pi$ & $n$ & Total variation & Mass gap (pp) & Reaches \probThresh & Size ratio \\
\midrule
PRB & 0.00 & 135 & 0.000 & $0.0$ [$0.0$, $0.0$] & 100.0\% & 1.00$\times$ \\
 & 0.10 & 135 & 0.146 & $-0.5$ [$-1.3$, $+0.3$] & 87.4\% & 1.43$\times$ \\
 & 0.20 & 135 & 0.263 & $+1.5$ [$+0.1$, $+2.7$] & 67.4\% & 1.67$\times$ \\
 & 0.30 & 135 & 0.352 & $+2.7$ [$+1.1$, $+4.2$] & 75.6\% & 2.00$\times$ \\
 & 0.50 & 135 & 0.450 & $-0.5$ [$-1.4$, $+0.8$] & 85.2\% & 2.00$\times$ \\
\midrule
GW & 0.00 & 182 & 0.000 & $0.0$ [$0.0$, $0.0$] & 100.0\% & 1.00$\times$ \\
 & 0.10 & 182 & 0.164 & $-3.2$ [$-3.3$, $-3.1$] & 100.0\% & 2.11$\times$ \\
 & 0.20 & 182 & 0.349 & $-3.5$ [$-3.7$, $-3.4$] & 100.0\% & 3.38$\times$ \\
 & 0.30 & 182 & 0.517 & $-2.9$ [$-3.1$, $-2.7$] & 100.0\% & 4.20$\times$ \\
 & 0.50 & 182 & 0.743 & $+0.1$ [$-0.0$, $+0.3$] & 50.0\% & 4.62$\times$ \\
\midrule
ALF & 0.00 & 192 & 0.000 & $0.0$ [$0.0$, $0.0$] & 100.0\% & 1.00$\times$ \\
 & 0.10 & 191 & 0.247 & $-2.9$ [$-3.3$, $-2.4$] & 93.2\% & 1.69$\times$ \\
 & 0.20 & 190 & 0.405 & $-2.5$ [$-3.2$, $-1.7$] & 88.4\% & 2.09$\times$ \\
 & 0.30 & 190 & 0.511 & $-1.8$ [$-2.7$, $-0.9$] & 88.9\% & 2.38$\times$ \\
 & 0.50 & 190 & 0.640 & $-0.2$ [$-0.8$, $+0.5$] & 81.6\% & 2.55$\times$ \\
\bottomrule
\end{tabular}
\end{table}

\begin{table}[t]
\centering
\small
\caption{Ground-truth coverage under correlated perception errors (mutex-coupled block, $\rho=1$, $\probThresh=0.9$, GPT-4.1), with the median returned set size in parentheses. The true-joint column is the oracle that knows the correlated joint.}
\label{tab:correlated-coverage}
\resizebox{\textwidth}{!}{%
\begin{tabular}{@{}lcccccc@{}}
\toprule
Domain & $\pi$ & $n$ & \gls{mbs} (product belief) & \gls{mbs} (true joint) & Valid \gls{map} & \gls{map} \\
\midrule
PRB & 0.00 & 135 & 89.6\% (2) & 89.6\% (2) & 67.4\% (1) & 66.7\% (1) \\
 & 0.10 & 135 & 95.6\% (4) & 91.1\% (3) & 67.4\% (1) & 66.7\% (1) \\
 & 0.20 & 135 & 97.0\% (5) & 91.1\% (3) & 67.4\% (1) & 66.7\% (1) \\
 & 0.30 & 135 & 97.8\% (6) & 91.1\% (3) & 67.4\% (1) & 66.7\% (1) \\
 & 0.50 & 135 & 94.8\% (8) & 91.9\% (3) & 32.6\% (1) & 32.6\% (1) \\
\midrule
GW & 0.00 & 182 & 84.1\% (22) & 84.1\% (22) & 25.8\% (1) & 19.2\% (1) \\
 & 0.10 & 182 & 89.0\% (43) & 84.1\% (22) & 25.8\% (1) & 19.2\% (1) \\
 & 0.20 & 182 & 91.2\% (68.5) & 84.1\% (22) & 25.8\% (1) & 19.2\% (1) \\
 & 0.30 & 182 & 89.6\% (76.5) & 84.1\% (22) & 25.8\% (1) & 19.2\% (1) \\
 & 0.50 & 182 & 86.8\% (88) & 84.1\% (22) & 3.8\% (1) & 3.8\% (1) \\
\midrule
ALF & 0.00 & 192 & 85.9\% (43) & 85.9\% (43) & 54.2\% (1) & 54.2\% (1) \\
 & 0.10 & 191 & 88.5\% (76) & 86.9\% (44) & 54.5\% (1) & 54.5\% (1) \\
 & 0.20 & 190 & 88.9\% (94) & 86.8\% (42.5) & 54.2\% (1) & 54.2\% (1) \\
 & 0.30 & 190 & 89.5\% (118.5) & 86.3\% (43) & 54.2\% (1) & 54.2\% (1) \\
 & 0.50 & 190 & 88.4\% (139) & 86.3\% (46) & 19.5\% (1) & 19.5\% (1) \\
\bottomrule
\end{tabular}%
}
\end{table}

\begin{table}[t]
\centering
\small
\caption{Mass gap (attributed minus true mass of the returned \gls{mbs}, percentage points, with 95\% bootstrap confidence intervals) and fraction of instances whose true mass reaches \probThresh, for the two corruption block modes ($\rho=1$, $\probThresh=0.9$, GPT-4.1).}
\label{tab:correlated-blocks}
\begin{tabular}{@{}lccccc@{}}
\toprule
 & & \multicolumn{2}{c}{Mutex-coupled block} & \multicolumn{2}{c}{Least-decided block} \\
\cmidrule(lr){3-4} \cmidrule(lr){5-6}
Domain & $\pi$ & Mass gap (pp) & Reaches \probThresh & Mass gap (pp) & Reaches \probThresh \\
\midrule
PRB & 0.10 & $-0.5$ [$-1.3$, $+0.3$] & 87\% & $+0.8$ [$-0.1$, $+1.8$] & 74\% \\
 & 0.20 & $+1.5$ [$+0.1$, $+2.7$] & 67\% & $+3.8$ [$+2.4$, $+5.2$] & 55\% \\
 & 0.30 & $+2.7$ [$+1.1$, $+4.2$] & 76\% & $+4.4$ [$+2.8$, $+6.4$] & 65\% \\
 & 0.50 & $-0.5$ [$-1.4$, $+0.8$] & 85\% & $-0.5$ [$-1.8$, $+1.1$] & 93\% \\
\midrule
GW & 0.10 & $-3.2$ [$-3.3$, $-3.1$] & 100\% & $-1.5$ [$-1.9$, $-1.0$] & 90\% \\
 & 0.20 & $-3.5$ [$-3.7$, $-3.4$] & 100\% & $-2.2$ [$-2.5$, $-1.7$] & 87\% \\
 & 0.30 & $-2.9$ [$-3.1$, $-2.7$] & 100\% & $-2.3$ [$-2.7$, $-1.9$] & 88\% \\
 & 0.50 & $+0.1$ [$-0.0$, $+0.3$] & 50\% & $-1.7$ [$-2.0$, $-1.3$] & 84\% \\
\midrule
ALF & 0.10 & $-2.9$ [$-3.3$, $-2.4$] & 93\% & $+1.2$ [$+0.7$, $+1.8$] & 30\% \\
 & 0.20 & $-2.5$ [$-3.2$, $-1.7$] & 88\% & $+3.8$ [$+2.9$, $+4.6$] & 27\% \\
 & 0.30 & $-1.8$ [$-2.7$, $-0.9$] & 89\% & $+2.3$ [$+1.2$, $+3.4$] & 51\% \\
 & 0.50 & $-0.2$ [$-0.8$, $+0.5$] & 82\% & $-0.1$ [$-0.9$, $+0.7$] & 65\% \\
\bottomrule
\end{tabular}
\end{table}

\FloatBarrier
\section{Computational resources used}\label{ap:compute}

\subsection{Worker type and hardware specifications}
\begin{itemize}
    \item Architecture \& CPU: Internal HPC cluster using x86\_64 architecture with dual-socket Intel(R) Xeon(R) 6710E CPUs (128 total cores per node, 64 cores per socket, 1 thread per core).
    \item Memory: 1 TB (1031821 MB) of total RAM available per compute node.
    \item Storage: 438 GB local NVMe storage (/dev/nvme0n1p2) per node, backed by network-attached storage (e.g., Vast Data NFS).
\end{itemize}

\subsection{Compute and memory per experiment}
\begin{itemize}
    \item Execution Time: Individual CPU experiment runs ranged from less than a minute to a maximum wall-clock time of 10 hours.
    \item Memory Utilization: Peak memory usage (MaxRSS) varied significantly by job, ranging from 3.5 MB up to a maximum of 226.8 GB (221460920K).
    \item Open-weight \gls{vlm} inference (\cref{ap:backbones}) is the only GPU workload. Each of the five open-weight backbones was run once over all evaluation splits as a single job allocated four NVIDIA L40 GPUs (48 GB each) on the same cluster, with the model loaded in bfloat16. Active inference wall-clock time per backbone ranged from roughly 30 hours (Qwen3-VL-8B) to 80 hours (Gemma 4 31B), about 270 hours in total, estimated from output timestamps. All subsequent calibration, \gls{mbs} construction, and planning experiments consume the cached predictions and run on CPU only, as for GPT-4.1.
\end{itemize}

\section{Full runtime results}\label{ap:runtime-full}

\begin{figure}[t]
    \centering
    \begin{minipage}[t][][b]{0.49\textwidth}
        \includegraphics[width=1\linewidth]{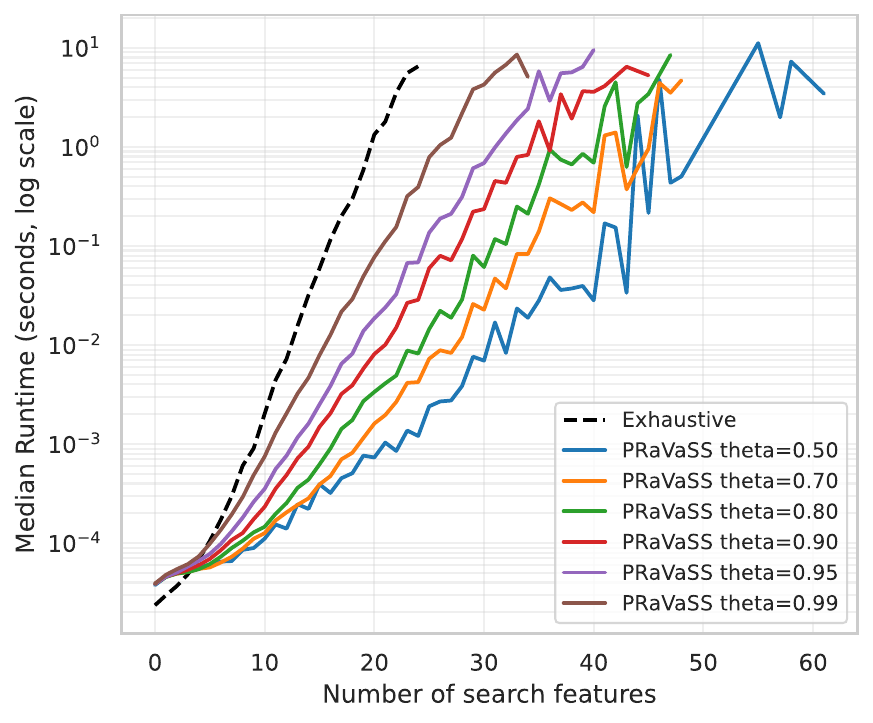}
        \subcaption{Runtime vs.\ number of search features.}
        \label{fig:scaling:sfc-time}
    \end{minipage}\hfill
    \begin{minipage}[t][][b]{0.49\textwidth}
        \includegraphics[width=1\linewidth]{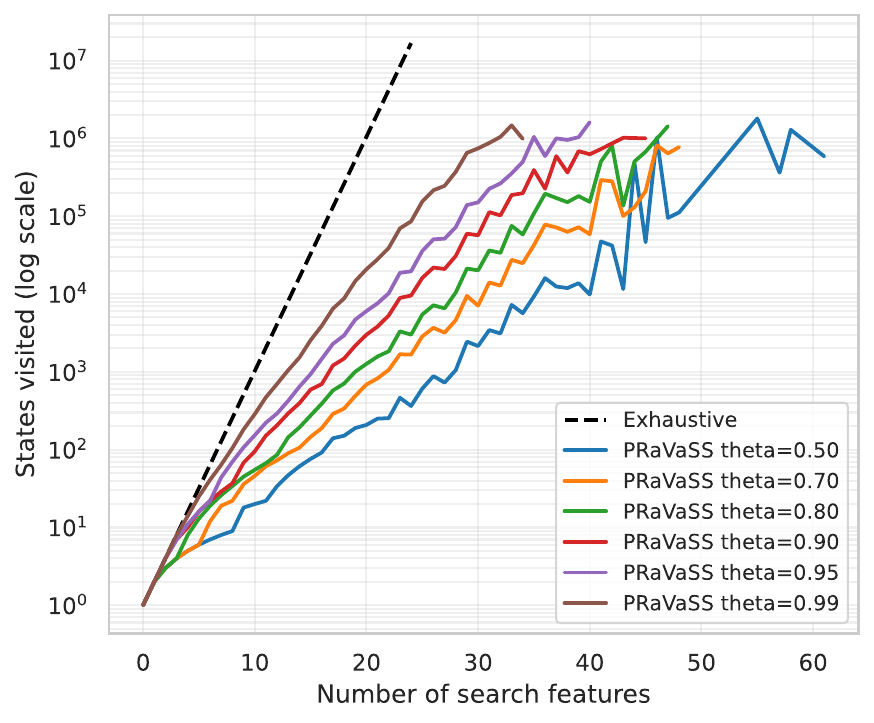}
        \subcaption{States visited vs.\ number of search features.}
        \label{fig:scaling:sfc-mem}
    \end{minipage}\hfill
    \caption{Scaling behavior of \gls{pravas} across test instances and \probThresh values. (a) Runtime scaling as a function of the number of search features for PRB and ALF. (b) Memory scaling as a function of the number of search features for PRB and ALF. GW is omitted because it has a fixed number of features.}
    \label{fig:scaling}
\end{figure}

\cref{fig:scaling:sfc-time} and \cref{fig:scaling:sfc-mem} show the runtime and memory scaling of \gls{pravas} as a function of the number of features considered in the search, for the PRB and ALF domains.
The GW domain was omitted because it has a fixed number of features, and thus does not provide insight into feature-count scaling behavior.
The runtime plot shows the growing gap between \gls{pravas} and exhaustive enumeration as the number of search features increases, with \gls{pravas} scaling much more gracefully.
At $\probThresh=0.99$, \gls{pravas} can find the \gls{mbs} in problems with 42\% more features than exhaustive enumeration, meaning that \gls{pravas} can scale significantly just by forgoing 0.01 probability mass.
This advantage grows up to 154\% at $\probThresh=0.5$, scaling significantly as we place more trust in the \gls{sse}'s output.
Similarly, the memory plot shows that \gls{pravas} maintains significantly fewer states than exhaustive enumeration, and thus uses much less memory.
The gap is significant even at the highest-coverage operating point of $\probThresh=0.99$, showing that there is a big mass of low-probability states that \gls{pravas} can safely ignore to achieve significant efficiency gains.

\cref{tab:scaling} reports runtime, memory, and scaling behavior of \gls{pravas} over different \probThresh values compared to exhaustive enumeration across all domains. Results show that exhaustive enumeration was unable to scale beyond 24 search features.

\begin{table}[t]
\centering
\small
\caption{Runtime and scaling behavior of \gls{pravas} and exhaustive enumeration. Solved entries report counts with percentages in parentheses; bracketed values are 95\% bootstrap confidence intervals using 10,000 paired resamples over test instances.}
\label{tab:scaling}
\resizebox{\columnwidth}{!}{%
\begin{tabular}{@{}lcccc@{}}
\toprule
$\theta$ & \shortstack{Solved\\instances} & \shortstack{Max search\\features solved} & \shortstack{Median\\runtime (s)} & \shortstack{Median states\\visited} \\
\midrule
0.50 & 3{,}789 (99.71\%) & 61 & $1.56{\times}10^{-4}$ & 16.00 \\
  &  &  & $[1.43{\times}10^{-4}, 1.67{\times}10^{-4}]$ & $[12.48, 18.00]$ \\
 0.70 & 3{,}779 (99.45\%) & 48 & $1.14{\times}10^{-4}$ & 36.00 \\
  &  &  & $[1.02{\times}10^{-4}, 1.24{\times}10^{-4}]$ & $[31.00, 43.52]$ \\
 0.80 & 3{,}774 (99.32\%) & 47 & $1.34{\times}10^{-4}$ & 45.00 \\
  &  &  & $[1.20{\times}10^{-4}, 1.47{\times}10^{-4}]$ & $[43.00, 55.00]$ \\
 0.90 & 3{,}756 (98.84\%) & 45 & $1.80{\times}10^{-4}$ & 68.00 \\
  &  &  & $[1.49{\times}10^{-4}, 2.17{\times}10^{-4}]$ & $[55.00, 86.26]$ \\
 0.95 & 3{,}737 (98.34\%) & 40 & $2.64{\times}10^{-4}$ & 106.00 \\
  &  &  & $[2.15{\times}10^{-4}, 3.09{\times}10^{-4}]$ & $[82.00, 121.00]$ \\
 0.99 & 3{,}674 (96.68\%) & 34 & $4.32{\times}10^{-4}$ & 159.00 \\
  &  &  & $[3.38{\times}10^{-4}, 5.19{\times}10^{-4}]$ & $[122.00, 186.00]$ \\
\midrule
Exhaustive & 2{,}906 (76.47\%) & 24 & $1.66{\times}10^{-4}$ & 64.00 \\
  &  &  & $[1.31{\times}10^{-4}, 2.04{\times}10^{-4}]$ & $[32.00, 64.00]$ \\
\bottomrule
\end{tabular}%
}
\end{table}

\FloatBarrier
\section{Extended results for Experiment 3}\label{ap:ex3-full}

Experiment 3 used an iterative approach that we call \gls{ams} to find the largest \probThresh for which a robust plan can be found from an \gls{mbs} produced by \gls{pravas}.
First, we run \gls{pravas} with a large \probThresh to find a large \gls{mbs}, where states are sorted by their probability mass in descending order.
This is a side effect of how \gls{pravas} expands different states throughout its search.
With this large \gls{mbs}, we perform a binary search for the largest prefix of the sorted \gls{mbs} for which the CPOR planner finds some plan.
The probability mass of the resulting \gls{mbs} is used as a confidence score for plan success.
Similarly, adaptive abstention performs a binary search for an abstention threshold \absThresh for which CPOR finds a plan from the feature-abstention \gls{belSet}.

The \gls{ams} variant reported in \cref{tab:planning} uses a more comprehensive set of constraints than in previous experiments.
These constraints are not disjoint mutex groups, and thus an exact normalizer requires weighted model counting, which is exponential in the induced width in the worst case (\cref{ap:constraints:norm}).
Instead, we use a relaxed normalizer as suggested in \cref{sec:belief:comp}.
A comparison with \gls{ams} using looser restrictions is available in \cref{tab:planning-full}.
In the GW domain, \gls{ams}-Loose achieves a 3.5\% improvement over \gls{map}, and matches adaptive abstention.
In the PRB domain, \gls{ams}-Loose performs 3\% worse than \gls{map} and 3.6\% worse than adaptive abstention.
\gls{ams}, however, outperforms all other methods in both domains, achieving a 13\% improvement over \gls{map} in GW, and a 95\% improvement in PRB.
This shows the power of constraint-based belief set construction and the versatility of \gls{pravas}, even in the presence of approximated probability masses obtained with a relaxed normalizer.

\begin{table}[t]
\centering
\small
\caption{Downstream planning success rate in the PRB and GW domains. Bracketed rows are 95\% bootstrap confidence intervals using 10,000 paired resamples over test instances. Bold values indicate the best-performing method in each domain, underlined values indicate the second-best.}
\label{tab:planning-full}
\begin{tabular}{@{}lcc@{}}
\toprule
Method & PRB & GW \\
\midrule
MAP & 16.7\% & 70.7\% \\
 & {\scriptsize [14.4\%, 18.8\%]} & {\scriptsize [67.2\%, 73.8\%]} \\
\addlinespace[0.2em]
Adaptive Abst. & \underline{16.8\%} & \underline{73.3\%} \\
 & {\scriptsize [14.7\%, 19.1\%]} & {\scriptsize [69.8\%, 76.8\%]} \\
\addlinespace[0.2em]
\midrule
\gls{ams}-Loose & 16.2\% & 73.2\% \\
 & {\scriptsize [14.1\%, 18.0\%]} & {\scriptsize [69.5\%, 76.3\%]} \\
\addlinespace[0.2em]
\gls{ams} & \textbf{32.6\%} & \textbf{80.0\%} \\
 & {\scriptsize [30.1\%, 35.3\%]} & {\scriptsize [76.7\%, 83.0\%]} \\
\bottomrule
\end{tabular}
\end{table}

\paragraph{Constraint-strength and normalizer ablation.}
To separate the belief-set construction from the constraint information, \cref{tab:ablation} runs the identical \gls{ams} pipeline with the loose (disjoint selected groups) and strict (full) constraint sets of \cref{ap:constraints}, the latter under both the relaxed and the exact variable-elimination normalizer, alongside \gls{map} and adaptive abstention fed the same calibrated marginals and the same strict constraints.
The rows come from a single run of the Experiment~3 pipeline covering all backbones of \cref{ap:backbones}, so the GPT-4.1 values differ marginally from those of \cref{tab:planning}.
\cref{tab:ablation-paired} reports paired bootstrap contrasts over matched instances using 10,000 resamples.
Every method except \gls{ams}-Loose receives the same constraint information, yet only the belief-set construction converts it into a large gain: with identical strict constraints, \gls{ams} exceeds adaptive abstention by 15.7 points on PRB and 5.2 points on GW, and exceeds its own loose-constraint variant by 16.2 and 4.5 points.
The exact normalizer shifts success by at most 0.7 points, an order of magnitude below the constraint-layer effect, which confirms that the prefix search consumes only the ordering of states, which the relaxed normalizer preserves while its reported masses remain lower bounds (\cref{cor:relaxed}).
Valid \gls{map} in \cref{ap:backbones}, i.e., \gls{map} with the same strict constraints, likewise does not close the coverage gap to \gls{mbs}.
The constraints are extracted automatically by \gls{fd} for PRB and GW, so the strict sets are not privileged hand-crafted information.

\begin{table}[t]
\centering
\small
\caption{Constraint-strength and normalizer ablation for GPT-4.1: downstream planning success rate with 95\% bootstrap confidence intervals over the 1,200 PRB and 600 GW test instances. All methods receive the same calibrated marginals and, except \gls{ams}-Loose, the same strict constraint set. TO is the fraction of instances on which the method exhausted the planning budget, counted as failures.}
\label{tab:ablation}
\begin{tabular}{@{}lcccc@{}}
\toprule
 & \multicolumn{2}{c}{PRB} & \multicolumn{2}{c}{GW} \\
\cmidrule(lr){2-3} \cmidrule(lr){4-5}
Method & Success [95\% CI] & TO & Success [95\% CI] & TO \\
\midrule
\gls{map} & 16.7\% [14.7, 18.9] & 0\% & 71.7\% [67.9, 75.1] & 0\% \\
Adaptive Abst. & 16.8\% [14.7, 19.0] & 0\% & 74.0\% [70.3, 77.3] & 0\% \\
\gls{ams}-Loose & 16.2\% [14.2, 18.4] & 0\% & 74.7\% [71.0, 78.0] & 1\% \\
\gls{ams} (strict, relaxed \normalizer) & 32.4\% [29.8, 35.1] & 0\% & 79.2\% [75.7, 82.2] & 2\% \\
\gls{ams} (strict, exact \normalizer) & 32.3\% [29.7, 35.0] & 0\% & 79.8\% [76.4, 82.8] & 0\% \\
\bottomrule
\end{tabular}
\end{table}

\begin{table}[t]
\centering
\small
\caption{Paired contrasts for the ablation of \cref{tab:ablation}: difference in planning success (percentage points) on matched test instances, with 95\% paired bootstrap confidence intervals using 10,000 resamples.}
\label{tab:ablation-paired}
\begin{tabular}{@{}llcc@{}}
\toprule
Domain & Contrast & Difference & 95\% CI \\
\midrule
PRB & \gls{ams} (strict, relaxed \normalizer) $-$ \gls{map} & $+$15.8 & [$+$13.7, $+$17.9] \\
 & \gls{ams} (strict, relaxed \normalizer) $-$ Adaptive Abst. & $+$15.7 & [$+$13.6, $+$17.8] \\
 & \gls{ams} (strict, relaxed \normalizer) $-$ \gls{ams}-Loose & $+$16.2 & [$+$14.2, $+$18.4] \\
 & \gls{ams} (strict, exact \normalizer) $-$ \gls{ams} (strict, relaxed \normalizer) & $-$0.1 & [$-$0.2, $+$0.0] \\
\midrule
GW & \gls{ams} (strict, relaxed \normalizer) $-$ \gls{map} & $+$7.5 & [$+$5.3, $+$9.8] \\
 & \gls{ams} (strict, relaxed \normalizer) $-$ Adaptive Abst. & $+$5.2 & [$+$3.2, $+$7.3] \\
 & \gls{ams} (strict, relaxed \normalizer) $-$ \gls{ams}-Loose & $+$4.5 & [$+$2.7, $+$6.5] \\
 & \gls{ams} (strict, exact \normalizer) $-$ \gls{ams} (strict, relaxed \normalizer) & $+$0.7 & [$+$0.2, $+$1.3] \\
\bottomrule
\end{tabular}
\end{table}

\begin{figure}
    \centering
    \includegraphics[width=1\textwidth]{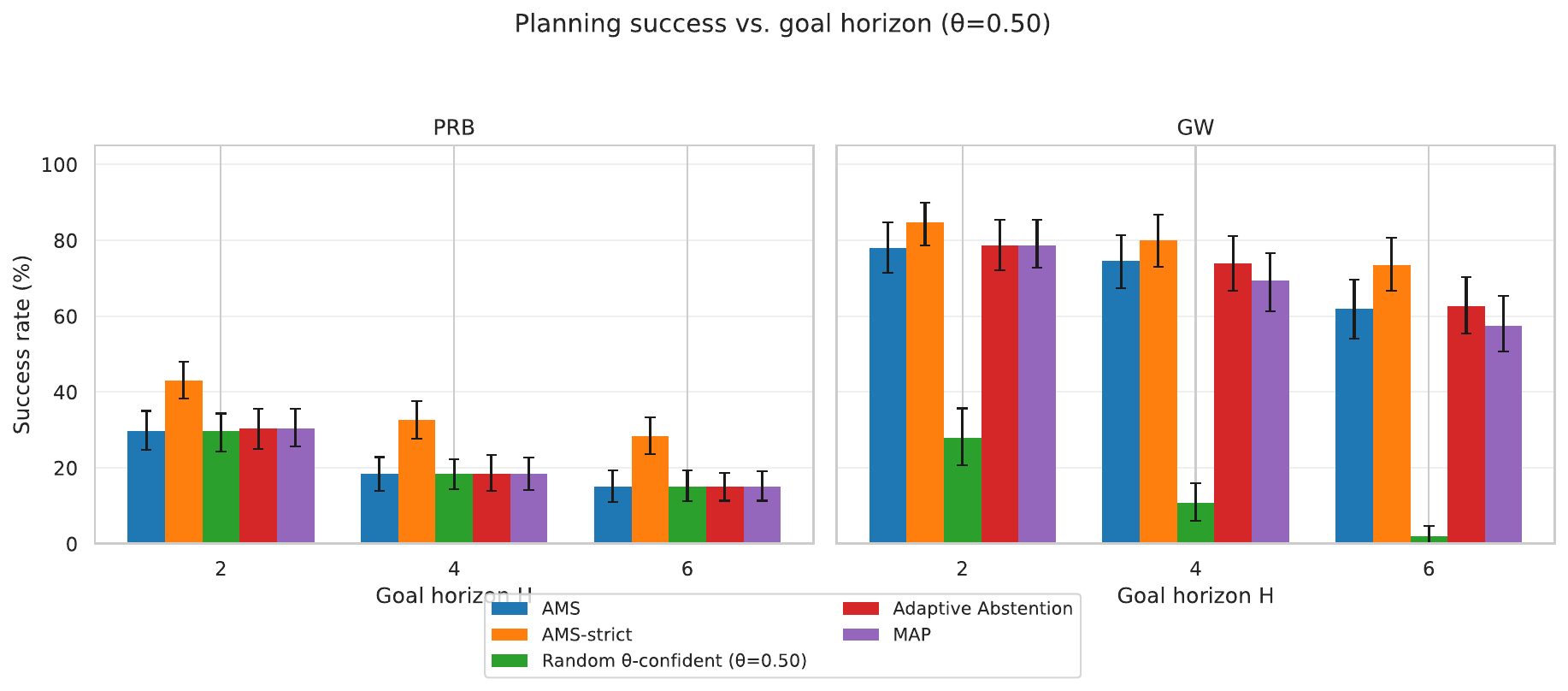}
    \caption{Downstream planning success rate breakdown by horizon in the PRB and GW domains. Error bars are 95\% bootstrap confidence intervals using 10,000 paired resamples over test instances. In the legend, AMS-strict is \gls{ams}, AMS is \gls{ams}-Loose, and Random $\theta$-confident is Random \probThresh-Mass.}
    \label{fig:plan-horizon}
\end{figure}

\cref{fig:plan-horizon} breaks down the results seen in \cref{tab:planning} by horizon, showing the downstream planning success rate for each method as a function of the horizon of the test instance.

\begin{figure}
\centering
\includegraphics[width=1\textwidth]{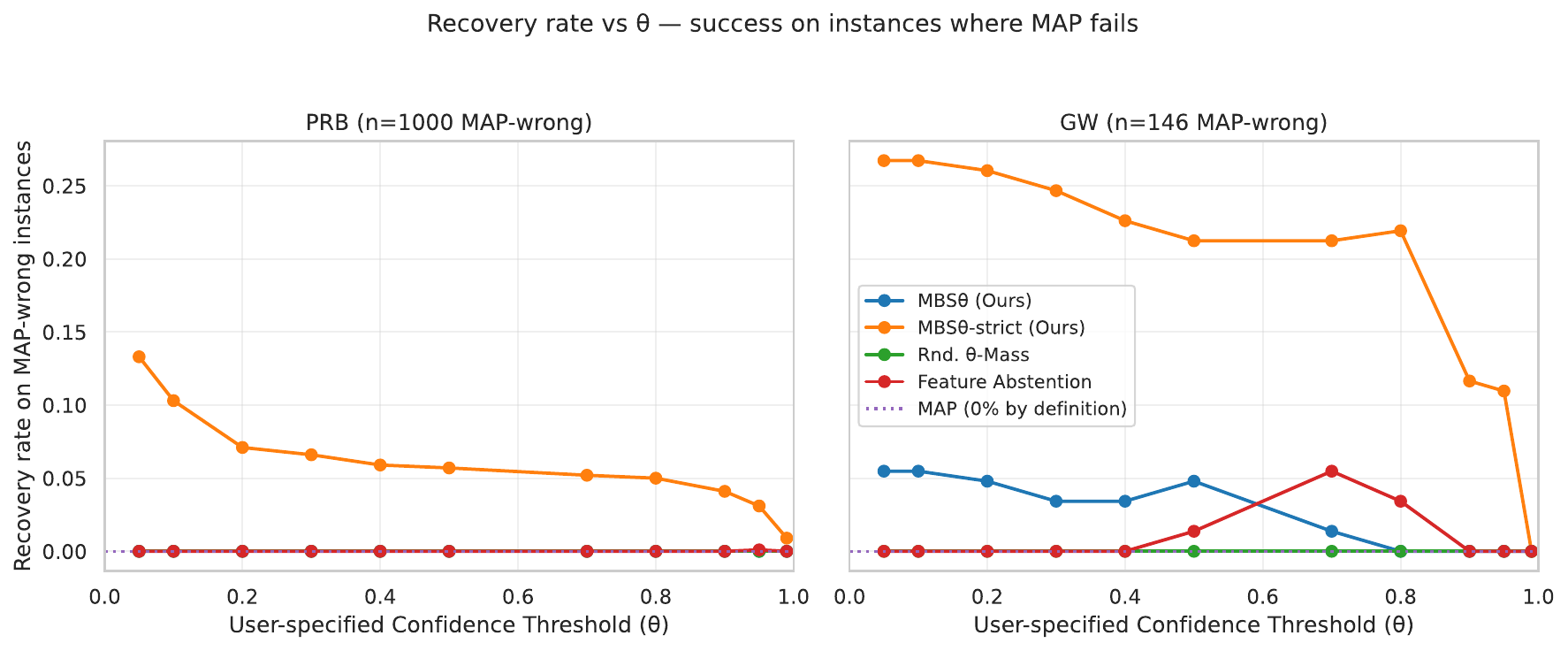}
\caption{Recovery performance in the PRB and GW domains as a function of \probThresh. In the legend, \gls{mbs}-strict is \gls{mbs} and \gls{mbs} is \gls{mbs}-loose of \cref{tab:retention}.}
\label{fig:retention}
\end{figure}

\cref{tab:retention} shows how much performance is retained on solvable test instances where the most likely state is the true state (MAP-correct) vs.\ test instances where the most likely state is not the true state (MAP-wrong), for each method in the PRB and GW domains.
We report the success rate of each compared method on the MAP-correct and MAP-wrong strata, which we refer to as the \textit{retention} and \textit{recovery} performance of each method, respectively.
In this table, we show performance for $\probThresh=0.5$, which corresponds to the \gls{belSet} containing the majority of the probability mass. \cref{fig:retention} shows the recovery performance of each method as a function of $\probThresh$.

\begin{sidewaystable}[p]
\centering
\small
\caption{Retention and recovery performance in the PRB and GW domains for $\probThresh=0.5$. Bracketed values are 95\% bootstrap confidence intervals using 10,000 paired resamples over test instances.}
\label{tab:retention}
\resizebox{\columnwidth}{!}{
\begin{tabular}{lllll}
\toprule
Domain & \multicolumn{2}{r}{PRB} & \multicolumn{2}{r}{GW} \\
Stratum & MAP-correct (Retention) & MAP-wrong (Recovery) & MAP-correct (Retention) & MAP-wrong (Recovery) \\
\midrule
(stratum size) & n=200 & n=1000 & n=424 & n=146 \\
\gls{mbs}-loose (Ours) & 97.0\% [94.7\%, 99.0\%] & 0.0\% [0.0\%, 0.0\%] & 90.8\% [87.4\%, 93.4\%] & 4.8\% [1.4\%, 8.2\%] \\
\gls{mbs} (Ours) & 97.5\% [95.5\%, 99.3\%] & 5.7\% [4.4\%, 7.4\%] & 98.6\% [97.4\%, 99.5\%] & 21.2\% [15.1\%, 28.8\%] \\
Rnd. $\theta$-Mass & 95.5\% [92.5\%, 98.0\%] & 0.0\% [0.0\%, 0.0\%] & 14.4\% [11.2\%, 17.7\%] & 0.0\% [0.0\%, 0.0\%] \\
Feature Abstention & 100.0\% [100.0\%, 100.0\%] & 0.0\% [0.0\%, 0.0\%] & 98.8\% [97.6\%, 99.8\%] & 1.4\% [0.0\%, 3.4\%] \\
MAP & 100.0\% [100.0\%, 100.0\%] & 0.0\% [0.0\%, 0.0\%] & 100.0\% [100.0\%, 100.0\%] & 0.0\% [0.0\%, 0.0\%] \\
\bottomrule
\end{tabular}
}
\end{sidewaystable}

\FloatBarrier
\section{Closed-loop planning with hidden features}\label{ap:closed-loop}

The claims of this paper concern single-shot \gls{belSet} construction, and ALF is excluded from Experiment 3 because robust decision-making there requires the agent to gather information.
To test whether the advantage of \gls{ams} carries over to a closed loop, we ran an observe-update-replan study on ALF with GPT-4.1 as the backbone.

\paragraph{Protocol.}
Of the 2,000 ALF test tasks, 417 have between two and six declared non-visible, goal- or action-relevant uncertainty units with a constraint-valid alternative.
A unit is either a Boolean feature or a categorical group that keeps its exactly-one constraint, such as the location of an object.
We froze a seeded sample of 120 of them, of which 5 were excluded for every method because the closed-world initial state contained no true member of a declared location group, leaving 115 paired tasks.
Every atom in the declared units is set to probability 0.5 and all other features are clamped to the same constraint-valid \gls{map} assignment for every method, so the belief is a uniform prior over the legal completions of the hidden units.
Each method plans (round $k=0$), one shared relevant unit selected by a seeded uniform policy is then revealed noiselessly, the belief is updated for that unit only, the common constrained state ordering is recomputed, and each method replans ($k=1$).
Constraint-valid \gls{map} (hereafter \gls{map}), \gls{ams}, and adaptive abstention receive the identical uncertainty vocabulary, constraints, observation, exact normalizer, and 30\,s per-planner-call budget, and an oracle that plans from the true state supplies a ceiling.
Success means that the returned action sequence reaches the goal when executed from the true initial state, and planner failures remain in the denominator.
Integrity checks found no method errors, timeouts, or denominator mismatches, and whenever \gls{map} finds a plan, \gls{ams} and adaptive abstention retain at least that plan.

\begin{figure}[t]
\centering
\includegraphics[trim=0 0 0 50,clip,width=0.9\textwidth]{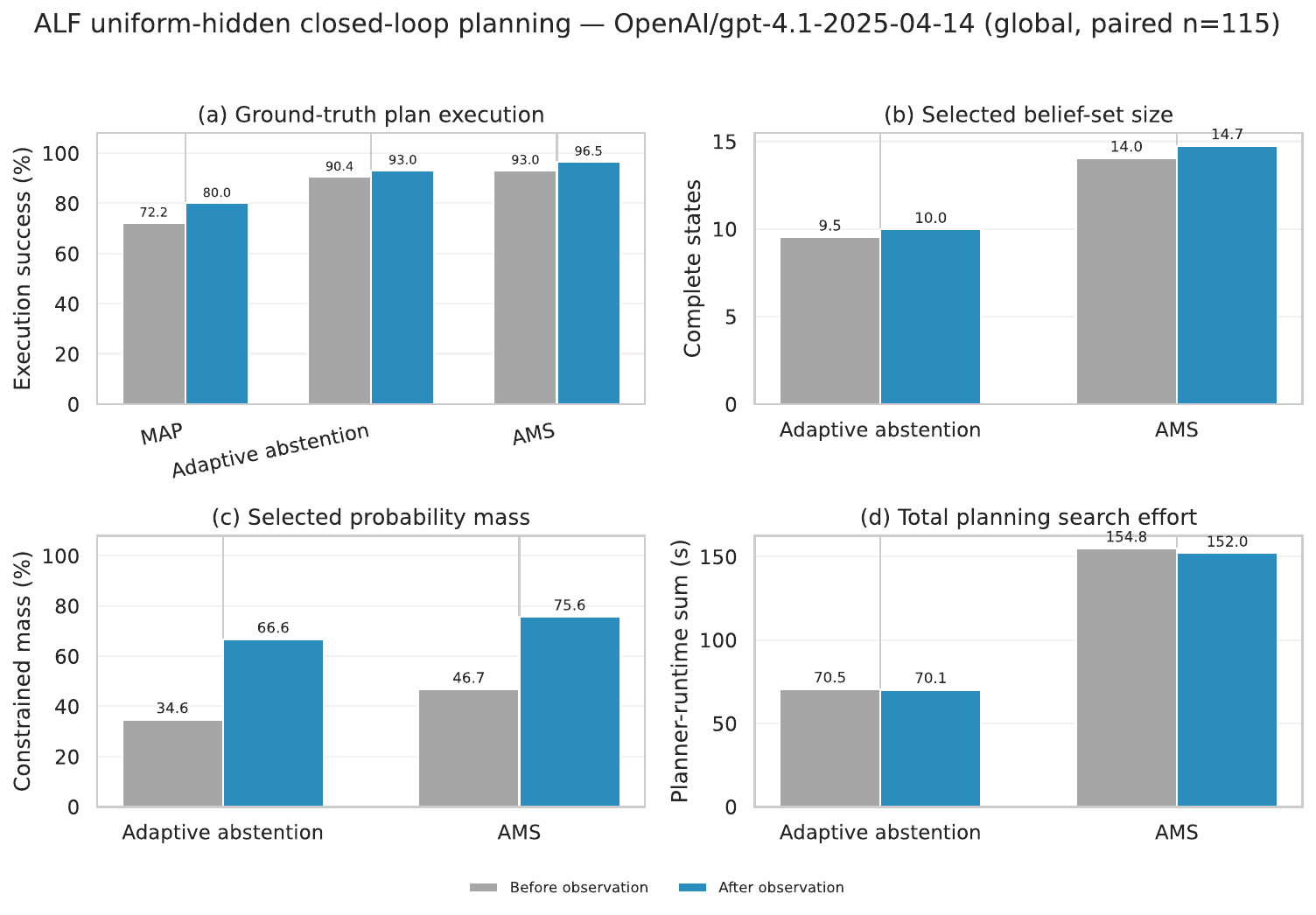}
\caption{Closed-loop planning on ALF with GPT-4.1 under global Platt calibration (115 paired tasks), before and after one shared observation. (a) Ground-truth plan-execution success. (b) Mean number of complete states in the selected planning set. (c) Mean exact constrained mass of that set. (d) Mean sum of planner runtimes over all candidate attempts.}
\label{fig:closed-loop}
\end{figure}

\paragraph{Results.}
\cref{tab:closed-loop,tab:closed-loop-eff} and \cref{fig:closed-loop} report the outcome.
Before the observation, \gls{map} succeeds on 72.2\% of the tasks, adaptive abstention on 90.4\%, \gls{ams} on 93.0\%, and the oracle on 100\%.
After it, these rise to 80.0\%, 93.0\%, 96.5\%, and 100\%.
\gls{ams} exceeds \gls{map} by 20.9 percentage points before the observation (paired 95\% CI $[13.9, 28.7]$) and by 16.5 points after it ($[9.6, 24.3]$).
Its advantage over adaptive abstention is not significant before the observation ($+2.6$ points, $[-0.9, 7.0]$) and becomes significant after it ($+3.5$ points, $[0.9, 7.8]$).
The observation recovers 9 \gls{map}, 4 adaptive-abstention, and 4 \gls{ams} failures, with 0, 1, and 0 regressions, respectively, and it concentrates the belief: the median mass retained by the selected sets rises to approximately 100\% after the observation at essentially unchanged set sizes.
Adaptive abstention offers the better efficiency tradeoff, reaching near-\gls{ams} success with smaller sets and under half the search time, whereas \gls{ams} attains the highest success.

\paragraph{Scope.}
This is a controlled stress test under synthetic maximal uncertainty on the declared hidden units.
It demonstrates a complete observe-update-replan loop with discriminative planning gains, and it does not estimate natural \gls{vlm} uncertainty rates, simulate a physical information-gathering action, or solve many-step active sensing, which we leave to future work.

\begin{table}[t]
\centering
\small
\caption{Closed-loop planning on ALF with GPT-4.1 (115 paired tasks). Top: ground-truth plan-execution success before ($k=0$) and after ($k=1$) one shared observation; the oracle, which plans from the true state, succeeds on all 115 tasks in both rounds. Bottom: paired differences in percentage points with 95\% paired bootstrap confidence intervals.}
\label{tab:closed-loop}
\begin{tabular}{@{}lccc@{}}
\toprule
Method & Before observation & After observation & Change \\
\midrule
\gls{map} & 83/115 (72.2\%) & 92/115 (80.0\%) & $+$7.8 \\
Adaptive Abst. & 104/115 (90.4\%) & 107/115 (93.0\%) & $+$2.6 \\
\gls{ams} & 107/115 (93.0\%) & 111/115 (96.5\%) & $+$3.5 \\
\midrule
Paired contrast & Before observation & After observation & \\
\midrule
\gls{ams} $-$ \gls{map} & $+20.9$ [$+13.9$, $+28.7$] & $+16.5$ [$+9.6$, $+24.3$] & \\
Adaptive Abst. $-$ \gls{map} & $+18.3$ [$+12.2$, $+25.2$] & $+13.0$ [$+7.8$, $+19.1$] & \\
\gls{ams} $-$ Adaptive Abst. & $+2.6$ [$-0.9$, $+7.0$] & $+3.5$ [$+0.9$, $+7.8$] & \\
\bottomrule
\end{tabular}
\end{table}

\begin{table}[t]
\centering
\small
\caption{Efficiency of the set-valued methods in the closed-loop study, mean (median) over the 115 paired tasks. Belief-set size is the number of complete states in the selected planning set, probability mass is the exact constrained mass of that set, search time is the sum of planner runtimes over all candidate attempts including the \gls{map} fallback, and attempts is the mean number of planner calls.}
\label{tab:closed-loop-eff}
\begin{tabular}{@{}llcccc@{}}
\toprule
Round & Method & Belief-set size & Probability mass & Search time & Attempts \\
\midrule
Before observation & \gls{ams} & 14.0 (20) & 46.7\% (45.5\%) & 154.8\,s (186.4\,s) & 6.4 \\
 & Adaptive Abst. & 9.5 (12) & 34.6\% (25.0\%) & 70.5\,s (73.2\,s) & 3.6 \\
\midrule
After observation & \gls{ams} & 14.7 (20) & 75.6\% (100.0\%) & 152.0\,s (174.6\,s) & 6.5 \\
 & Adaptive Abst. & 10.0 (12) & 66.6\% (100.0\%) & 70.1\,s (72.9\,s) & 3.7 \\
\bottomrule
\end{tabular}
\end{table}

\FloatBarrier
\section{Broader impact}\label{ap:impact}

This paper proposes a methodological tool for translating uncertain perception outputs into compact symbolic belief sets that can be consumed by classical reasoners.
Its primary intended societal upside is improved robustness and safety of decision-making systems built on top of pretrained perception, particularly in robotic and embodied settings.
By exposing perceptual ambiguity, the method lets downstream planners avoid actions that are unsafe or infeasible under genuinely ambiguous conditions, supporting the development of more trustworthy neurosymbolic systems.

We see two main avenues for potential negative impact.
First, the method relies on assumptions that may be violated in deployment, including calibrated per-feature probability estimates, conditional independence in the belief converter, and faithful task constraint extraction.
If these assumptions are violated, the resulting belief set may misrepresent the true uncertainty, which could be more harmful than acknowledged uncertainty in safety-critical applications.
Second, formal correctness guarantees on the belief representation may encourage over-trust in downstream pipelines that still depend on the perception model, the constraint specification, and the planner.
Practitioners deploying such systems should validate the calibration of the underlying perception model on the target domain, audit the extracted task constraints, and treat the probability mass parameter \probThresh as a tunable risk budget rather than a guarantee of plan success.

Because the work is evaluated on benchmark and simulated task-planning domains rather than deployed robotic systems, we do not foresee a direct path to immediate harm, but we encourage future deployments to pair this representation with monitoring of perception calibration and validation of task constraints.

\clearpage
\printglossary[title=Terms]
\printglossary[title=Acronyms,type=\acronymtype]

\newpage
\section*{NeurIPS Paper Checklist}

\begin{enumerate}

\item {\bf Claims}
    \item[] Question: Do the main claims made in the abstract and introduction accurately reflect the paper's contributions and scope?
    \item[] Answer: \answerYes{}
    \item[] Justification: The abstract and introduction state the paper's main contributions: the \gls{mbs} representation, the \gls{pravas} algorithm, correctness and complexity guarantees, and empirical evaluation in task-planning domains. These claims are supported by the formal development in Sections 2--3 and the experiments in Section 4.

\item {\bf Limitations}
    \item[] Question: Does the paper discuss the limitations of the work performed by the authors?
    \item[] Answer: \answerYes{}
    \item[] Justification: The paper discusses important limitations and assumptions, including the conditional independence assumption in the belief converter (with a controlled sensitivity study in \cref{ap:correlated}), the distinction between the attributed mass bounded by \probThresh and true-state coverage (\cref{sec:mlss,ap:static-full}), the exponential worst-case search behavior under constraints with the polynomial bound restricted to the unconstrained case, the single-shot scope of the claims (\cref{ap:closed-loop}), and future extensions to temporal refinement, learned dependencies, and closed-loop robotic execution.

\item {\bf Theory assumptions and proofs}
    \item[] Question: For each theoretical result, does the paper provide the full set of assumptions and a complete (and correct) proof?
    \item[] Answer: \answerYes{}
    \item[] Justification: The paper states its theoretical results in Theorem 1, Corollary 1, and Proposition 1, and provides a proof sketch of Theorem 1 in the main text together with full proofs in \cref{ap:proof-mlss,ap:proof-cor,ap:proof-efficiency}.

    \item {\bf Experimental result reproducibility}
    \item[] Question: Does the paper fully disclose all the information needed to reproduce the main experimental results of the paper to the extent that it affects the main claims and/or conclusions of the paper (regardless of whether the code and data are provided or not)?
    \item[] Answer: \answerYes{}
    \item[] Justification: \cref{sec:experiments} describes the domains, baselines, metrics, state estimator, calibration procedure, and constraint extraction, while \cref{ap:domains,ap:data,ap:constraints,ap:static-full,ap:compute,ap:runtime-full} provide additional details on domains, data construction, constraints, compute, and full results. The paper also links the public repository containing all experiment code (\coderepo).

\item {\bf Open access to data and code}
    \item[] Question: Does the paper provide open access to the data and code, with sufficient instructions to faithfully reproduce the main experimental results, as described in supplemental material?
    \item[] Answer: \answerYes{}
    \item[] Justification: All experiment code is publicly available at \coderepo, with step-by-step instructions for reproducing every result, and the final dataset is linked from that repository. The appendices describe the domains, data construction, constraint extraction, and compute resources needed to reproduce the experiments.

\item {\bf Experimental setting/details}
    \item[] Question: Does the paper specify all the training and test details (e.g., data splits, hyperparameters, how they were chosen, type of optimizer) necessary to understand the results?
    \item[] Answer: \answerYes{}
    \item[] Justification: \cref{sec:experiments} specifies the evaluation domains, calibration setup, VLM backbone, constraint extraction procedure, and baseline inputs. \cref{ap:data} provides validation and test split sizes for PRB, GW, and ALF and lists the additional VLM backbones of \cref{ap:backbones}; \cref{ap:correlated,ap:closed-loop} specify the protocols of the sensitivity and closed-loop studies.

\item {\bf Experiment statistical significance}
    \item[] Question: Does the paper report error bars suitably and correctly defined or other appropriate information about the statistical significance of the experiments?
    \item[] Answer: \answerYes{}
    \item[] Justification: \cref{ap:static-full,ap:runtime-full} report 95\% bootstrap confidence intervals for coverage, set size, mass, runtime, and states visited, and \cref{ap:backbones,ap:correlated,ap:ex3-full,ap:closed-loop} report bootstrap confidence intervals, including paired contrasts, for the multi-backbone, sensitivity, ablation, and closed-loop results. These intervals accompany the main empirical comparisons and scaling results.

\item {\bf Experiments compute resources}
    \item[] Question: For each experiment, does the paper provide sufficient information on the computer resources (type of compute workers, memory, time of execution) needed to reproduce the experiments?
    \item[] Answer: \answerYes{}
    \item[] Justification: \cref{ap:compute} reports the compute environment, including the internal HPC cluster, CPU architecture, number of cores, RAM, local storage, execution-time range, memory-utilization range, and the hardware used for open-weight VLM inference. Experiment 2 in \cref{sec:experiments} and \cref{ap:runtime-full} further report runtime and memory-related scaling behavior.
    
\item {\bf Code of ethics}
    \item[] Question: Does the research conducted in the paper conform, in every respect, with the NeurIPS Code of Ethics \url{https://neurips.cc/public/EthicsGuidelines}?
    \item[] Answer: \answerYes{}
    \item[] Justification: The work evaluates algorithms on benchmark or generated task-planning domains and does not report human-subject experiments, crowdsourcing, or collection of sensitive personal data. We are not aware of any aspect of the research that conflicts with the NeurIPS Code of Ethics.

\item {\bf Broader impacts}
    \item[] Question: Does the paper discuss both potential positive societal impacts and negative societal impacts of the work performed?
    \item[] Answer: \answerYes{}
    \item[] Justification: \cref{ap:impact} provides a broader impact statement covering positive impacts (safer task planning under uncertain perception) and negative impacts (risk of over-trust if the assumptions of calibrated perception, conditional independence, and faithful constraint extraction are violated in deployment).
    
\item {\bf Safeguards}
    \item[] Question: Does the paper describe safeguards that have been put in place for responsible release of data or models that have a high risk for misuse (e.g., pre-trained language models, image generators, or scraped datasets)?
    \item[] Answer: \answerNA{}
    \item[] Justification: The paper does not release a high-risk pretrained model, image generator, language model, or scraped dataset. The released assets are experiment code and generated or benchmark-based task-planning data, so special misuse safeguards are not applicable.

\item {\bf Licenses for existing assets}
    \item[] Question: Are the creators or original owners of assets (e.g., code, data, models), used in the paper, properly credited and are the license and terms of use explicitly mentioned and properly respected?
    \item[] Answer: \answerYes{}
    \item[] Justification: The paper cites the existing datasets, tools, models, and libraries it uses, including PRB, GW/S3E, ALFWorld/ALFRED, Fast Downward, GPT-4.1, Qwen3-VL, and Gemma 4. The appendix explicitly states the licenses or terms of use for the evaluation domains, S3E, Fast Downward, GPT-4.1 API usage, and the open-weight VLMs.

\item {\bf New assets}
    \item[] Question: Are new assets introduced in the paper well documented and is the documentation provided alongside the assets?
    \item[] Answer: \answerYes{}
    \item[] Justification: The paper introduces experiment code and generated evaluation data, provides them with documentation in a public repository (\coderepo), and documents the data construction process in \cref{ap:data}. Since the data are generated from benchmark or simulated task domains, consent from human participants is not applicable.

\item {\bf Crowdsourcing and research with human subjects}
    \item[] Question: For crowdsourcing experiments and research with human subjects, does the paper include the full text of instructions given to participants and screenshots, if applicable, as well as details about compensation (if any)? 
    \item[] Answer: \answerNA{}
    \item[] Justification: The paper does not involve crowdsourcing or research with human subjects. The experiments use generated or benchmark-based task-planning instances and VLM-based state estimation.

\item {\bf Institutional review board (IRB) approvals or equivalent for research with human subjects}
    \item[] Question: Does the paper describe potential risks incurred by study participants, whether such risks were disclosed to the subjects, and whether Institutional Review Board (IRB) approvals (or an equivalent approval/review based on the requirements of your country or institution) were obtained?
    \item[] Answer: \answerNA{}
    \item[] Justification: The paper does not involve crowdsourcing or research with human subjects. Therefore, IRB approval or equivalent review is not applicable.

\item {\bf Declaration of LLM usage}
    \item[] Question: Does the paper describe the usage of LLMs if it is an important, original, or non-standard component of the core methods in this research? Note that if the LLM is used only for writing, editing, or formatting purposes and does \emph{not} impact the core methodology, scientific rigor, or originality of the research, declaration is not required.
    \item[] Answer: \answerYes{}
    \item[] Justification: The paper explicitly describes the use of a VLM as the symbolic state estimator and explains how token probabilities for answers such as ``Yes'' and ``No'' are normalized into per-feature probabilities. \cref{sec:experiments} further states that GPT-4.1 is used as the fixed VLM backbone in the main text, and \cref{ap:backbones} lists the five open-weight VLMs used in the multi-backbone study.

\end{enumerate}

\end{document}